\documentclass[sigconf]{acmart}

\AtBeginDocument{%
  }

\setcopyright{acmlicensed}
\copyrightyear{2024} 
\acmYear{2024}
\acmDOI{XXXXXXX.XXXXXXX}

\acmConference[KDD' 24]{KDD'24 August 25--29, 2024, Barcelona, BCN, Spain}{August 25--29, 2024}{Barcelona, BCN, Spain}
\acmISBN{978-1-4503-XXXX-X/18/06}

\usepackage{multirow}
\usepackage{algorithm}
\usepackage{algpseudocode}
\usepackage{epstopdf}
\usepackage{array} 
\usepackage{hyperref}
\usepackage[graphicx]{realboxes}

\usepackage[utf8]{inputenc} % allow utf-8 input
\usepackage[T1]{fontenc}    % use 8-bit T1 fonts

\definecolor{roarblue}{HTML}{2A2A6B}
\definecolor{roarred}{HTML}{DC536F}
\definecolor{customcolor}{HTML}{DA335C}
\definecolor{citecustomcolor}{HTML}{2B6691}
\hypersetup{
    colorlinks=true,
    linkcolor=customcolor,
    filecolor=customcolor,
    urlcolor=customcolor,
    citecolor=citecustomcolor
}
\usepackage{url}            % simple URL typesetting
\usepackage{booktabs}       % professional-quality tables
\usepackage{amsfonts}       % blackboard math symbols
\usepackage{nicefrac}       % compact symbols for 1/2, etc.
\usepackage{microtype}      % microtypography
\usepackage{xcolor}         % colors
\usepackage{algorithm}
\usepackage{algpseudocode} 
\usepackage[capitalize]{cleveref}
\crefname{section}{Sec.}{Secs.}
\Crefname{section}{Section}{Sections}
\Crefname{table}{Table}{Tables}
\crefname{table}{Tab.}{Tabs.}
\usepackage{colortbl}
\usepackage{multirow}
\usepackage{arydshln}
\usepackage{bibunits}
\usepackage{xspace}

\newcommand\blfootnote[1]{%
  \begingroup
  \renewcommand\thefootnote{}\footnote{#1}%
  \addtocounter{footnote}{-1}%
  \endgroup
}
\usepackage[font=small,labelfont=bf]{caption}
\copyrightyear{2024}
\acmYear{2024}
\setcopyright{rightsretained}
\acmConference[KDD '24]{Proceedings of the 30th ACM SIGKDD Conference on Knowledge Discovery and Data Mining}{August 25--29, 2024}{Barcelona, Spain}
\acmBooktitle{Proceedings of the 30th ACM SIGKDD Conference on Knowledge Discovery and Data Mining (KDD '24), August 25--29, 2024, Barcelona, Spain}\acmDOI{10.1145/3637528.3671724}
\acmISBN{979-8-4007-0490-1/24/08}
\begin{document}

\title{CAFO: Feature-Centric Explanation on Time Series Classification}

\author{Jaeho Kim}
    \email{kjh3690@unist.ac.kr}
\affiliation{%
    \small\institution{Ulsan National Institute of Science
and Technology (UNIST) \\
Artificial Intelligence Graduate School (AIGS)}
    \city{Ulsan}
    \country{South Korea}
}

\author{Seok-Ju Hahn}
\email{seokjuhahn@unist.ac.kr}
\affiliation{%
    \small\institution{Ulsan National Institute of Science
and Technology (UNIST) \\
Department of Industrial Engineering (IE)}
    \city{Ulsan}
    \country{South Korea}
}
\author{Yoontae Hwang}
\email{yoontae@unist.ac.kr}
\affiliation{%
    \small\institution{Ulsan National Institute of Science
and Technology (UNIST) \\
Department of Industrial Engineering (IE)}
    \city{Ulsan}
    \country{South Korea}
}
\author{Junghye Lee}
\email{junghye@snu.ac.kr}
\affiliation{%
    \small\institution{Seoul National University (SNU) \\
Technology Management, Economics and Policy Program \& \\
Graduate School of Engineering Practice \& \\
Institute of Engineering Research}
    \city{Seoul}
    \country{South Korea}
}
\author{Seulki Lee$^\dag$}
\email{seulki.lee@unist.ac.kr}
\affiliation{%
    \small\institution{Ulsan National Institute of Science
and Technology (UNIST) \\
Computer Science and Engineering (CSE) \& \\ 
Artificial Intelligence Graduate School (AIGS)}
    \city{Ulsan}
    \country{South Korea}
}

\begin{abstract}
In multivariate time series (MTS) classification, finding the important features (e.g., sensors) for model performance is crucial yet challenging due to the complex, high-dimensional nature of MTS data, intricate temporal dynamics, and the necessity for domain-specific interpretations. Current explanation methods for MTS mostly focus on time-centric explanations, apt for pinpointing important time periods but less effective in identifying key features. This limitation underscores the pressing need for a feature-centric approach, a vital yet often overlooked perspective that complements time-centric analysis. To bridge this gap, our study introduces a novel feature-centric explanation and evaluation framework for MTS, named  \textbf{CAFO} (\textbf{C}hannel \textbf{A}ttention and \textbf{F}eature \textbf{O}rthgonalization). CAFO employs a convolution-based approach with channel attention mechanisms, incorporating a depth-wise separable channel attention module (DepCA) and a QR decomposition-based loss for promoting feature-wise orthogonality. We demonstrate that this orthogonalization enhances the separability of attention distributions, thereby refining and stabilizing the ranking of feature importance. This improvement in feature-wise ranking enhances our understanding of feature explainability in MTS. Furthermore, we develop metrics to evaluate global and class-specific feature importance. Our framework's efficacy is validated through extensive empirical analyses on two major public benchmarks and real-world datasets, both synthetic and self-collected, specifically designed to highlight class-wise discriminative features. The results confirm CAFO's robustness and informative capacity in assessing feature importance in MTS classification tasks. This study not only advances the understanding of feature-centric explanations in MTS but also sets a foundation for future explorations in feature-centric explanations. The codes are available at \href{https://github.com/eai-lab/CAFO}{https://github.com/eai-lab/CAFO}.

\end{abstract}






\maketitle
\section{Introduction}
With the advancement of Internet of Things (IoT) technologies, time series classification (TSC) tasks have proliferated in recent years. A notable characteristic of TSC data derived from these sources is that they are usually 1) multivariate; that is, they contain multiple measurements or sensors and 2)~characterized by patterns that are complex and intertwined, which poses challenges for semantic interpretation~\cite{beuchert2020overcoming,siddiqui2019tsviz} by humans, a stark contrast to more intuitively graspable domains of image and text data. These multivariate time series~(MTS) data have found practical applications ranging from the classification of human activities to the detection of industrial faults~\cite{filonov2016multivariate,kim2023multi}, demonstrating its broad applicability. \blfootnote{$^{\dag}$ S. Lee is the corresponding author}

\begin{figure*}[ht]
\centerline{\includegraphics[scale=0.27]{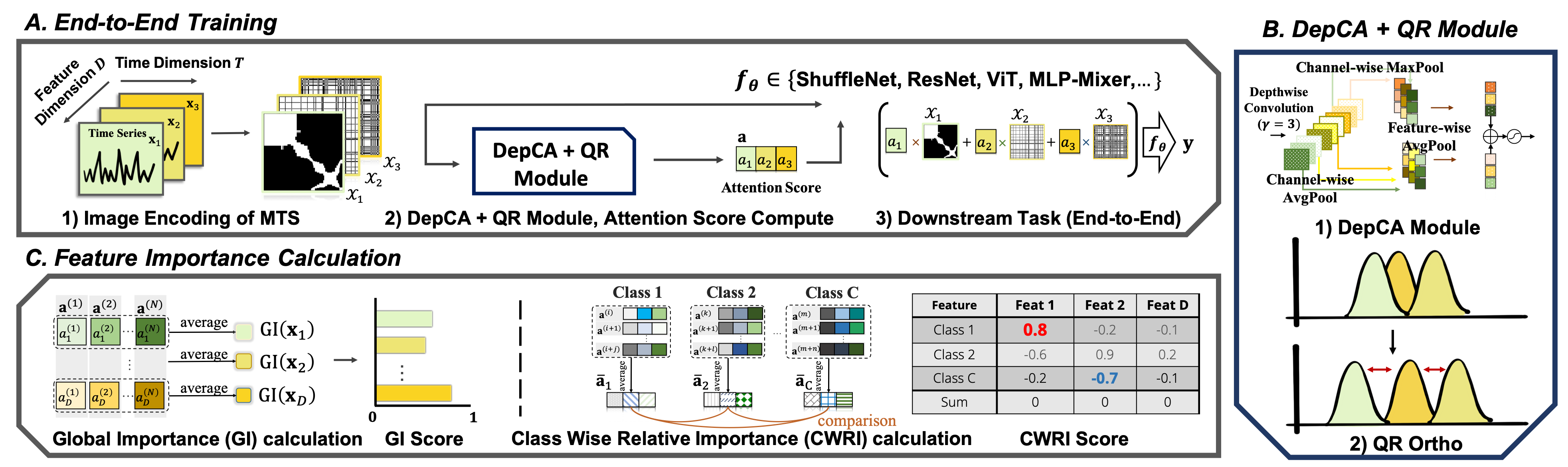}}
\vspace{-10pt}
\caption{Overview of CAFO: (A) End-to-end training. Raw time series are converted into images using image encoding methods, followed by the extraction of channel-wise attention scores using the DepCA+QR Module. These attention scores are element-wise multiplied to image features for end-to-end model training. (B)~DepCA assesses feature contributions, while QR-Ortho Loss minimizes feature redundancy through orthogonality regularization. (C) Feature Importance Calculation. The calculated attention scores are utilized to explain MTS data via Global Importance (GI) and Class-Wise Relative Importance (CWRI) metrics.}
\label{fig:overview}
\vspace{-10pt}
\end{figure*}

As MTS data find broader applications, an essential need emerges in the phase of model development. Engineers and domain experts seek not just to use these models but to understand how they process data. This understanding is vital; it can drastically reduce computational and manufacturing costs and foster confidence in the model's deployment, ensuring it leverages features recognized as important~\cite{xu2019explainable}. In this context, the role of explainable AI (XAI) is not just beneficial but indispensable. Yet, a concerning observation arises: XAI research in MTS has predominantly concentrated on generating time-step-specific or instance-specific explanations~\cite{crabbe2021explaining,tonekaboni2020went,turbe2023evaluation,ismail2020benchmarking,schlegel2019towards}, focusing narrowly on segments of time critical to the model's decision-making in a given instance. Such local explanations, while invaluable in contexts like healthcare, reveal a significant gap for a more comprehensive, feature-centric overview that can provide a broader understanding of the data. In MTS, a `feature' \textit{is commonly identified as a separate channel or measurement variable, which is independent of the time axis}. This need is particularly acute in both industry and academia, where a global understanding of TSC tasks, at both the class level and across the model as a whole, is crucial. However, previous MTS XAI works~\cite{bento2021timeshap,hsieh2021explainable} have addressed this topic in a limited manner, leaving room for more extensive exploration and discussion. 

\textbf{Herein lies the main theme of our paper: we address the imperative need for a feature-centric explanation} - a perspective that is not just complementary but essential to the time-centric explanation in the MTS classification task.

To illustrate the practical impact of our approach, consider the example of a smart shoe manufacturer employing a deep learning model for classifying user activities based on sensor data (accelerometers, force sensors, etc.). Unlike previous methods that used fourteen sensors (or features) for~95\% accuracy~\cite{kim2023multi}, our feature-centric analysis achieves a near-comparable~94\% accuracy with just three key sensors (See \cref{roar_overall} where we sequentially dropped important and unimportant features and measured the model performance). This insightful identification of key sensors would not have been easily attainable through previous time-step or local instance-based explanation methods. Conversely, relying on the three least important sensors, as identified by our model, drops accuracy to~71\%, highlighting the critical nature of sensor selection. Moreover, this revelation goes beyond mere accuracy; it offers manufacturers and engineers invaluable insights. Manufacturers can reduce costs by focusing on essential sensors, potentially eliminating redundant ones. Engineers gain insights for optimizing model performance. These scenarios, spanning industrial and academic fields, introduce the broader application of feature-centric explanation in MTS classification tasks. We present detailed use cases motivating feature-centric explanation throughout this paper and in \cref{appendix:motivation}.

While a number of studies have explored time-step based explanations in depth, feature-based explanations have typically been addressed as a secondary focus or in a limited scope~\cite{bento2021timeshap, hsieh2021explainable}. To our knowledge, there has been a lack of discussion regarding feature-centric explanations in TSC in deep learning, especially with regard to multivariate data (i.e., MTS)\footnote{Our focus is on multivariate time series as we provide feature-centric explanations.}, presented as a comprehensive research paper. The lack of explanation strategy, appropriate MTS datasets, characterized by clearly defined feature importance, alongside matching evaluation protocols, presents a considerable challenge~\cite{schlegel2019towards} in the realm of MTS XAI. Our paper bridges this gap by presenting the \textbf{following contributions}:
\begin{enumerate}
    \item \textbf{Methodologies.} Introduction of Channel Attention and Feature Orthogonalization (CAFO): (1) DepCA, a novel convolution-based framework that utilizes channel attention, which measures feature importance and (2) QR-Ortho, a QR decomposition-based regularizer that ensures feature separability for improved feature-centric explanation.
    \item \textbf{Datasets.} Compilation of both synthetic and real world datasets collected with known class-discriminative features.
    \item \textbf{Metrics.} Development of a comprehensive set of metrics aimed at quantifying global and class-specific feature importance, complete with a corresponding evaluation protocol.
    \item \textbf{Experiments.} Extensive number of empirical evaluations confirming the practical value of the proposed work.
\end{enumerate}

\section{Preliminaries and Related Works}
We first formalize the notation used in our work. To better understand, it is beneficial to have a big overview of CAFO depicted in \cref{fig:overview}. Subsequently, the rest of the section illustrates prior works that are helpful in understanding our method. 

\subsection{Preliminaries}
Given $N$ number of MTS samples, the $i$-th MTS instance $\mathbf{X}^{(i)}=\left[\mathbf{x}^{(i)}_1,...,\mathbf{x}^{(i)}_D\right]\in\mathbb{R}^{T \times D}$ encompasses $T$ times steps and $D$ features. In this context, a univariate time series is defined as a single $j$-th feature sequence $\mathbf{x}^{(i)}_j=[x_1,...,x_T]^{\top}$, and an aggregation of such univariate sequences constitutes an MTS. Traditional time-step-based explanations offer insights along the temporal ($T$) dimension, but our research pivots towards elucidating the feature ($D$) dimension. Consequently, our primary interest lies in discerning the significance of each feature, establishing a hierarchy of feature importance that is both global and class-specific. Specific to our problem setting, rather than using the raw MTS as-is, we transform each feature $\mathbf{x}_j$ into a single-channel image of size $\{0,1\}^{T \times T}$ using image encoding methods. As a whole, the raw MTS input $\mathbf{X}^{(i)}\in\mathbb{R}^{T \times D}$ is converted into an image of size $\mathcal{X}^{(i)}\in\mathbb{R}^{D\times T \times T}$. Employing a channel attention (CA) module, we compute a set of attention scores along the $D$ dimension $\text{CA}(\mathcal{X}^{(i)}) = \mathbf{a}^{(i)} = [a_1,...,a_D]^{\top}$ within the range $[0,1]^{D}$, where each $a_{j}$ represents the attention allocated to the $j$-th feature channel $\mathcal{X}_{j}$ (equivalent to $\mathbf{x}_{j}$) .

\begin{figure*}[ht!]
\vspace{-2pt}
\centerline{\includegraphics[scale=0.315]{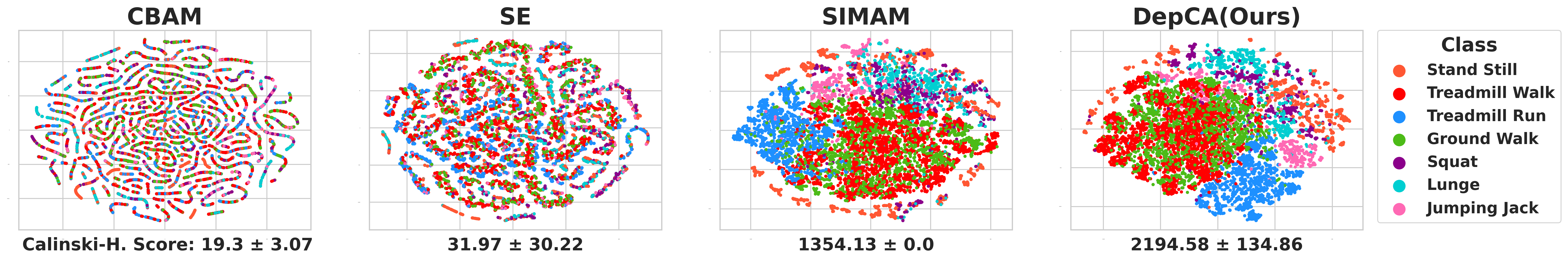}}
\vspace{-8pt}
\caption{Visualization of the channel attention (CA) values using t-SNE~\cite{van2008visualizing} for CBAM~\cite{woo2018cbam}, SE~\cite{hu2018squeeze}, SIMAM~\cite{yang2021simam}, and the proposed DepCA module on the GILON dataset~\cite{kim2023multi}. The Calinski-Harabasz score~\cite{calinski1974dendrite} at the bottom indicates their clustering performance (the higher, the better). As observed, DepCA effectively captures sample and class-specific information even though the CA scores are computed in the early layer of the network, in contrast to existing methods~\cite{woo2018cbam,hu2018squeeze} that compute the CA scores in latent channel spaces (middle layers of the network)}.
\label{fig:tsne_comparison}
\vspace{-12pt}
\end{figure*}

\subsection{Image Encoding of MTS}
Image encoding of MTS involves transforming time series data into image formats, such as a Recurrence Plot (RP)~\cite{eckmann1995recurrence} or Gramian Angular Fields (GAF)~\cite{wang2015encoding}. Encoding time series into images offers several benefits in analyzing feature-wise importance. First, image encoding operates independently of specific standardization methods~\cite{eckmann1995recurrence}, which is often crucial due to the heterogeneity of features, e.g., the varying scales of accelerometers and gyroscopes. This can even significantly impact the end performance of a modeling~\cite{zhao2020deep}. Image encoding, however, primarily relies on point-wise relations (e.g., inner products with threshold) to represent time series, thereby liberating the feature representation from the implicit biases of any particular scaling method. This aspect ensures a more equitable comparison and ranking of features. Second, image encoding enhances the representation of temporal dependency within features. By converting the original feature $\mathbf{x}_{j}\in\mathbb{R}^{T}$ into a  $\mathcal{X}_{j}\in\mathbb{R}^{T\times T}$ image, recurrent patterns become more explicit and discernible~\cite{eckmann1995recurrence}, which allows for the use of well-curated vision models, e.g., ViT~\cite{dosovitskiy2020image}. Our empirical results~(\cref{appendix:GI_FULL}) suggest these models more effectively discern feature importance in our study, potentially owing to image encoding or inherent model capabilities. We use Recurrence Plot to capture the recurrence patterns in MTS, and explore alternative encoding techniques such as GAF in \cref{appendix:UseOfOtherImageEncodings}.

\subsection{Channel Attention (CA) Modules} CA is primarily used in the image classification domain to improve model performance by emphasizing relevant feature channels. The pioneering SENet~\cite{hu2018squeeze}, BAM~\cite{park2018bam}, CBAM~\cite{woo2018cbam}, and SIMAM~\cite{yang2021simam} harness CA by collating channel specific statistics (e.g. global average), passing them through parametrized functions to obtain channel or spatial attention. In contrast to the use of CA in the image domain, which integrates CA at multiple points within the latent space, \textbf{our approach applies CA singularly and directly to the input representation, to obtain the attention scores for each feature}. We note that in the time series domain, the joint usage of image encoding and CA has been previously explored in temporal~\cite{lee2021hierarchical}, frequency~\cite{jiang2022fecam}, and wavelet~\cite{he2021wavelet}-based literature, often to augment model performance, and occasionally to offer interpretative insights via raw attention visualizations. Our research pioneers the use of CA scores to systematically evaluate feature importance on a global and class-specific scale in MTS data.

\subsection{Multivariate Time Series Explanation}

\textbf{Post-hoc explanation in multivariate time series~(MTS)} elucidate model decisions by deriving explanations from their output, making them generally agnostic to the underlying model. In MTS explanation, several post-hoc methods employed in the image domain have been repurposed for MTS by viewing the raw time series as a $T\times D$ image. A recent study by Turbé et al.~\cite{turbe2023evaluation} has undertaken a comprehensive assessment of various post-hoc interpretability methods— Integrated Gradients~\cite{sundararajan2017axiomatic}, GradientSHAP~\cite{lundberg2017unified}, and Shapley value sampling~\cite{castro2009polynomial}—in the context of TSC, highlighting substantial discrepancies in time-centric explanation across methods, also noted by Schlegel et al~\cite{schlegel2019towards}. Our research extends these observations, confirming these inconsistencies in feature-centric explanations of MTS, and first identifying the impact of train/validation distribution on feature importance inconsistency. The study~\cite{turbe2023evaluation} also highlights the limitations on the use of synthetic data in prior time-centric explanation research~\cite{ismail2020benchmarking}, emphasizing the need for real-world datasets with clear discriminative features for validating MTS explanation methods. In the pursuit of enhancing these methods, past works have applied these post-hoc methods on LSTM~\cite{hochreiter1997long}, TCN~\cite{lea2017temporal}, and Transformer\cite{vaswani2017attention} models for time-based explanations. Orthogonal to these approaches, DynaMask~\cite{crabbe2021explaining} is a post-hoc method, providing an explanation based on optimizing perturbation masks for MTS. However, its requirement for numerous optimization steps per instance presents a challenge, limiting its efficiency in global and class-specific importance calculation.

\noindent\textbf{Model-based explanation for time series} rely on specific neural architecture such as recurrent neural networks (RNNs), as these models inherently handle sequential inputs. Nevertheless, recent works show that they suffer from saliency vanishing~\cite{hochreiter1998vanishing} and may have limitations in explaining time series data~\cite{ismail2019input}. For example, TimeSHAP~\cite{bento2021timeshap} is a recurrent explainer extending KernelSHAP~\cite{lundberg2017unified} to the temporal domain by grouping sequential data into coalitions. Shapley-based methods are known to be computationally-intensive~\cite{lundberg2017unified}, while TimeSHAP provides efficient pruning methods to overcome this. However, pruning relies on the assumption that recent events have a predominant influence on model outcome might not apply universally, such as in continuous event recording like human activity monitoring. Another recurrent-based approach, FIT~\cite{tonekaboni2020went}, assigns significance to events using counterfactuals within a generative model. Unfortunately, training a generator adds an extra cost, and the explanation depends on the generator's performance. LAXCAT~\cite{hsieh2021explainable} is another model-based explanation method utilizing both temporal and variable attention scores using 1D convolution methods. Our CAFO method, however, distinguishes itself by employing 2D convolutions and channel attention (CA), offering a unique structural approach to derive attention scores.

\section{CAFO: Channel Attention and Feature Orthogonalization} 

\cref{fig:overview} provides an overview of CAFO for extracting feature-centric importances from MTS data. Starting with image encoding (specifically recurrence plot (RP)~\cite{eckmann1995recurrence}), the raw MTS is transformed into image-like data, where our DepCA module (\cref{subsec:depca}) is used to compute the channel attention score $\mathbf{a}\in\mathbb{R}^D$. These scores are then element-wise multiplied with their respective channels, which are further processed for \textbf{end-to-end training with downstream backbone models} such as a ResNet~\cite{he2016deep}. Throughout the training, we employ a novel QR-Ortho loss~(\cref{subsec:qr-ortho}) to ensure the orthogonality along the feature dimension of the attention, effectively reducing feature redundancy through orthogonalization. Upon completion of the training, we harness these attention scores to compute the Global Importance~(GI) and Class-Wise Relative Importance~(CWRI) metrics,  with further details regarding the metric described in \cref{section:featuremeasures}.

\subsection{Depthwise Channel Attention (DepCA)}
\label{subsec:depca}
In our work, a raw time series $\mathbf{X} {\in} \mathbb{R}^{T\times D}$ is encoded into an image-like format $\mathcal{X}\in\mathbb{R}^{D\times T\times T}$, with each channel representing a distinct feature. The CA (Channel Attention) module evaluates these channels, employing attention scores $\mathbf{a} = [a_1,...,a_D]^{\top}$ for global and class-specific metric computation. It's essential that attention scores,~$\mathbf{a}$, precisely captures each feature's unique information. To achieve this, we introduce the \textbf{Depthwise Channel Attention (DepCA)} module, which surpasses traditional CA techniques in extracting comprehensive information from each channel. Where previous CA methods~\cite{woo2018cbam} rely on simple statistics like global maxima—apt for latent channel spaces for efficiency—our methodology, with its depthwise convolution, allows the model \textit{to learn} informative statistics from each feature representation. By applying depthwise convolutions, we treat each feature independently, capturing distinct details without inter-feature interference. It efficiently extracts clear, differentiated information, as shown by our t-SNE visualizations in \cref{fig:tsne_comparison}.

The DepCA module begins by applying a set of depthwise convolutional filters to the input $\mathcal{X}$. We use $\gamma$ number of filters per each feature channel (with $\gamma=3$ in all experiments). This yields the feature descriptor as $\text{Conv}_{\gamma}(\mathcal{X}) = \mathbf{F_{out}} \in~\mathbb{R}^{(\gamma \times D) \times \text{H} \times \text{W}}$, where $D$ is the number of channels, $\text{H}$ and $\text{W}$ denotes the height and width of the output channels, respectively. Following this, DepCA performs two pooling operations on $\mathbf{F_{out}}$: an average and max pooling, executed channel-wise (CW)~\cite{woo2018cbam} to produce $\mathbf{F_{avg}},\mathbf{F_{max}} \in \mathbb{R}^{\gamma\times D}$. These pooled features are then averaged across the channels coming from the same original feature (i.e., feature-wise; FW). As we aim to provide a feature-centric explanation, we performed FW pooling to get an attention score for each feature. Finally, the two output features are combined through element-wise summation to obtain the aggregated feature representation and then passed to the sigmoid function $\sigma(x)=1/(1+\text{exp}(-x))$, to constrain the CA score between zero and one. Putting it all together, the CA score of the image~$\mathcal{X}$, denoted as $\text{CA}(\mathcal{X}) = [a_1,...,a_D]^{\top} \in [0,1]^{D}
$ where $a_{j}$ is the attention score of~$\mathcal{X}_{j}$, is computed as:
\begin{equation}
\begin{split}
    \mathbf{F_{out}} &= \text{Conv}_{\gamma}(\mathcal{X})
    \\
    \mathbf{F_{avg}} &= \text{CWAvgPool}(\mathbf{F_{out}}), \quad \mathbf{F_{max}} = \text{CWMaxPool}(\mathbf{F_{out}})
    \\
    \text{CA}(\mathcal{X}) & \triangleq\sigma(\text{FWAvgPool}(\mathbf{F_{avg}}) + \text{FWAvgPool}(\mathbf{F_{max}})) \equiv \mathbf{a}
\end{split}
\end{equation}
The CA score, $\mathbf{a}$, is element-wise multiplied with the image representation $\mathcal{X}$, expressed as $\mathbf{a}~\odot~\mathcal{X}$. Here, \textbf{attention values determine the retention or suppression of features}; values near 1 retain features, while those close to 0 suppress them. Consequently, these attention scores are crucial in constructing feature importance metrics and determining the relevance of different features.

\begin{figure}[t]
\centerline{\includegraphics[scale=0.37]{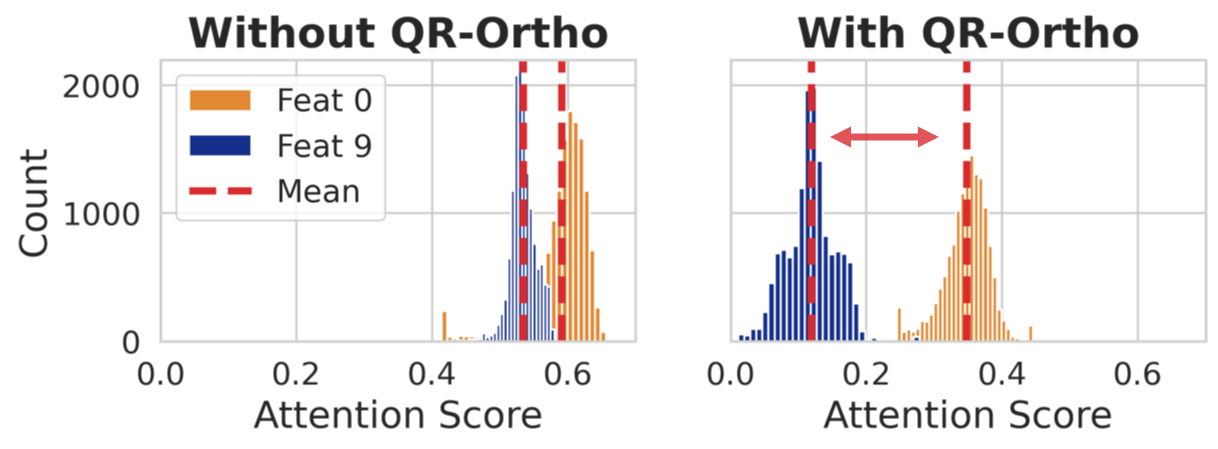}}
\vspace{-12pt}
\caption{Orthogonal regularization on the feature-dimension of the attentions enhances separability. Using QR-Ortho loss, we demonstrate an enhanced distinction between previously overlapping attentions in the Gilon dataset~\cite{kim2023multi}, consistent across five-fold CV.
}
\label{fig:qr_ortho_sep}
\vspace{-12pt}
\end{figure}

% \vspace{-5pt}
\begin{figure*} [htb]
\centerline{\includegraphics[scale=0.3]{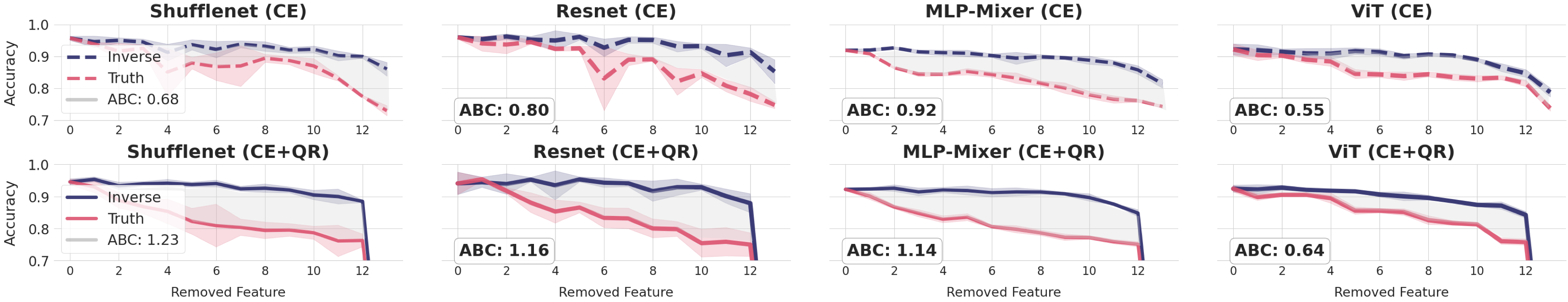}}
\vspace{-8pt}
\caption{RemOve And Retrain (ROAR) with Gilon~\cite{kim2023multi}. The feature ranks of the Gilon task were first identified by our CAFO using the whole 14 feature set, with potential rank variations across figures. To assess the importance of each feature, we systematically removed them from the train and test datasets, ensuring consistency in distribution. This process involved the progressive subtraction of more important (\textcolor{roarred}{red} as `Truth') and less important (\textcolor{roarblue}{blue} as `Inverse') features. After each removal, the model was retrained, and its accuracy was evaluated. The X-axis represents the number of features removed (with zero indicating no removal), while the Y-axis shows the model's accuracy. A notable decline in accuracy is observed with the removal of key features, in contrast to a minimal impact when less important features are omitted. The area between the curve (ABC) metric quantifies the gap between the two curves, where a higher ABC indicates superior feature-wise ranking. The first row exhibits the model's performance using cross-entropy (CE) alone, while the second row shows integration of QR-Ortho (our approach) with CE. A marked improvement in ABC scores across all models is evident, underscoring QR-Ortho's efficacy in identifying pivotal features.}
\label{roar_overall}
\vspace{-12pt}
\end{figure*}

\subsection{Enhancing Feature Separability: QR-Ortho}
\label{subsec:qr-ortho}
During the development of DepCA, we observed an overlapping distribution of CA (channel attention) scores between each feature (see \cref{fig:qr_ortho_sep}). This overlap complicates the derivation of precise feature importance rankings, which are essential for computing both global and class-specific metrics.  To address this issue, we enforce an orthogonal regularization on the feature average of the CA scores to obtain distinguished CA distribution, leading to enhanced and distinct feature rankings of MTS data. \cref{fig:qr_ortho_sep} illustrates the clear separation of previously overlapping CA distributions when orthogonality is enforced. This enhanced separation contributes to three key outcomes: (1) distinct CA distributions for each feature, (2) improved ranking measures for global and class-wise feature importance, and as a result (3) overall better explainability of the MTS data. In our evaluation, we observe a substantial improvement in the explainability of MTS data with feature-wise orthogonality, empirically verifying its critical role in TSC model explanations.

To this end, we propose \textbf{QR-Ortho Loss} that enforces feature-wise orthogonality along the feature dimension of the CA scores through QR decomposition \cite{goodall199313} that factorizes a given matrix $\mathbf{\bar{A}}$ into an orthonormal matrix $\mathbf{Q}$ and a residual upper triangular matrix~$\mathbf{R}$, i.e., $\mathbf{\bar{A}} {=} \textbf{QR}$. Here, the class prototype matrix $\mathbf{\bar{A}}$ is constructed by stacking $\mathbf{\bar{a}}_{c}$ in a row-wise manner, where row-wise means arranging the class-prototype vectors into a matrix format where each class prototype occupies a single row. Given $C$ classes in a TSC task, MTS data can be represented as $\textit{\textbf{K}} {=} \{(\mathbf{X}^{(1)},y^{(1)}),...,(\mathbf{X}^{(N)},y^{(N)})\}$, where $y^{(i)} {\in} \{1,...,C\}$ is the corresponding class label. Then, the class prototype of class~$c$ denoted as $\mathbf{\bar{a}}_{c}$, is defined as the average CA scores of samples belonging to class~$c$, given by:
\begin{equation}
    \mathbf{\bar{a}}_{c} = \frac{1}{|\textit{\textbf{K}}_c|}\sum_{i \in \textit{\textbf{K}}_c}\mathbf{a}^{(i)} 
    \label{eq:class_prototype}
\end{equation}
where $\textit{\textbf{K}}_c=\left\{\left(\mathbf{X}, y\right) \mid y \in c\right\}$ denotes the MTS instances in class $c$.

As such, we denote $\bar{\mathbf{A}}_{:,j}$$(j {=} 1,...,D)$ as the column (feature) vector of $\mathbf{\bar{A}}$ used to perform QR decomposition as $\mathbf{\bar{A}} {=} \textbf{QR}$, given by: %\note{add the dimension of each matrix (A, Q, and R) in the below QR decomposition?, i.e., [C x D] = [C x D][D x D]?}
\begin{equation}
\begin{split}
    % \underbrace{\begin{bmatrix}
    %     | & | & & | \\
    %     \mathbf{\bar{a}}^\top_1 & \mathbf{\bar{a}}^\top_2 & \cdots & \mathbf{\bar{a}}^\top_D \\
    %     | & | & & |
    % \end{bmatrix}}_{\mathbf{\bar{A}}}\\ 
    % &=&
    \underbrace{\begin{bmatrix}
        | & | & & | \\
        \mathbf{q}_1 & \mathbf{q}_2 & \cdots & \mathbf{q}_D \\
        | & | & & |
    \end{bmatrix}}_{\mathbf{Q}}
    \underbrace{\begin{bmatrix}
        \langle \mathbf{q}_1, \bar{\mathbf{A}}_{:,1} \rangle & \langle \mathbf{q}_1, \bar{\mathbf{A}}_{:,2} \rangle & \cdots & \langle \mathbf{q}_1, \bar{\mathbf{A}}_{:,D} \rangle \\
        0                       & \langle \mathbf{q}_2, \bar{\mathbf{A}}_{:,2} \rangle & \cdots & \langle \mathbf{q}_2, \bar{\mathbf{A}}_{:,D} \rangle \\
        \vdots                  & \vdots                  & \ddots & \vdots                  \\
        0                       & 0                       & \cdots & \langle \mathbf{q}_D, \bar{\mathbf{A}}_{:,D} \rangle
    \end{bmatrix}}_{\mathbf{R}}
\end{split}
\end{equation}

The \textbf{Q} matrix embodies the orthogonal basis of the feature dimension of $\mathbf{\bar{A}}$, and the upper diagonal elements of the \textbf{R} matrix, i.e., $\mathbf{R}_{ij}(i \leq j) = \langle\mathbf{q}_{i}, \bar{\mathbf{A}}_{:,j}\rangle$, signify the dot products between class feature representation $\bar{\mathbf{A}}_{:,j}$ and the orthonormal basis $\mathbf{q}_{i}$. The decomposition process ensures direct orthogonality, as the orthonormal columns of $\mathbf{Q}$ inherently exhibit orthogonal properties. Also, by leveraging the widely-used Gram-Schmidt~\cite{bjorck1994numerics} or Householder~\cite{schreiber1989storage} algorithms, it maintains numerical stability while guaranteeing a unique set of orthogonal vectors.

Thus, by penalizing the upper off-diagonals of~$\mathbf{R}$,~i.e.,~$\mathbf{R}_{ij}(i {<} j)$, feature-wise orthogonality of the channel attentions can be effectively regularized. From this, we define QR-Ortho loss as:
\begin{equation}
\label{eq:qr_ortho}
\mathcal{L}_{\texttt{QR}}=\sum\nolimits_{i < j} |\mathbf{R}_{ij}|
\end{equation}
which is to be minimized in addition to the cross-entropy (CE) loss $\mathcal{L}_{\texttt{CE}}$ with the hyperparameter $\lambda$ that controls the strength of orthogonality as $\mathcal{L} = \mathcal{L}_{\texttt{CE}} + \lambda  \mathcal{L}_{\texttt{QR}}$. The loss jointly optimizes the DepCA module and the downstream model end-to-end. In practice, the class prototype matrix $\mathbf{\bar{A}}$ is formed in a mini-batch, with QR-Ortho Loss details in \cref{appendix:qr_algo}.
\section{Feature Explanation Measures}
\label{section:featuremeasures}
\begin{figure}[t]
\centerline{\includegraphics[scale=0.18]{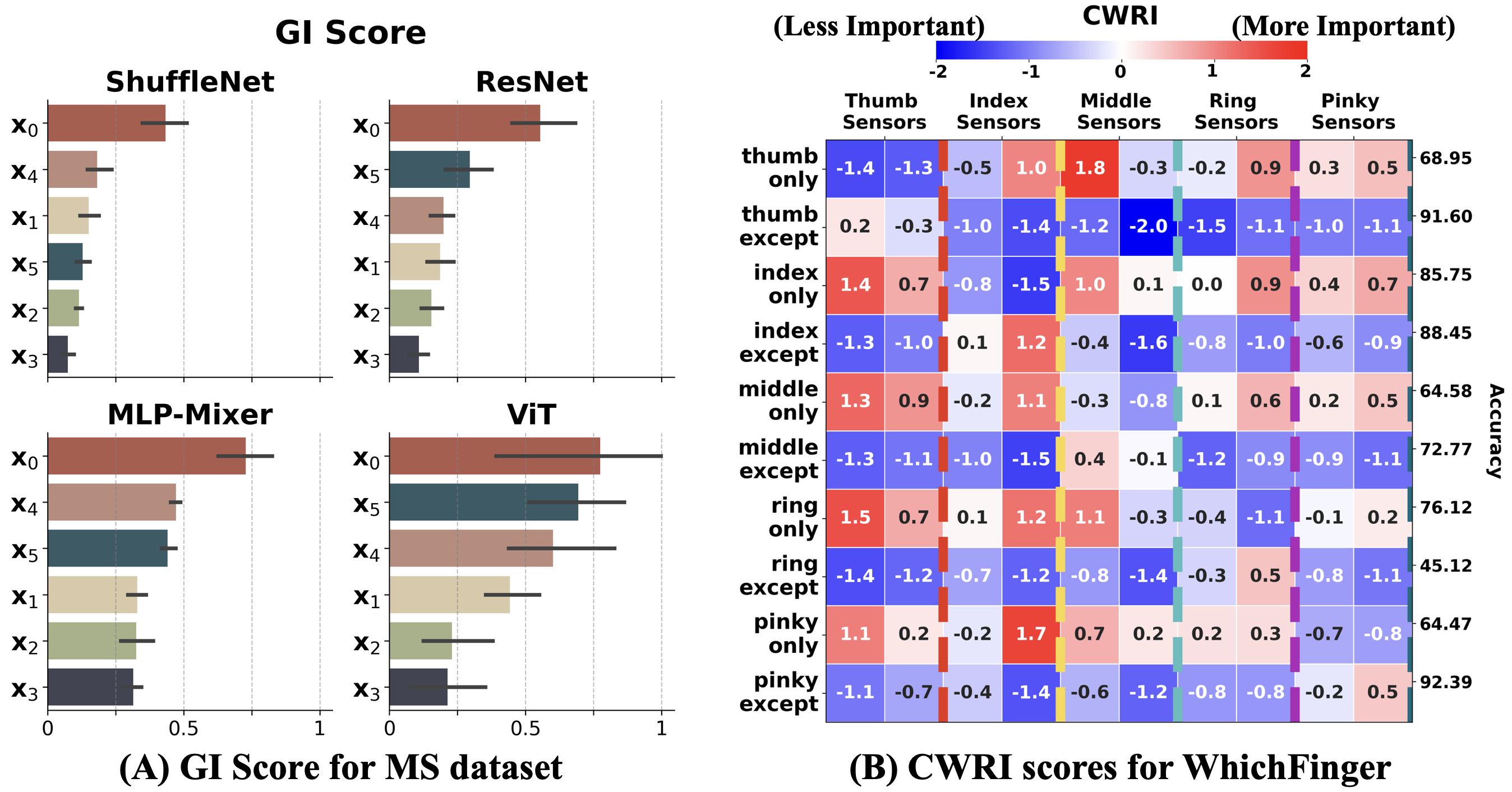}} %0.18 defa
\vspace{-8pt}
\caption{(A) The GI (Global Importance) score for the MS Dataset~\cite{morris2014recofit} is provided. $\mathbf{x}_{0}$ to $\mathbf{x}_{5}$ denotes the feature index. (B)~An example of WhichFinger's CWRI (Class-Wise Relative Importance) score: columns represent sensors (features), rows denote classes, and cell values convey CWRI scores. Red indicates the higher relative importance of the feature for the class, whereas blue denotes features of lesser importance in the context of the specific class.
}
\label{fig:cwri_gi_scores}
\vspace{-12pt}
\end{figure}

We present two feature importance measures: \textbf{(1) Global Importance (GI) and (2) Class-Wise Relative Importance (CWRI)}, which provide reliable feature-wise explanations of MTS data (\cref{fig:cwri_gi_scores}). While we elucidate GI and CWRI in terms of the attention scores, the computation of both measures does not favor or rely on attention scores. Rather, both measures can be effectively applied in conjunction with any attribution method capable of producing instance-wise attributions, as in prior studies \cite{tonekaboni2020went,crabbe2021explaining}. This is achieved by averaging the scores across the time dimension, aligning with~\cref{eq:gi_eq}. This comprehensiveness ensures that our measures are broadly applicable across various attribution methodologies.

\noindent\textbf{Global Importance (GI)}.
The GI score quantifies the significance of each feature in relation to classification performance over the entire data and thus simplifies the interpretation and comparison between features. A feature with a high GI score is globally essential for accurate classifications, while a low GI score suggests negligible influence on the overall model performance. We denote the GI score of the $j$-th feature $\mathbf{x}_j$ as $\text{GI}(\mathbf{x}_j)$, and it is calculated by averaging the $j$-th channel attention (CA) scores $a_{j} \in [0,1]$ of all data samples over all classes, as shown in~\cref{eq:gi_eq}.
\begin{equation}
    \textbf{GI}(\mathbf{x}_j) = \frac{1}{N}\sum_{i=1}^{N}a_{j}^{(i)}
    \label{eq:gi_eq}
\end{equation}

\noindent\textbf{GI Evaluation}.
The evaluation on the GI score focuses on two aspects: (1) the removal of high GI ranked feature should have a higher drop in model performance compared to low GI ranked feature, and (2)~the order of GI ranks should be consistent within models. For model performance, we employ the renowned RemOve And Retrain (ROAR) method~\cite{hooker2019benchmark}, which sequentially eliminates the most important (truth) and least important (inverse) features before retraining the model to maintain consistent train and test distributions (See~\cref{roar_overall}). Based on ROAR, we report the \textit{Drop-in Accuracy~(DA)}, which is the drop in model performance after excluding 20\% of the important features. Second, to complement the manual selection of~K\% of features to be removed, we introduce the \textit{Weighted Drop in Accuracy~(WDA)} metric, which measures the decrease in accuracy when important features are removed sequentially, giving greater weight to scenarios where a smaller fraction of features is dropped. In real-world scenarios, practitioners and developers are often more interested in discerning the most or least important features rather than knowing mediocre features. Such scenarios arise when we have to extensively reduce the number of features to fit in a small memory budget or remove features that do not contribute to model performance. This weighting scheme ensures that the impact of removing high ranked GI feature is more pronounced (see \cref{appendix:gi_metric_calculation}). Third, as an enhancement to the ROAR, we introduce the \textit{Area Between Curves~(ABC)} metric (\cref{eq:ABC}) based on the trapezoid rule~\cite{yeh2002using}, quantifying the area enclosed by the inverse~($f(x)$) and truth curve ($g(x)$) between the interval $[a,b]$; a larger ABC value denotes a more precise feature ranking assessment, highlighted as gray areas in \cref{roar_overall}.
\vspace{-2pt}
\begin{equation}
\begin{split}
    \textbf{ABC}&\triangleq\int_a^b (f(x)-g(x))dx \\
    &\simeq \frac{1}{2}\sum_{i=1}^{n-1}(f(x_i)-g(x_i)+f(x_{i+1}) - g(x_{i+1})) \Delta x_i
    \label{eq:ABC}
\end{split}
\end{equation}

To assess the consistency in GI rankings produced for different model runs, we utilize the Spearman correlation ($\rho_{S}$)~\cite{myers2004spearman} and Kendall correlation ($\rho_{K}$)~\cite{abdi2007kendall} based on a 5-fold cross-validation~(CV). 

\noindent\textbf{Class-Wise Relative Importance (CWRI)}.
While GI offers a global view of feature importance, CWRI provides detailed, class-specific insights into the role of each feature for every class $c\in\{1, ... , C\}$. This approach is particularly valuable because a feature with a high GI score may not necessarily be of high importance for each individual class. CWRI, therefore, provides a class-centric perspective of feature importance in MTS data. We define the CWRI score for class $c$, denoted as $\text{CWRI}(c)\in\mathbb{R}^{D}$, as outlined in~\cref{eq:cwri_eq}. For instance, the rows of \cref{fig:cwri_gi_scores}-B shows the CWRI score for each class in the WhichFinger dataset. This class-specific score is derived by evaluating the deviation in the class prototype of class~$c$ in comparison to other classes $k\neq c$.
\begin{equation}
\begin{split}
    \textbf{CWRI}(c) \triangleq {\mathbf{\bar{a}}_{c} {-} \frac{1}{C{-}1}\sum_{k\neq c}\mathbf{\bar{a}}_{k}}, ~~ \mathbf{\bar{a}}_{c} \text{: see \cref{eq:class_prototype}}%\\ \mathbf{\bar{a}}_{c} \text{: see (\cref{eq:class_prototype})}
    \label{eq:cwri_eq}
\end{split}
\end{equation}

A key aspect of CWRI is its relative computation across different classes, ensuring that the sum of CWRI scores for any given feature is zero. This method naturally produces both positive and negative CWRI values for the same feature across various classes. To illustrate, in a situation with three classes, the CWRI for feature $\mathbf{x}_j$ might be +1.7 for Class~1, -1.4 for Class 2, and -0.3 for Class 3. Here, a positive CWRI of feature $\mathbf{x}_j$ for Class 1 indicates higher relative importance of this feature in classifying Class 1 compared to Classes 2 and 3. The negative scores in Classes 2 and 3 indicate that these features were relatively unimportant. This approach in relatively calculating the difference is advantageous over simple class average scoring, which can be ambiguous and less informative, particularly when classes exhibit similar scores.
\vspace{4pt}

\noindent\textbf{CWRI Evaluation.}
We utilize the CWRI scores to categorize each class and feature into relatively important (positive scores, $\mathbf{x}_{j} \geq 0$; red cells in~\cref{fig:cwri_gi_scores}-B) and relatively unimportant sets (negative scores, $\mathbf{x}_{j} < 0$; blue cells). Our evaluation compares these categorized feature sets against the established ground truth to determine the accuracy of feature importance identification. Given the lack of public datasets with known class discriminative feature importance, we created both real-world and synthetic datasets specifically for this purpose (detailed in~\cref{sec:dataset}). The comparison between our identified important/unimportant sets and the ground truths employs binary classification metrics like the F1 score, Jaccard index, and accuracy (we use the term \textit{interpretative accuracy (IACC)} for clarity over standard model accuracy). The methodology for establishing these ground truths is elaborated in~\cref{appendix:cwri_metrics}.
\section{Dataset and Baseline}
\label{sec:dataset}
\textbf{GI Datasets.} For GI measure, we curated two large public datasets, chosen for their substantial size and relevance (\cref{appendix:gi_data}). The Gilon Activity (7-class) comprises 14 features collected from smart insoles utilized by 72 users~\cite{kim2023multi}. The Microsoft (MS) Activity (10-class) contains 6 features from armbands worn by 92 users~\cite{morris2014recofit}. The details of all the datasets can be found in \cref{appendix:gi_data}.

\begin{figure}[t]
\centerline{\includegraphics[scale=0.17]{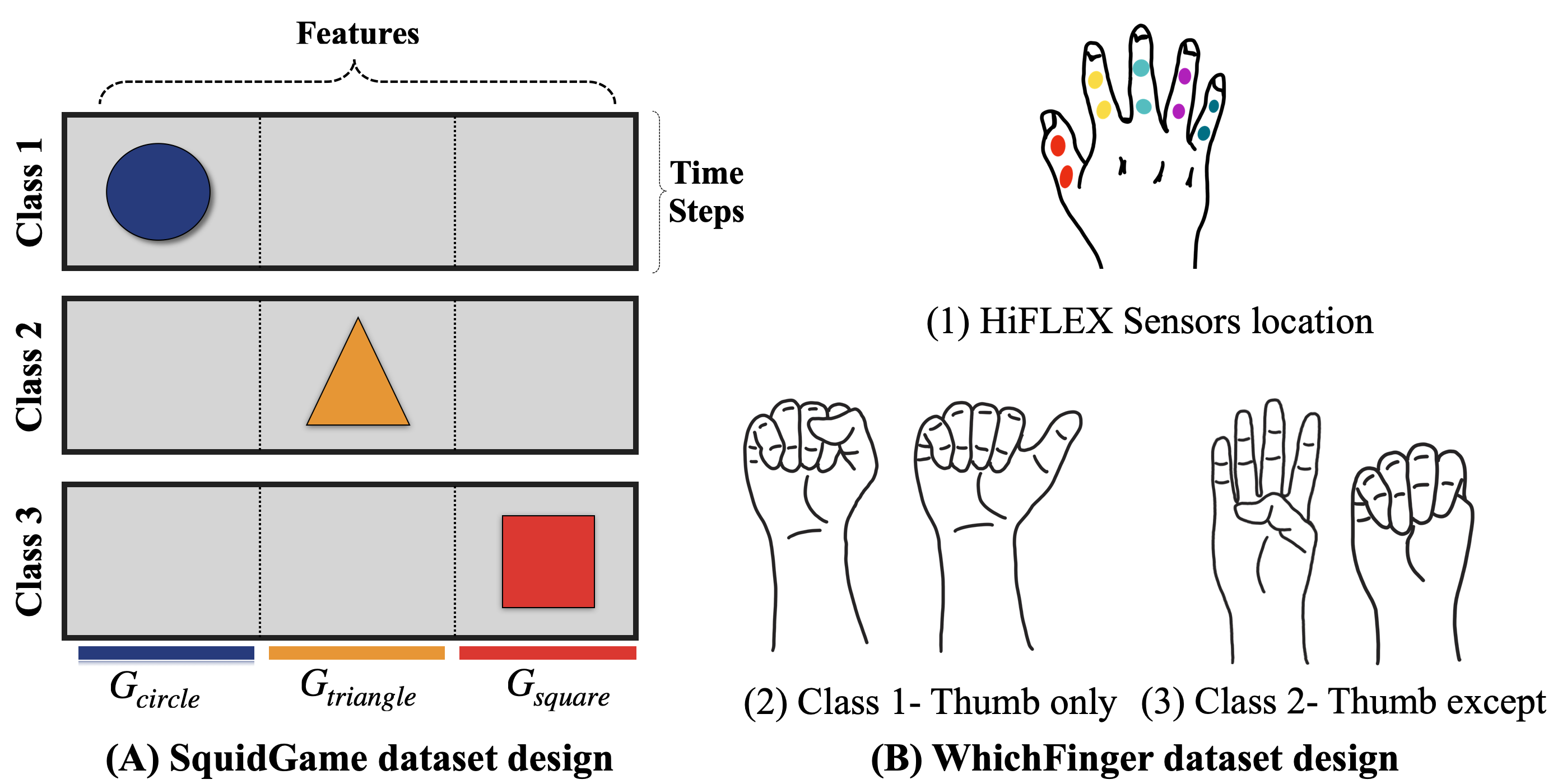}}
\vspace{-12pt}
\caption{\textbf{(A)} Class-specific signals in Circle, Triangle, and Square masks; grey regions are filled with Gaussian noise. \textbf{(B-1)} The smart glove has 10 sensors, two per finger. \textbf{(B-2,~3)} We depict the data acquisition process for Class 1 and 2 for the WhichFinger. Specifically, Class 1 involves folding and unfolding movements of the thumb. In contrast, Class~2 is the complement of Class 1, focusing on the folding and unfolding of the remaining four fingers (See \cref{appendix:whichfinger}). 
}
\label{fig:cwri_design}
\vspace{-12pt}
\end{figure}

\noindent\textbf{CWRI Datasets.} Evaluating the CWRI measure on existing public datasets is challenging due to: (1) lacking in-depth comprehension of the features that influence the performance for each class \cite{turbe2023evaluation}, and (2) unmet requirements for ample classes ($C \geq 3$), features ($D\geq3$), and samples ($N \geq 10,000$) for generalization. Thus, we introduce synthetic and real-world MTS datasets, as follows.

\begin{table*}[h!]
% \usepackage{multirow}
\centering
\fontsize{7.5}{7.7}\selectfont{
\begin{tabular}{cc|cccccc|cccccc}
\hline
 &
   &
  \multicolumn{6}{c|}{\textbf{GILON}} &
  \multicolumn{6}{c}{\textbf{MS}} \\ \cline{3-14} 
\multirow{-2}{*}{\textbf{Models}} &
  \multirow{-2}{*}{\textbf{Methods}} &
  \textbf{ABC($\uparrow$)} &
  \textbf{DA($\uparrow$)} &
  \textbf{WDA($\uparrow$)} &
  \textbf{$\rho_{\text{S}}(\uparrow)$} &
  \textbf{$\rho_{\text{K}}(\uparrow)$} &
  \textbf{ACC($\uparrow$)} &
  \textbf{ABC($\uparrow$)} &
  \textbf{DA($\uparrow$)} &
  \textbf{WDA($\uparrow$)} &
  \textbf{$\rho_{\text{S}}(\uparrow)$} &
  \textbf{$\rho_{\text{K}}(\uparrow)$} &
  \textbf{ACC($\uparrow$)} \\ \hline
 &
  GS &
  0.152 &
  -0.451 &
  0.195 &
  0.105 &
  0.081 &
  {\color[HTML]{FF0000} } &
  -0.159 &
  7.014 &
  0.166 &
  0.051 &
  \textbf{0.066} &
  {\color[HTML]{FF0000} } \\
 &
  SVS &
  0.819 &
  0.092 &
  0.331 &
  0.100 &
  0.068 &
  {\color[HTML]{FF0000} } &
  0.225 &
  \textbf{11.062} &
  0.329 &
  -0.068 &
  -0.066 &
  {\color[HTML]{FF0000} } \\
 &
  Saliency &
  \textbf{1.192} &
  \textbf{4.015} &
  \textbf{0.546} &
  {\color[HTML]{FF0000} \textbf{0.371}} &
  \textbf{0.257} &
  {\color[HTML]{FF0000} } &
  0.039 &
  -4.861 &
  0.142 &
  0.102 &
  0.146 &
  {\color[HTML]{FF0000} } \\
 &
  FA &
  0.714 &
  0.313 &
  0.331 &
  0.148 &
  0.107 &
  {\color[HTML]{FF0000} } &
  {\color[HTML]{FF0000} \textbf{0.491}} &
  4.934 &
  0.287 &
  -0.091 &
  -0.040 &
  {\color[HTML]{FF0000} } \\
 &
  IG &
  0.210 &
  -0.118 &
  0.193 &
  0.045 &
  0.033 &
  {\color[HTML]{FF0000} } &
  -0.015 &
  -2.084 &
  0.241 &
  \textbf{0.074} &
  \textbf{0.066} &
  {\color[HTML]{FF0000} } \\
 &
  CE &
  0.684 &
  3.325 &
  0.478 &
  0.207 &
  0.142 &
  \multirow{-6}{*}{{\color[HTML]{FF0000} \textbf{0.958}}} &
  \textbf{0.355} &
  7.781 &
  {\color[HTML]{FF0000} \textbf{0.475}} &
  0.062 &
  0.013 &
  \multirow{-6}{*}{{\color[HTML]{FF0000} \textbf{0.846}}} \\
\multirow{-7}{*}{ShuffleNet} &
  \cellcolor[HTML]{F2F2F2}CE+QR(Ours) &
  \cellcolor[HTML]{EFEFEF}{\color[HTML]{FF0000} \textbf{1.227}} &
  \cellcolor[HTML]{F2F2F2}{\color[HTML]{FF0000} \textbf{8.157}} &
  \cellcolor[HTML]{EFEFEF}{\color[HTML]{FF0000} \textbf{0.760}} &
  \cellcolor[HTML]{EFEFEF}\textbf{0.352} &
  \cellcolor[HTML]{F2F2F2}{\color[HTML]{FF0000} \textbf{0.270}} &
  \cellcolor[HTML]{F2F2F2}\textbf{0.945} &
  \cellcolor[HTML]{F2F2F2}0.337 &
  \cellcolor[HTML]{EFEFEF}{\color[HTML]{FF0000} \textbf{21.03}} &
  \cellcolor[HTML]{F2F2F2}\textbf{0.450} &
  \cellcolor[HTML]{EFEFEF}{\color[HTML]{FF0000} \textbf{0.640}} &
  \cellcolor[HTML]{EFEFEF}{\color[HTML]{FF0000} \textbf{0.546}} &
  \cellcolor[HTML]{F2F2F2}\textbf{0.810} \\ \hline
 &
  GS &
  1.284 &
  \textbf{3.935} &
  {\color[HTML]{FF0000} \textbf{0.671}} &
  \textbf{0.453} &
  \textbf{0.318} &
  {\color[HTML]{FF0000} } &
  0.059 &
  {\color[HTML]{FF0000} \textbf{8.438}} &
  \textbf{0.356} &
  0.051 &
  0.013 &
   \\
 &
  SVS &
  0.577 &
  0.835 &
  0.281 &
  0.134 &
  0.116 &
  {\color[HTML]{FF0000} } &
  0.160 &
  4.520 &
  0.344 &
  \textbf{0.360} &
  \textbf{0.280} &
   \\
 &
  Saliency &
  {\color[HTML]{FF0000} \textbf{1.338}} &
  2.652 &
  \textbf{0.663} &
  0.364 &
  0.287 &
  {\color[HTML]{FF0000} } &
  0.154 &
  \textbf{8.327} &
  0.345 &
  0.251 &
  0.200 &
   \\
 &
  FA &
  0.847 &
  0.118 &
  0.440 &
  -0.014 &
  -0.024 &
  {\color[HTML]{FF0000} } &
  {\color[HTML]{FF0000} \textbf{0.350}} &
  7.037 &
  {\color[HTML]{FF0000} \textbf{0.373}} &
  0.108 &
  0.066 &
   \\
 &
  IG &
  \textbf{1.289} &
  2.137 &
  0.620 &
  \textbf{0.453} &
  \textbf{0.318} &
  {\color[HTML]{FF0000} } &
  -0.005 &
  7.935 &
  0.244 &
  0.051 &
  0.013 &
   \\
 &
  CE &
  0.801 &
  1.553 &
  0.420 &
  0.370 &
  0.270 &
  \multirow{-6}{*}{{\color[HTML]{FF0000} \textbf{0.960}}} &
  \textbf{0.175} &
  -0.667 &
  0.007 &
  -0.062 &
  -0.066 &
  \multirow{-6}{*}{\textbf{0.752}} \\
\multirow{-7}{*}{ResNet} &
  \cellcolor[HTML]{F2F2F2}CE+QR(Ours) &
  \cellcolor[HTML]{F2F2F2}1.163 &
  \cellcolor[HTML]{F2F2F2}{\color[HTML]{FE0000} \textbf{6.393}} &
  \cellcolor[HTML]{EFEFEF}0.615 &
  \cellcolor[HTML]{EFEFEF}{\color[HTML]{FF0000} \textbf{0.581}} &
  \cellcolor[HTML]{F2F2F2}{\color[HTML]{FF0000} \textbf{0.441}} &
  \cellcolor[HTML]{F2F2F2}\textbf{0.940} &
  \cellcolor[HTML]{EFEFEF}0.117 &
  \cellcolor[HTML]{EFEFEF}1.266 &
  \cellcolor[HTML]{EFEFEF}0.254 &
  \cellcolor[HTML]{EFEFEF}{\color[HTML]{FF0000} \textbf{0.440}} &
  \cellcolor[HTML]{EFEFEF}{\color[HTML]{FF0000} \textbf{0.333}} &
  \cellcolor[HTML]{F2F2F2}{\color[HTML]{FF0000} \textbf{0.787}} \\ \hline
 &
  GS &
  -0.307 &
  0.494 &
  0.112 &
  0.230 &
  0.156 &
   &
  0.294 &
  2.572 &
  0.218 &
  -0.040 &
  0.013 &
  {\color[HTML]{FF0000} } \\
 &
  SVS &
  -0.708 &
  3.330 &
  0.063 &
  0.164 &
  0.112 &
   &
  {\color[HTML]{FF0000} \textbf{0.328}} &
  3.983 &
  0.226 &
  -0.137 &
  -0.120 &
  {\color[HTML]{FF0000} } \\
 &
  Saliency &
  0.986 &
  3.446 &
  0.490 &
  {\color[HTML]{FF0000} \textbf{0.525}} &
  {\color[HTML]{FF0000} \textbf{0.411}} &
   &
  0.321 &
  {\color[HTML]{FF0000} \textbf{6.642}} &
  \textbf{0.263} &
  \textbf{0.040} &
  \textbf{0.040} &
  {\color[HTML]{FF0000} } \\
 &
  FA &
  0.457 &
  1.041 &
  0.269 &
  -0.065 &
  -0.046 &
   &
  0.300 &
  3.209 &
  0.226 &
  -0.045 &
  -0.013 &
  {\color[HTML]{FF0000} } \\
 &
  IG &
  -0.301 &
  0.787 &
  0.115 &
  0.235 &
  0.169 &
   &
  \textbf{0.326} &
  2.788 &
  \textbf{0.263} &
  -0.040 &
  0.013 &
  {\color[HTML]{FF0000} } \\
 &
  CE &
  \textbf{0.920} &
  {\color[HTML]{FF0000} \textbf{8.117}} &
  \textbf{0.531} &
  \textbf{0.330} &
  \textbf{0.239} &
  \multirow{-6}{*}{\textbf{0.920}} &
  0.125 &
  0.730 &
  0.165 &
  -0.142 &
  -0.093 &
  \multirow{-6}{*}{{\color[HTML]{FF0000} \textbf{0.736}}} \\
\multirow{-7}{*}{MLP-Mixer} &
  \cellcolor[HTML]{EFEFEF}CE+QR(Ours) &
  \cellcolor[HTML]{EFEFEF}{\color[HTML]{FF0000} \textbf{1.144}} &
  \cellcolor[HTML]{F2F2F2}\textbf{8.105} &
  \cellcolor[HTML]{EFEFEF}{\color[HTML]{FF0000} \textbf{0.697}} &
  \cellcolor[HTML]{EFEFEF}0.165 &
  \cellcolor[HTML]{EFEFEF}0.129 &
  \cellcolor[HTML]{F2F2F2}{\color[HTML]{FF0000} \textbf{0.922}} &
  \cellcolor[HTML]{EFEFEF}0.276 &
  \cellcolor[HTML]{EFEFEF}\textbf{6.093} &
  \cellcolor[HTML]{EFEFEF}{\color[HTML]{FF0000} \textbf{0.289}} &
  \cellcolor[HTML]{EFEFEF}{\color[HTML]{FF0000} \textbf{0.908}} &
  \cellcolor[HTML]{EFEFEF}{\color[HTML]{FF0000} \textbf{0.813}} &
  \cellcolor[HTML]{F2F2F2}\textbf{0.726} \\ \hline
 &
  GS &
  -0.197 &
  1.244 &
  0.116 &
  0.218 &
  0.147 &
   &
  \textbf{0.169} &
  5.219 &
  0.361 &
  {\color[HTML]{FF0000} \textbf{0.360}} &
  {\color[HTML]{FF0000} \textbf{0.280}} &
   \\
 &
  SVS &
  -0.607 &
  \textbf{2.830} &
  0.083 &
  0.131 &
  0.081 &
   &
  {\color[HTML]{FF0000} \textbf{0.182}} &
  \textbf{5.227} &
  0.318 &
  -0.040 &
  -0.066 &
   \\
 &
  Saliency &
  \textbf{0.587} &
  1.216 &
  0.376 &
  0.120 &
  0.098 &
   &
  0.122 &
  5.219 &
  0.329 &
  0.051 &
  0.040 &
   \\
 &
  FA &
  0.244 &
  0.367 &
  0.223 &
  -0.054 &
  -0.024 &
   &
  {\color[HTML]{FF0000} \textbf{0.182}} &
  \textbf{5.227} &
  0.318 &
  0.141 &
  0.120 &
   \\
 &
  IG &
  -0.176 &
  1.244 &
  0.119 &
  0.200 &
  0.138 &
   &
  0.169 &
  5.219 &
  0.361 &
  \textbf{0.177} &
  \textbf{0.173} &
   \\
 &
  CE &
  0.553 &
  {\color[HTML]{FE0000} \textbf{3.512}} &
  \textbf{0.387} &
  \textbf{0.407} &
  \textbf{0.292} &
  \multirow{-6}{*}{\textbf{0.922}} &
  -0.054 &
  -1.109 &
  0.151 &
  -0.097 &
  -0.066 &
  \multirow{-6}{*}{\textbf{0.703}} \\
\multirow{-7}{*}{ViT} &
  \cellcolor[HTML]{EFEFEF}CE+QR(Ours) &
  \cellcolor[HTML]{EFEFEF}{\color[HTML]{FF0000} \textbf{0.636}} &
  \cellcolor[HTML]{F2F2F2}2.046 &
  \cellcolor[HTML]{EFEFEF}{\color[HTML]{FF0000} \textbf{0.456}} &
  \cellcolor[HTML]{EFEFEF}{\color[HTML]{FF0000} \textbf{0.502}} &
  \cellcolor[HTML]{EFEFEF}{\color[HTML]{FF0000} \textbf{0.389}} &
  \cellcolor[HTML]{F2F2F2}{\color[HTML]{FF0000} \textbf{0.924}} &
  \cellcolor[HTML]{EFEFEF}0.128 &
  \cellcolor[HTML]{EFEFEF}{\color[HTML]{FF0000} \textbf{10.962}} &
  \cellcolor[HTML]{EFEFEF}{\color[HTML]{FF0000} \textbf{0.530}} &
  \cellcolor[HTML]{EFEFEF}0.120 &
  \cellcolor[HTML]{EFEFEF}0.093 &
  \cellcolor[HTML]{F2F2F2}{\color[HTML]{FF0000} \textbf{0.706}} \\ \hline
\end{tabular}}
    \caption{Performance Evaluation of GI Metrics on Gilon and MS datasets. Each performance metric is explained in \cref{section:featuremeasures}. Optimal performance is indicated by values in bold red, while the second-highest performance is marked in bold black. See \cref{appendix:GI_FULL} for full comparison (including standard deviation from five-fold cross validation) of explainers based on raw MTS data such as LSTM, and TCN.}
    \label{tab:GI_full}
\vspace{-20pt}
\end{table*}

\noindent\textbf{Dataset 1: SquidGame} ($C=3; D=10; N=54,000$).  We designed the SquidGame dataset (\cref{fig:cwri_design}-A), a 3-class synthetic MTS data, comprising of 30 features, which are divided into sets $\textbf{\textit{G}}_{\text{circle}}=\{1,...,10\}$, $\textbf{\textit{G}}_{\text{triangle}}=\{11,...,20\}$, and $\textbf{\textit{G}}_{\text{square}}=\{21,...,30\}$. For Class 1, distinct time signals, such as sine waves, are produced within the circular mask in the feature set $\textbf{\textit{G}}_{\text{circle}}$. Meanwhile, Gaussian noise fills the remaining areas (grey region)  outside the circular mask in $\textbf{\textit{G}}_{\text{circle}}$. This process mirrors the approach taken by Ismail et al.~\cite{ismail2020benchmarking}. Similarly, Classes 2 and 3 have unique time signals within their respective feature sets, $\textbf{\textit{G}}_{\text{triangle}}$ and $\textbf{\textit{G}}_{\text{square}}$. For each MTS instance, the location and size of the masks are randomly generated within the three feature sets to increase complexity.

\noindent\textbf{Dataset 2: WhichFinger} ($C=10; D=10; N=18,010$). 
We gathered a real-world MTS dataset using a smart glove \cite{lee2020deep} from 19 users, called WhichFinger  (\cref{fig:cwri_design}-B), to validate the CWRI measure. The smart glove incorporated with two sensors for each finger measures the resistance change in response to the tensile force exerted by each finger. We capture ten unique finger movements, achieved by either flexing and extending a single finger or a group of four fingers. Owing to the interlinked nature of hand muscles, we observe realistic correlations among features, making the task both intricate and non-trivial, providing a valuable MTS dataset for XAI applications. Detailed descriptions of the task design and data collection methodologies are provided in \cref{appendix:whichfinger}.

\noindent\textbf{Implementation and Baselines.} 
We compare CAFO to several post-hoc explanation methods i.e., Gradient Shap~(GS) \cite{lundberg2017unified}, Shapley Value Sampling~(SVS)~\cite{castro2009polynomial}, Saliency~\cite{simonyan2013deep}, Feature Ablation (FA)~\cite{suresh2017clinical}, Integrated Gradients (IG)~\cite{sundararajan2017axiomatic}, and DynaMask (DM) \cite{crabbe2021explaining}. We utilized several deep architectures including vision-based deep models: ShuffleNet~\cite{zhang2018shufflenet}, ResNet~\cite{he2016deep}, MLP-Mixer~\cite{tolstikhin2021mlp}, and Vision Transformer (ViT)~\cite{dosovitskiy2020image}, and sequence based deep models like LSTM~\cite{hochreiter1997long} and TCN. We adopt FIT~\cite{tonekaboni2020went}, an explainer designed for recurrent models. We also employ LAXCAT~\cite{hsieh2021explainable}, a 1-D CNN based MTS explainer. A detailed description is in \cref{appendix:ModelHyperParameters}.

\section{Results}
\subsection{Evaluation of Global Importance}
We evaluated CAFO's performance in the GI measure using Gilon and MS datasets. The results for vision-based models like ShuffleNet, ResNet, MLP-Mixer, and ViT are in \cref{tab:GI_full}. Models using raw MTS format (e.g., LSTM, TCN, LAXCAT) and post-hoc methods relying on raw MTS (e.g., FIT, DynaMask) are detailed in \cref{appendix:GI_FULL}. Generally, vision-based models excel in most scenarios.

CAFO consistently demonstrates superior performance across key metrics (ABC, DA, WDA), highlighted in bold red in the Gilon dataset: notably, ShuffleNet (1.227), MLP-Mixer (1.144), and ViT~(0.636) in the ABC metric. The use of QR-Ortho with cross-entropy (CE) significantly improved GI metric performance in most cases: 10 of 12 in Gilon and 9 of 12 in MS datasets. Notably, several baselines showed negative ABC scores, indicating a mismatch between critical feature identification and the drop in model accuracy, as seen in ROAR's inverse and truth lines (\cref{roar_overall}). This suggests some baseline explainers inadequately measure GI rankings. Our findings show minimal difference in model accuracy between with and without QR-Ortho integration. The regularization parameter $\lambda$ in $\mathcal{L} = \mathcal{L}_{\texttt{CE}} + \lambda \mathcal{L}_{\texttt{QR}}$ was selected based on dataset characteristics, not through exhaustive tuning. We believe a more tailored selection of $\lambda$ for each model may offer additional performance enhancements.

\subsection{Consistency in GI Ranks}
\subsubsection{Within Models}
Establishing consistency in model explanations is imperative for fostering user trust, as highlighted by Riberio et al.~\cite{ribeiro2016should}. Conversely, models that yield divergent feature rankings across multiple runs undermine confidence in user's decision-making processes. Although prior research~\cite{turbe2023evaluation} has underscored the variability in time-step importance produced with post-hoc explanation methods, our study is the initial effort to identify and quantitatively evaluate such variability in the context of feature-based importance. Our evaluation involves executing each model through 5 iterations of CV, with each fold serving once as a validation set and the remaining as training. This process yields 5 distinct feature rankings for the same left-out test set, from which we compute the pairwise Spearman's $\rho_{s}$ and Kendall's $\rho_{k}$ coefficients to gauge rank consistency, with the findings presented in~\cref{tab:GI_full}. Notably, CAFO demonstrates the highest consistency in 11 out of 16 instances. Across the board, our results reveal there exists a huge variability in feature rankings, even under constant model architectures and explanatory methods. Alarmingly, certain explanatory methods yield negative correlations, indicating inverted ranking orders across different runs. These findings raise the need for more robust explanatory frameworks that can deliver dependable and stable feature rankings.

\begin{figure}[t]
\centerline{\includegraphics[scale=0.27]{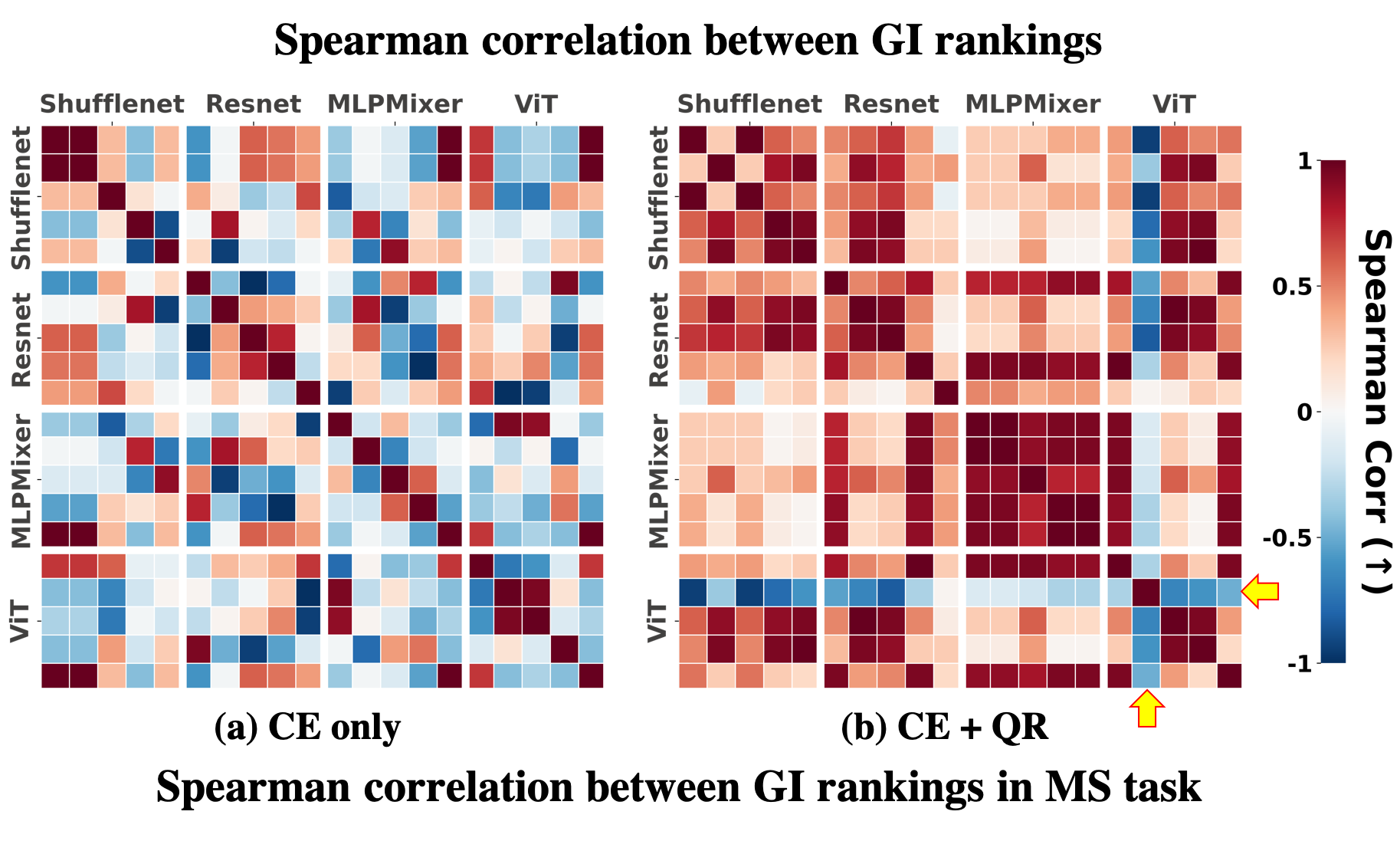}}
\vspace{-12pt}
\caption{
(a) Pairwise Spearman correlation ($\rho_S$) of GI ranks from four models with cross-entropy~(CE) loss in five-fold CV, revealing lower correlations and inconsistent GI ranks. (b)~Enhanced GI rank consistency observed with QR-Ortho integration, demonstrated by higher $\rho_S$ values in both inter and intra-model comparisons.
}
\label{fig:gi_consistency}
\vspace{-16pt}
\end{figure}

\subsubsection{Between Models}
Our analysis extends to assessing feature rank consistency across different models. The Spearman correlation coefficients, visualized as a heatmap in \cref{fig:gi_consistency}, reveal that the use of QR-Ortho significantly improves feature ranking consistency across models compared to CE alone, demonstrating CAFO's robustness in providing consistent feature rankings, independent of model architecture. While different models naturally prioritize varying features for optimal performance, a degree of ranking consistency is a robustness indicator, fostering model trust. Additionally, analyzing ranking discrepancies offers deep insights. For example, as indicated by the yellow arrow in~\cref{fig:gi_consistency}, we observe an anomaly where a single run from the ViT model presents an entirely reversed feature ranking relative to all other runs. Such an outlier warrants further investigation by developers to ascertain the presence of potential errors or anomalies in the model training or data processing. These insights can prove invaluable in enhancing model reliability.

\begin{figure}[t]
\centerline{\includegraphics[scale=0.28]{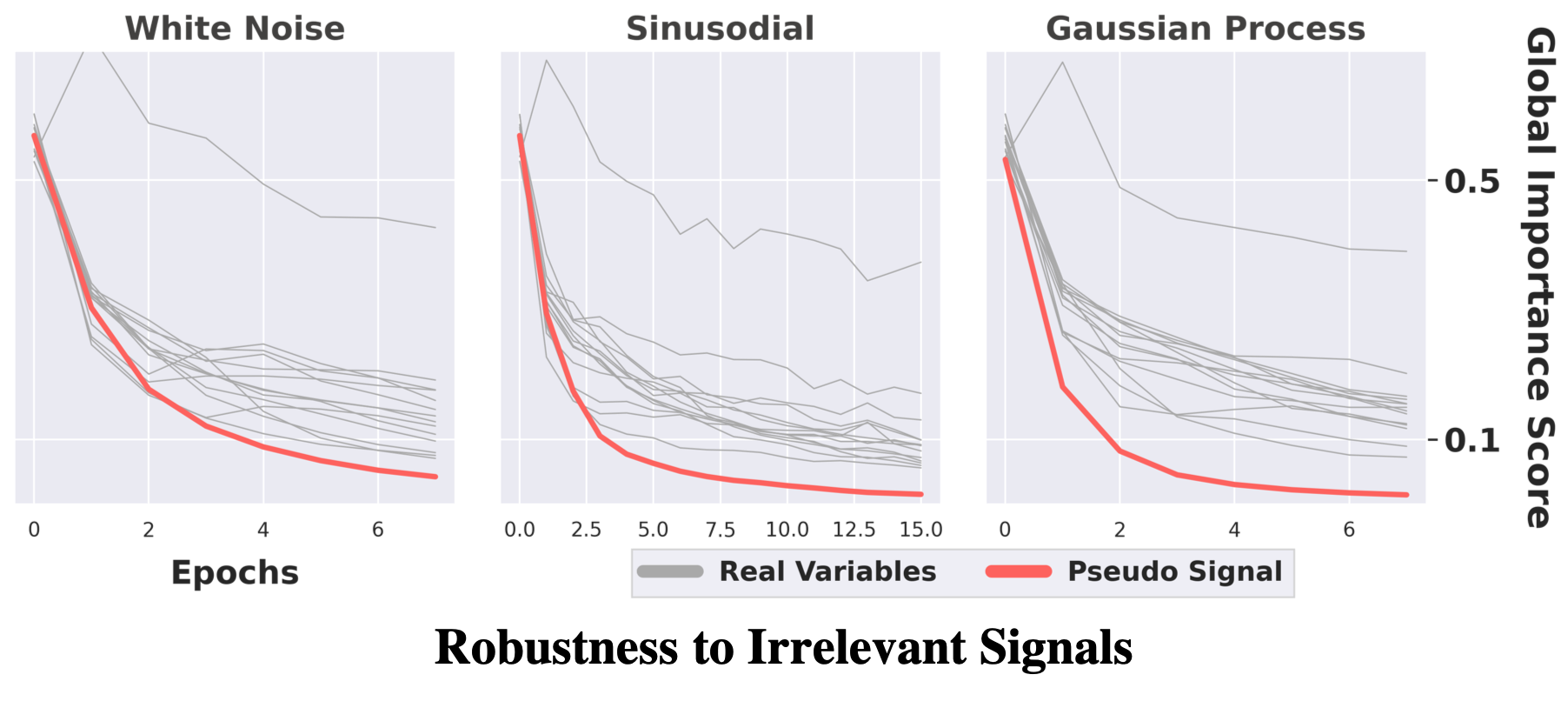}}
\vspace{-12pt}
\caption{The GI metric at each training epoch is visualized, with bold red lines representing the irrelevant signal, while the thin grey lines correspond to the actual variables in the Gilon task. Over the course of training, the pseudo signals (i.e., bold red lines) consistently converge to the lowest GI values.
}
\label{fig:gi_pseudo}
\vspace{-16pt}
\end{figure}

\subsection{Robustness to Irrelevant Signals}
In real world MTS problems, the overabundance of data often results in the accumulation of measurements from superfluous sensors~\cite{west2011automated}. Eliminating such irrelevant feature is, therefore, a critical task for practitioners. To assess the efficacy of CAFO in filtering out insignificant variables during model training, we generate pseudo signals~\cite{maat2017timesynth} from time series processes: White noise, Sinusoidal, and Gaussian Process (detailed in~\cref{appendix:pseudo_vars}). The GI measure from each training epoch is visualized in~\cref{fig:gi_pseudo}. Initially, the pseudo-variable's GI value is near 0.5, but it converges to the lowest GI ranking as training advances. This demonstrates CAFO's robustness against non-significant variables and its potential to identify and discard non-significant features.

\subsection{Class-Wise Relative Importance}
Assessing feature relevance for specific classes is critical in applications like predictive maintenance in Heating, Ventilating, and Air Conditioning (HVAC) systems, where sensor importance varies by fault (class) type~\cite{taheri2021fault, zhang2019systematic} (\cref{appendix:motivation}). Our CWRI methodology offers valuable information for sensor prioritization for each class.

In this study, we evaluate the ability of CAFO to identify critical class-wise features using the CWRI metric on two datasets: SquidGame and WhichFinger. As discussed in \cref{sec:dataset}, these datasets come with ground truth labels indicating the relevance of features on a class-wise basis. As detailed in \cref{table_CWRI}, CAFO surpasses other explanatory models in accurately identifying class-specific features in 13 out of 24 scenarios. We also note that integrating QR-Ortho Loss consistently enhances the discernment of class-wise relevant features across the table, compared to the standalone use of CE~(cross-entropy) loss. Moreover, we observe a general performance degradation of all explainers in the WhichFinger dataset compared to SquidGame dataset, which may be attributed to the increased complexity inherent in real-world data. Such observation underscores the need for the development of real-world data oriented for XAI, especially in the time series domain.

\begin{table}

\centering
\fontsize{5.8}{5.8}\selectfont{
\begin{tabular}{cccccccc}
\hline
 &          & \multicolumn{3}{c}{\textbf{SquidGame}}                                & \multicolumn{3}{c}{\textbf{WhichFinger}}                          \\ \cline{3-8} 
\multirow{-2}{*}{\textbf{Models}} &
  \multirow{-2}{*}{\textbf{Methods}} &
  \textbf{F1($\uparrow$)} &
  \textbf{Jaccard($\uparrow$)} &
  \multicolumn{1}{c:}{\textbf{IACC($\uparrow$)}} &
  \textbf{F1($\uparrow$)} &
  \textbf{Jaccard($\uparrow$)} &
  \textbf{IACC($\uparrow$)} \\ \hline
 & GS       & 0.689          & 0.531          & \multicolumn{1}{c:}{0.649}          & 0.590          & 0.424          & 0.596                                 \\
 & SVS      & \textbf{0.811} & \textbf{0.699} & \multicolumn{1}{c:}{\textbf{0.853}} & \textbf{0.635} & \textbf{0.466} & \textbf{0.636}                        \\
 & Saliency & 0.533          & 0.371          & \multicolumn{1}{c:}{0.489}          & 0.601          & 0.436          & {\color[HTML]{FF0000} \textbf{0.644}} \\
 & FA       & 0.765          & 0.624          & \multicolumn{1}{c:}{0.818}          & 0.552          & 0.383          & 0.588                                 \\
 & IG       & 0.690          & 0.531          & \multicolumn{1}{c:}{0.649}          & 0.582          & 0.414          & 0.588                                 \\
 & CE       & 0.561          & 0.394          & \multicolumn{1}{c:}{0.518}          & 0.429          & 0.273          & 0.530                                 \\
\multirow{-7}{*}{\rotatebox[origin=c]{90}{ShuffleNet}} &
  \cellcolor[HTML]{EFEFEF}CE+QR(Ours) &
  \cellcolor[HTML]{EFEFEF}{\color[HTML]{FF0000} \textbf{0.983}} &
  \cellcolor[HTML]{EFEFEF}{\color[HTML]{FF0000} \textbf{0.967}} &
  \multicolumn{1}{c:}{\cellcolor[HTML]{EFEFEF}{\color[HTML]{FF0000} \textbf{0.978}}} &
  \cellcolor[HTML]{EFEFEF}{\color[HTML]{FF0000} \textbf{0.679}} &
  \cellcolor[HTML]{EFEFEF}{\color[HTML]{FF0000} \textbf{0.514}} &
  \cellcolor[HTML]{EFEFEF}0.606 \\ \hline
 & GS       & 0.747          & 0.598          & \multicolumn{1}{c:}{0.693}          & 0.630          & 0.462          & 0.636                                 \\
 & SVS      & \textbf{0.758} & \textbf{0.621} & \multicolumn{1}{c:}{\textbf{0.800}} & 0.626          & 0.456          & 0.634                                 \\
 & Saliency & 0.489          & 0.329          & \multicolumn{1}{c:}{0.467}          & 0.588          & 0.419          & 0.604                                 \\
 & FA       & 0.739          & 0.603          & \multicolumn{1}{c:}{0.789}          & \textbf{0.670} & \textbf{0.503} & \textbf{0.696}                        \\
 & IG       & 0.746          & 0.597          & \multicolumn{1}{c:}{0.693}          & 0.632          & 0.463          & 0.638                                 \\
 & CE       & 0.524          & 0.362          & \multicolumn{1}{c:}{0.580}          & 0.357          & 0.232          & 0.530                                 \\
\multirow{-7}{*}{\rotatebox[origin=c]{90}{ResNet}} &
  \cellcolor[HTML]{EFEFEF}CE+QR(Ours) &
  \cellcolor[HTML]{EFEFEF}{\color[HTML]{FF0000} \textbf{0.987}} &
  \cellcolor[HTML]{EFEFEF}{\color[HTML]{FF0000} \textbf{0.974}} &
  \multicolumn{1}{c:}{\cellcolor[HTML]{EFEFEF}{\color[HTML]{FF0000} \textbf{0.982}}} &
  \cellcolor[HTML]{EFEFEF}{\color[HTML]{FF0000} \textbf{0.724}} &
  \cellcolor[HTML]{EFEFEF}{\color[HTML]{FF0000} \textbf{0.570}} &
  \cellcolor[HTML]{EFEFEF}{\color[HTML]{FF0000} \textbf{0.698}} \\ \hline
 & GS       & 0.929          & 0.877          & \multicolumn{1}{c:}{0.918}          & \textbf{0.859} & \textbf{0.753} & \textbf{0.862}                        \\
 &
  SVS &
  {\color[HTML]{FF0000} \textbf{0.994}} &
  {\color[HTML]{FF0000} \textbf{0.988}} &
  \multicolumn{1}{c:}{{\color[HTML]{FF0000} \textbf{0.996}}} &
  0.778 &
  0.641 &
  0.784 \\
 & Saliency & 0.514          & 0.349          & \multicolumn{1}{c:}{0.453}          & 0.726          & 0.571          & 0.744                                 \\
 & FA       & \textbf{0.987} & \textbf{0.975} & \multicolumn{1}{c:}{\textbf{0.991}} & 0.732          & 0.586          & 0.752                                 \\
 & IG       & 0.929          & 0.877          & \multicolumn{1}{c:}{0.918}          & \color[HTML]{FF0000} \textbf{0.861}         & \color[HTML]{FF0000} \textbf{0.756}           & \color[HTML]{FF0000} \textbf{0.864}                                 \\
 & CE       & 0.596          & 0.427          & \multicolumn{1}{c:}{0.736}          & 0.496          & 0.333          & 0.600                                 \\
\multirow{-7}{*}{\rotatebox[origin=c]{90}{MLP-Mixer}}&
  \cellcolor[HTML]{EFEFEF}CE+QR(Ours) &
  \cellcolor[HTML]{EFEFEF}0.949 &
  \cellcolor[HTML]{EFEFEF}0.904 &
  \multicolumn{1}{c:}{\cellcolor[HTML]{EFEFEF}0.933} &
  \cellcolor[HTML]{EFEFEF}{0.709} &
  \cellcolor[HTML]{EFEFEF}{0.551} &
  \cellcolor[HTML]{EFEFEF}{0.660} \\ \hline
 & GS       & 0.861          & 0.763          & \multicolumn{1}{c:}{0.842}          & 0.685          & 0.527          & 0.690                                 \\
 &
  SVS &
  \textbf{0.883} &
  {\color[HTML]{000000} \textbf{0.803}} &
  \multicolumn{1}{c:}{{\color[HTML]{FF0000} \textbf{0.902}}} &
  {\color[HTML]{FF0000} \textbf{0.811}} &
  {\color[HTML]{FF0000} \textbf{0.686}} &
  {\color[HTML]{FF0000} \textbf{0.812}} \\
 & Saliency & 0.526          & 0.359          & \multicolumn{1}{c:}{0.484}          & \textbf{0.722} & \textbf{0.570} & \textbf{0.750}                        \\
 & FA       & 0.779          & 0.640          & \multicolumn{1}{c:}{0.809}          & 0.674          & 0.512          & 0.684                                 \\
 & IG       & 0.861          & 0.763          & \multicolumn{1}{c:}{0.842}          & 0.702          & 0.544          & 0.710                                 \\
 & CE       & 0.700          & 0.546          & \multicolumn{1}{c:}{0.796}          & 0.520          & 0.352          & 0.536                                 \\
\multirow{-7}{*}{\rotatebox[origin=c]{90}{ViT}} &
  \cellcolor[HTML]{EFEFEF}CE+QR(Ours) &
  \cellcolor[HTML]{EFEFEF}{\color[HTML]{FF0000} \textbf{0.925}} &
  \cellcolor[HTML]{EFEFEF}{\color[HTML]{FF0000} \textbf{0.863}} &
  \multicolumn{1}{c:}{\cellcolor[HTML]{EFEFEF}{\color[HTML]{000000} \textbf{0.898}}} &
  \cellcolor[HTML]{EFEFEF}0.531 &
  \cellcolor[HTML]{EFEFEF}0.363 &
  \cellcolor[HTML]{EFEFEF}0.572 \\ \hline
\end{tabular}}
    \caption{Performance Evaluation of CWRI Metrics. Here, the features identified as important by the model against the established ground truth importance is evaluated using binary metrics, including F1~score, Jaccard, and Accuracy (distinguished from model accuracy.)}
    \label{table_CWRI}
\vspace{-16pt}
\end{table}

\subsection{Additional Experiments}
Due to the space constraint, additional experimental results are presented in the appendix, with key highlights summarized below.

\subsubsection{Other Image Encoding Methods}
We provide several main experiment results with the Gramian Angular Field image encoding method in \cref{appendix:UseOfOtherImageEncodings}.

\subsubsection{Effect of $\lambda$}
We evaluated the effect of $\lambda$ -a key hyperparameter in our model which modulates the QR-Ortho Loss (\cref{eq:qr_ortho})-(ranging from 0 to 1) on two tasks: SquidGame and WhichFinger. Results indicate that increasing $\lambda$ improves CWRI-related metrics, but excessively high values can reduce model accuracy. Detailed findings are in \cref{appendix:lambdaeffect}.

\subsubsection{Alignment with Domain Knowledge}
Using Gilon and MS datasets, we demonstrated CAFO's alignment with established domain insights. For the Gilon task, accelerometer features were crucial for speed differentiation, and in the MS task, similar activities yielded similar attention scores. Visual evidence of these correlations is provided in \cref{appendix:alignmentwithdomain}. 

\subsection{Limitations and Discussions}
We discuss the following limitations of CAFO. As our evaluation strategy for the GI method inherits the ROAR method~\cite{hooker2019benchmark}, the retraining and re-evaluation cost is computationally intensive. Consequently, there is a need for alternative explanation methodologies that either do not rely on model accuracy or employ more computationally efficient evaluation techniques for the GI method. Additionally, our CAFO leverages image encoding to represent a time series into an image-like representation. While this approach has its merits, it also restricts the type of models used. As such, our research agenda includes the development of evaluation methods that are not only less demanding in terms of computational resources but also architecture-agnostic. 

\section{Conclusion}
In this paper, we introduce CAFO, a feature-centric explanation framework for MTS classification. An in-depth discussion regarding the feature-centric explanation for MTS has been missing in much of the previous literature despite its huge importance, due to the lack of evaluation protocols, pertinent benchmarks, and methodologies. Addressing these problems, our contribution is threefold: First, we present CAFO, a channel attention-based feature explainer which combines a novel depth-wise channel attention module, DepCA, with QR-Ortho regularization for feature explanation in time series. Second, we curate a collection of both real-world and synthetic datasets, each annotated with known discriminative feature importance. Third, we introduce a set of feature importance metrics designed to quantify both global and class-specific importance, complete with corresponding evaluation schemes. We believe that our work will serve as a new groundwork for understanding feature importance within MTS classification.

\begin{acks}
This work was supported by the National Research Foundation of Korea(NRF) grant funded by the Korea government(MSIT) (RS-2023-00277383), and Institute of Information \& communications Technology Planning \& Evaluation(IITP) grant funded by the Korea government(MSIT) (No.RS-2020-II201336, Artificial Intelligence graduate school support(UNIST)). The authors extend their gratitude to the Gilon Corporation for inspiring this work. Special thanks are due to Prof. Sunghoon Lim, Gyeongho Kim, Sujin Jeon, and Jae Gyeong Choi for providing the smart glove essential for the WhichFinger data collection and for their valuable insights into our research. We are also grateful to MyoungHoon Lee, Prof.~Suhyeon Kim, Wonho Sohn, and Hyewon Kang for their initial discussions that shaped our paper. Appreciation is further extended to Yeonjoo Kim, Solang Kim, Bosung Kim, Isu Jeong, Jaewook Lee, and Changhyeon Lee for their thorough review of our manuscript. Lastly, we thank the numerous anonymous reviewers whose constructive feedback significantly enhanced our work.
\end{acks}

\bibliographystyle{ACM-Reference-Format}
\bibliography{main}


\begin{thebibliography}{9}

%%% the Web address.

\ifx \showCODEN    \undefined \def \showCODEN     #1{\unskip}     \fi
\ifx \showDOI      \undefined \def \showDOI       #1{#1}\fi
\ifx \showISBNx    \undefined \def \showISBNx     #1{\unskip}     \fi
\ifx \showISBNxiii \undefined \def \showISBNxiii  #1{\unskip}     \fi
\ifx \showISSN     \undefined \def \showISSN      #1{\unskip}     \fi
\ifx \showLCCN     \undefined \def \showLCCN      #1{\unskip}     \fi
\ifx \shownote     \undefined \def \shownote      #1{#1}          \fi
\ifx \showarticletitle \undefined \def \showarticletitle #1{#1}   \fi
\ifx \showURL      \undefined \def \showURL       {\relax}        \fi
% invisible to TeX
\providecommand\bibfield[2]{#2}
\providecommand\bibinfo[2]{#2}
\providecommand\natexlab[1]{#1}
\providecommand\showeprint[2][]{arXiv:#2}

\bibitem[Es-Sakali et~al\mbox{.}(2022)]%
        {es2022review}
\bibfield{author}{\bibinfo{person}{Niima Es-Sakali}, \bibinfo{person}{Moha Cherkaoui}, \bibinfo{person}{Mohamed~Oualid Mghazli}, {and} \bibinfo{person}{Zakaria Naimi}.} \bibinfo{year}{2022}\natexlab{}.
\newblock \showarticletitle{Review of predictive maintenance algorithms applied to HVAC systems}.
\newblock \bibinfo{journal}{\emph{Energy Reports}}  \bibinfo{volume}{8} (\bibinfo{year}{2022}), \bibinfo{pages}{1003--1012}.
\newblock


\bibitem[Kokhlikyan et~al\mbox{.}(2020)]%
        {kokhlikyan2020captum}
\bibfield{author}{\bibinfo{person}{Narine Kokhlikyan}, \bibinfo{person}{Vivek Miglani}, \bibinfo{person}{Miguel Martin}, \bibinfo{person}{Edward Wang}, \bibinfo{person}{Bilal Alsallakh}, \bibinfo{person}{Jonathan Reynolds}, \bibinfo{person}{Alexander Melnikov}, \bibinfo{person}{Natalia Kliushkina}, \bibinfo{person}{Carlos Araya}, \bibinfo{person}{Siqi Yan}, {et~al\mbox{.}}} \bibinfo{year}{2020}\natexlab{}.
\newblock \showarticletitle{Captum: A unified and generic model interpretability library for pytorch}.
\newblock \bibinfo{journal}{\emph{arXiv preprint arXiv:2009.07896}} (\bibinfo{year}{2020}).
\newblock


\bibitem[Liu et~al\mbox{.}(2020)]%
        {liu2020data}
\bibfield{author}{\bibinfo{person}{Jiangyan Liu}, \bibinfo{person}{Daliang Shi}, \bibinfo{person}{Guannan Li}, \bibinfo{person}{Yi Xie}, \bibinfo{person}{Kuining Li}, \bibinfo{person}{Bin Liu}, {and} \bibinfo{person}{Zhipeng Ru}.} \bibinfo{year}{2020}\natexlab{}.
\newblock \showarticletitle{Data-driven and association rule mining-based fault diagnosis and action mechanism analysis for building chillers}.
\newblock \bibinfo{journal}{\emph{Energy and Buildings}}  \bibinfo{volume}{216} (\bibinfo{year}{2020}), \bibinfo{pages}{109957}.
\newblock


\bibitem[Loshchilov and Hutter(2017)]%
        {loshchilov2017decoupled}
\bibfield{author}{\bibinfo{person}{Ilya Loshchilov} {and} \bibinfo{person}{Frank Hutter}.} \bibinfo{year}{2017}\natexlab{}.
\newblock \showarticletitle{Decoupled weight decay regularization}.
\newblock \bibinfo{journal}{\emph{arXiv preprint arXiv:1711.05101}} (\bibinfo{year}{2017}).
\newblock


\bibitem[Namburu et~al\mbox{.}(2007)]%
        {namburu2007data}
\bibfield{author}{\bibinfo{person}{Setu~Madhavi Namburu}, \bibinfo{person}{Mohammad~S Azam}, \bibinfo{person}{Jianhui Luo}, \bibinfo{person}{Kihoon Choi}, {and} \bibinfo{person}{Krishna~R Pattipati}.} \bibinfo{year}{2007}\natexlab{}.
\newblock \showarticletitle{Data-driven modeling, fault diagnosis and optimal sensor selection for HVAC chillers}.
\newblock \bibinfo{journal}{\emph{IEEE transactions on automation science and engineering}} \bibinfo{volume}{4}, \bibinfo{number}{3} (\bibinfo{year}{2007}), \bibinfo{pages}{469--473}.
\newblock


\bibitem[Schein and Bushby(2006)]%
        {schein2006hierarchical}
\bibfield{author}{\bibinfo{person}{Jeffrey Schein} {and} \bibinfo{person}{Steven~T Bushby}.} \bibinfo{year}{2006}\natexlab{}.
\newblock \showarticletitle{A hierarchical rule-based fault detection and diagnostic method for HVAC systems}.
\newblock \bibinfo{journal}{\emph{Hvac\&r Research}} \bibinfo{volume}{12}, \bibinfo{number}{1} (\bibinfo{year}{2006}), \bibinfo{pages}{111--125}.
\newblock


\bibitem[Selvaraju et~al\mbox{.}(2017)]%
        {selvaraju2017grad}
\bibfield{author}{\bibinfo{person}{Ramprasaath~R Selvaraju}, \bibinfo{person}{Michael Cogswell}, \bibinfo{person}{Abhishek Das}, \bibinfo{person}{Ramakrishna Vedantam}, \bibinfo{person}{Devi Parikh}, {and} \bibinfo{person}{Dhruv Batra}.} \bibinfo{year}{2017}\natexlab{}.
\newblock \showarticletitle{Grad-cam: Visual explanations from deep networks via gradient-based localization}. In \bibinfo{booktitle}{\emph{Proceedings of the IEEE international conference on computer vision}}. \bibinfo{pages}{618--626}.
\newblock


\bibitem[Tang(2010)]%
        {tang2010hvac}
\bibfield{author}{\bibinfo{person}{Fan Tang}.} \bibinfo{year}{2010}\natexlab{}.
\newblock \emph{\bibinfo{title}{HVAC system modeling and optimization: a data-mining approach}}.
\newblock \bibinfo{thesistype}{Ph.\,D. Dissertation}. \bibinfo{school}{The University of Iowa}.
\newblock


\bibitem[Vaswani et~al\mbox{.}(2017)]%
        {vaswani2017attention}
\bibfield{author}{\bibinfo{person}{Ashish Vaswani}, \bibinfo{person}{Noam Shazeer}, \bibinfo{person}{Niki Parmar}, \bibinfo{person}{Jakob Uszkoreit}, \bibinfo{person}{Llion Jones}, \bibinfo{person}{Aidan~N Gomez}, \bibinfo{person}{{\L}ukasz Kaiser}, {and} \bibinfo{person}{Illia Polosukhin}.} \bibinfo{year}{2017}\natexlab{}.
\newblock \showarticletitle{Attention is all you need}.
\newblock \bibinfo{journal}{\emph{Advances in neural information processing systems}}  \bibinfo{volume}{30} (\bibinfo{year}{2017}).
\newblock


\end{thebibliography}



\begin{thebibliography}{52}

%%% the Web address.

\ifx \showCODEN    \undefined \def \showCODEN     #1{\unskip}     \fi
\ifx \showDOI      \undefined \def \showDOI       #1{#1}\fi
\ifx \showISBNx    \undefined \def \showISBNx     #1{\unskip}     \fi
\ifx \showISBNxiii \undefined \def \showISBNxiii  #1{\unskip}     \fi
\ifx \showISSN     \undefined \def \showISSN      #1{\unskip}     \fi
\ifx \showLCCN     \undefined \def \showLCCN      #1{\unskip}     \fi
\ifx \shownote     \undefined \def \shownote      #1{#1}          \fi
\ifx \showarticletitle \undefined \def \showarticletitle #1{#1}   \fi
\ifx \showURL      \undefined \def \showURL       {\relax}        \fi
% invisible to TeX
\providecommand\bibfield[2]{#2}
\providecommand\bibinfo[2]{#2}
\providecommand\natexlab[1]{#1}
\providecommand\showeprint[2][]{arXiv:#2}

\bibitem[Abdi(2007)]%
        {abdi2007kendall}
\bibfield{author}{\bibinfo{person}{Herv{\'e} Abdi}.} \bibinfo{year}{2007}\natexlab{}.
\newblock \showarticletitle{The Kendall rank correlation coefficient}.
\newblock \bibinfo{journal}{\emph{Encyclopedia of Measurement and Statistics. Sage, Thousand Oaks, CA}} (\bibinfo{year}{2007}), \bibinfo{pages}{508--510}.
\newblock


\bibitem[Bento et~al\mbox{.}(2021)]%
        {bento2021timeshap}
\bibfield{author}{\bibinfo{person}{Jo{\~a}o Bento}, \bibinfo{person}{Pedro Saleiro}, \bibinfo{person}{Andr{\'e}~F Cruz}, \bibinfo{person}{M{\'a}rio~AT Figueiredo}, {and} \bibinfo{person}{Pedro Bizarro}.} \bibinfo{year}{2021}\natexlab{}.
\newblock \showarticletitle{Timeshap: Explaining recurrent models through sequence perturbations}. In \bibinfo{booktitle}{\emph{Proceedings of the 27th ACM SIGKDD Conference on Knowledge Discovery \& Data Mining}}. \bibinfo{pages}{2565--2573}.
\newblock


\bibitem[Beuchert et~al\mbox{.}(2020)]%
        {beuchert2020overcoming}
\bibfield{author}{\bibinfo{person}{Jonas Beuchert}, \bibinfo{person}{Friedrich Solowjow}, \bibinfo{person}{Sebastian Trimpe}, {and} \bibinfo{person}{Thomas Seel}.} \bibinfo{year}{2020}\natexlab{}.
\newblock \showarticletitle{Overcoming bandwidth limitations in wireless sensor networks by exploitation of cyclic signal patterns: An event-triggered learning approach}.
\newblock \bibinfo{journal}{\emph{Sensors}} \bibinfo{volume}{20}, \bibinfo{number}{1} (\bibinfo{year}{2020}), \bibinfo{pages}{260}.
\newblock


\bibitem[Bj{\"o}rck(1994)]%
        {bjorck1994numerics}
\bibfield{author}{\bibinfo{person}{{\AA}ke Bj{\"o}rck}.} \bibinfo{year}{1994}\natexlab{}.
\newblock \showarticletitle{Numerics of gram-schmidt orthogonalization}.
\newblock \bibinfo{journal}{\emph{Linear Algebra and Its Applications}}  \bibinfo{volume}{197} (\bibinfo{year}{1994}), \bibinfo{pages}{297--316}.
\newblock


\bibitem[Cali{\'n}ski and Harabasz(1974)]%
        {calinski1974dendrite}
\bibfield{author}{\bibinfo{person}{Tadeusz Cali{\'n}ski} {and} \bibinfo{person}{Jerzy Harabasz}.} \bibinfo{year}{1974}\natexlab{}.
\newblock \showarticletitle{A dendrite method for cluster analysis}.
\newblock \bibinfo{journal}{\emph{Communications in Statistics-theory and Methods}} \bibinfo{volume}{3}, \bibinfo{number}{1} (\bibinfo{year}{1974}), \bibinfo{pages}{1--27}.
\newblock


\bibitem[Castro et~al\mbox{.}(2009)]%
        {castro2009polynomial}
\bibfield{author}{\bibinfo{person}{Javier Castro}, \bibinfo{person}{Daniel G{\'o}mez}, {and} \bibinfo{person}{Juan Tejada}.} \bibinfo{year}{2009}\natexlab{}.
\newblock \showarticletitle{Polynomial calculation of the Shapley value based on sampling}.
\newblock \bibinfo{journal}{\emph{Computers \& Operations Research}} \bibinfo{volume}{36}, \bibinfo{number}{5} (\bibinfo{year}{2009}), \bibinfo{pages}{1726--1730}.
\newblock


\bibitem[Crabb{\'e} and Van Der~Schaar(2021)]%
        {crabbe2021explaining}
\bibfield{author}{\bibinfo{person}{Jonathan Crabb{\'e}} {and} \bibinfo{person}{Mihaela Van Der~Schaar}.} \bibinfo{year}{2021}\natexlab{}.
\newblock \showarticletitle{Explaining time series predictions with dynamic masks}. In \bibinfo{booktitle}{\emph{International Conference on Machine Learning}}. PMLR, \bibinfo{pages}{2166--2177}.
\newblock


\bibitem[Dosovitskiy et~al\mbox{.}(2020)]%
        {dosovitskiy2020image}
\bibfield{author}{\bibinfo{person}{Alexey Dosovitskiy}, \bibinfo{person}{Lucas Beyer}, \bibinfo{person}{Alexander Kolesnikov}, \bibinfo{person}{Dirk Weissenborn}, \bibinfo{person}{Xiaohua Zhai}, \bibinfo{person}{Thomas Unterthiner}, \bibinfo{person}{Mostafa Dehghani}, \bibinfo{person}{Matthias Minderer}, \bibinfo{person}{Georg Heigold}, \bibinfo{person}{Sylvain Gelly}, {et~al\mbox{.}}} \bibinfo{year}{2020}\natexlab{}.
\newblock \showarticletitle{An image is worth 16x16 words: Transformers for image recognition at scale}.
\newblock \bibinfo{journal}{\emph{arXiv preprint arXiv:2010.11929}} (\bibinfo{year}{2020}).
\newblock


\bibitem[Eckmann et~al\mbox{.}(1995)]%
        {eckmann1995recurrence}
\bibfield{author}{\bibinfo{person}{Jean-Pierre Eckmann}, \bibinfo{person}{S~Oliffson Kamphorst}, \bibinfo{person}{David Ruelle}, {et~al\mbox{.}}} \bibinfo{year}{1995}\natexlab{}.
\newblock \showarticletitle{Recurrence plots of dynamical systems}.
\newblock \bibinfo{journal}{\emph{World Scientific Series on Nonlinear Science Series A}}  \bibinfo{volume}{16} (\bibinfo{year}{1995}), \bibinfo{pages}{441--446}.
\newblock


\bibitem[Filonov et~al\mbox{.}(2016)]%
        {filonov2016multivariate}
\bibfield{author}{\bibinfo{person}{Pavel Filonov}, \bibinfo{person}{Andrey Lavrentyev}, {and} \bibinfo{person}{Artem Vorontsov}.} \bibinfo{year}{2016}\natexlab{}.
\newblock \showarticletitle{Multivariate industrial time series with cyber-attack simulation: Fault detection using an lstm-based predictive data model}.
\newblock \bibinfo{journal}{\emph{arXiv preprint arXiv:1612.06676}} (\bibinfo{year}{2016}).
\newblock


\bibitem[Goodall(1993)]%
        {goodall199313}
\bibfield{author}{\bibinfo{person}{Colin~R Goodall}.} \bibinfo{year}{1993}\natexlab{}.
\newblock \showarticletitle{13 Computation using the QR decomposition}.
\newblock  (\bibinfo{year}{1993}).
\newblock


\bibitem[He et~al\mbox{.}(2021)]%
        {he2021wavelet}
\bibfield{author}{\bibinfo{person}{Jianing He}, \bibinfo{person}{Xiaolong Gong}, {and} \bibinfo{person}{Linpeng Huang}.} \bibinfo{year}{2021}\natexlab{}.
\newblock \showarticletitle{Wavelet-temporal neural network for multivariate time series prediction}. In \bibinfo{booktitle}{\emph{2021 International Joint Conference on Neural Networks (IJCNN)}}. IEEE, \bibinfo{pages}{1--8}.
\newblock


\bibitem[He et~al\mbox{.}(2016)]%
        {he2016deep}
\bibfield{author}{\bibinfo{person}{Kaiming He}, \bibinfo{person}{Xiangyu Zhang}, \bibinfo{person}{Shaoqing Ren}, {and} \bibinfo{person}{Jian Sun}.} \bibinfo{year}{2016}\natexlab{}.
\newblock \showarticletitle{Deep residual learning for image recognition}. In \bibinfo{booktitle}{\emph{Proceedings of the IEEE conference on computer vision and pattern recognition}}. \bibinfo{pages}{770--778}.
\newblock


\bibitem[Hochreiter(1998)]%
        {hochreiter1998vanishing}
\bibfield{author}{\bibinfo{person}{Sepp Hochreiter}.} \bibinfo{year}{1998}\natexlab{}.
\newblock \showarticletitle{The vanishing gradient problem during learning recurrent neural nets and problem solutions}.
\newblock \bibinfo{journal}{\emph{International Journal of Uncertainty, Fuzziness and Knowledge-Based Systems}} \bibinfo{volume}{6}, \bibinfo{number}{02} (\bibinfo{year}{1998}), \bibinfo{pages}{107--116}.
\newblock


\bibitem[Hochreiter and Schmidhuber(1997)]%
        {hochreiter1997long}
\bibfield{author}{\bibinfo{person}{Sepp Hochreiter} {and} \bibinfo{person}{J{\"u}rgen Schmidhuber}.} \bibinfo{year}{1997}\natexlab{}.
\newblock \showarticletitle{Long short-term memory}.
\newblock \bibinfo{journal}{\emph{Neural computation}} \bibinfo{volume}{9}, \bibinfo{number}{8} (\bibinfo{year}{1997}), \bibinfo{pages}{1735--1780}.
\newblock


\bibitem[Hooker et~al\mbox{.}(2019)]%
        {hooker2019benchmark}
\bibfield{author}{\bibinfo{person}{Sara Hooker}, \bibinfo{person}{Dumitru Erhan}, \bibinfo{person}{Pieter-Jan Kindermans}, {and} \bibinfo{person}{Been Kim}.} \bibinfo{year}{2019}\natexlab{}.
\newblock \showarticletitle{A benchmark for interpretability methods in deep neural networks}.
\newblock \bibinfo{journal}{\emph{Advances in neural information processing systems}}  \bibinfo{volume}{32} (\bibinfo{year}{2019}).
\newblock


\bibitem[Hsieh et~al\mbox{.}(2021)]%
        {hsieh2021explainable}
\bibfield{author}{\bibinfo{person}{Tsung-Yu Hsieh}, \bibinfo{person}{Suhang Wang}, \bibinfo{person}{Yiwei Sun}, {and} \bibinfo{person}{Vasant Honavar}.} \bibinfo{year}{2021}\natexlab{}.
\newblock \showarticletitle{Explainable multivariate time series classification: a deep neural network which learns to attend to important variables as well as time intervals}. In \bibinfo{booktitle}{\emph{Proceedings of the 14th ACM international conference on web search and data mining}}. \bibinfo{pages}{607--615}.
\newblock


\bibitem[Hu et~al\mbox{.}(2018)]%
        {hu2018squeeze}
\bibfield{author}{\bibinfo{person}{Jie Hu}, \bibinfo{person}{Li Shen}, {and} \bibinfo{person}{Gang Sun}.} \bibinfo{year}{2018}\natexlab{}.
\newblock \showarticletitle{Squeeze-and-excitation networks}. In \bibinfo{booktitle}{\emph{Proceedings of the IEEE conference on computer vision and pattern recognition}}. \bibinfo{pages}{7132--7141}.
\newblock


\bibitem[Ismail et~al\mbox{.}(2020)]%
        {ismail2020benchmarking}
\bibfield{author}{\bibinfo{person}{Aya~Abdelsalam Ismail}, \bibinfo{person}{Mohamed Gunady}, \bibinfo{person}{Hector Corrada~Bravo}, {and} \bibinfo{person}{Soheil Feizi}.} \bibinfo{year}{2020}\natexlab{}.
\newblock \showarticletitle{Benchmarking deep learning interpretability in time series predictions}.
\newblock \bibinfo{journal}{\emph{Advances in neural information processing systems}}  \bibinfo{volume}{33} (\bibinfo{year}{2020}), \bibinfo{pages}{6441--6452}.
\newblock


\bibitem[Ismail et~al\mbox{.}(2019)]%
        {ismail2019input}
\bibfield{author}{\bibinfo{person}{Aya~Abdelsalam Ismail}, \bibinfo{person}{Mohamed Gunady}, \bibinfo{person}{Luiz Pessoa}, \bibinfo{person}{Hector Corrada~Bravo}, {and} \bibinfo{person}{Soheil Feizi}.} \bibinfo{year}{2019}\natexlab{}.
\newblock \showarticletitle{Input-cell attention reduces vanishing saliency of recurrent neural networks}.
\newblock \bibinfo{journal}{\emph{Advances in Neural Information Processing Systems}}  \bibinfo{volume}{32} (\bibinfo{year}{2019}).
\newblock


\bibitem[Jiang et~al\mbox{.}(2022)]%
        {jiang2022fecam}
\bibfield{author}{\bibinfo{person}{Maowei Jiang}, \bibinfo{person}{Pengyu Zeng}, \bibinfo{person}{Kai Wang}, \bibinfo{person}{Huan Liu}, \bibinfo{person}{Wenbo Chen}, {and} \bibinfo{person}{Haoran Liu}.} \bibinfo{year}{2022}\natexlab{}.
\newblock \showarticletitle{FECAM: Frequency Enhanced Channel Attention Mechanism for Time Series Forecasting}.
\newblock \bibinfo{journal}{\emph{arXiv preprint arXiv:2212.01209}} (\bibinfo{year}{2022}).
\newblock


\bibitem[Kim et~al\mbox{.}(2023)]%
        {kim2023multi}
\bibfield{author}{\bibinfo{person}{Jaeho Kim}, \bibinfo{person}{Hyewon Kang}, \bibinfo{person}{Jaewan Yang}, \bibinfo{person}{Haneul Jung}, \bibinfo{person}{Seulki Lee}, {and} \bibinfo{person}{Junghye Lee}.} \bibinfo{year}{2023}\natexlab{}.
\newblock \showarticletitle{Multi-task Deep Learning for Human Activity, Speed, and Body Weight Estimation using Commercial Smart Insoles}.
\newblock \bibinfo{journal}{\emph{IEEE Internet of Things Journal}} (\bibinfo{year}{2023}).
\newblock


\bibitem[Lea et~al\mbox{.}(2017)]%
        {lea2017temporal}
\bibfield{author}{\bibinfo{person}{Colin Lea}, \bibinfo{person}{Michael~D Flynn}, \bibinfo{person}{Rene Vidal}, \bibinfo{person}{Austin Reiter}, {and} \bibinfo{person}{Gregory~D Hager}.} \bibinfo{year}{2017}\natexlab{}.
\newblock \showarticletitle{Temporal convolutional networks for action segmentation and detection}. In \bibinfo{booktitle}{\emph{proceedings of the IEEE Conference on Computer Vision and Pattern Recognition}}. \bibinfo{pages}{156--165}.
\newblock


\bibitem[Lee et~al\mbox{.}(2021)]%
        {lee2021hierarchical}
\bibfield{author}{\bibinfo{person}{Jiyoon Lee}, \bibinfo{person}{Hyungrok Do}, \bibinfo{person}{Mingu Kwak}, \bibinfo{person}{Hyungu Kahng}, {and} \bibinfo{person}{Seoung~Bum Kim}.} \bibinfo{year}{2021}\natexlab{}.
\newblock \showarticletitle{Hierarchical segment-channel attention network for explainable multichannel signal classification}.
\newblock \bibinfo{journal}{\emph{Information Sciences}}  \bibinfo{volume}{567} (\bibinfo{year}{2021}), \bibinfo{pages}{312--331}.
\newblock


\bibitem[Lee and Bae(2020)]%
        {lee2020deep}
\bibfield{author}{\bibinfo{person}{Minhyuk Lee} {and} \bibinfo{person}{Joonbum Bae}.} \bibinfo{year}{2020}\natexlab{}.
\newblock \showarticletitle{Deep learning based real-time recognition of dynamic finger gestures using a data glove}.
\newblock \bibinfo{journal}{\emph{IEEE Access}}  \bibinfo{volume}{8} (\bibinfo{year}{2020}), \bibinfo{pages}{219923--219933}.
\newblock


\bibitem[Lundberg and Lee(2017)]%
        {lundberg2017unified}
\bibfield{author}{\bibinfo{person}{Scott~M Lundberg} {and} \bibinfo{person}{Su-In Lee}.} \bibinfo{year}{2017}\natexlab{}.
\newblock \showarticletitle{A unified approach to interpreting model predictions}.
\newblock \bibinfo{journal}{\emph{Advances in neural information processing systems}}  \bibinfo{volume}{30} (\bibinfo{year}{2017}).
\newblock


\bibitem[Maat et~al\mbox{.}(2017)]%
        {maat2017timesynth}
\bibfield{author}{\bibinfo{person}{JR Maat}, \bibinfo{person}{A Malali}, {and} \bibinfo{person}{P Protopapas}.} \bibinfo{year}{2017}\natexlab{}.
\newblock \bibinfo{title}{Timesynth: A multipurpose library for synthetic time series in python}.
\newblock
\newblock


\bibitem[Morris et~al\mbox{.}(2014)]%
        {morris2014recofit}
\bibfield{author}{\bibinfo{person}{Dan Morris}, \bibinfo{person}{T~Scott Saponas}, \bibinfo{person}{Andrew Guillory}, {and} \bibinfo{person}{Ilya Kelner}.} \bibinfo{year}{2014}\natexlab{}.
\newblock \showarticletitle{RecoFit: using a wearable sensor to find, recognize, and count repetitive exercises}. In \bibinfo{booktitle}{\emph{Proceedings of the SIGCHI Conference on Human Factors in Computing Systems}}. \bibinfo{pages}{3225--3234}.
\newblock


\bibitem[Myers and Sirois(2004)]%
        {myers2004spearman}
\bibfield{author}{\bibinfo{person}{Leann Myers} {and} \bibinfo{person}{Maria~J Sirois}.} \bibinfo{year}{2004}\natexlab{}.
\newblock \showarticletitle{Spearman correlation coefficients, differences between}.
\newblock \bibinfo{journal}{\emph{Encyclopedia of statistical sciences}}  \bibinfo{volume}{12} (\bibinfo{year}{2004}).
\newblock


\bibitem[Park et~al\mbox{.}(2018)]%
        {park2018bam}
\bibfield{author}{\bibinfo{person}{Jongchan Park}, \bibinfo{person}{Sanghyun Woo}, \bibinfo{person}{Joon-Young Lee}, {and} \bibinfo{person}{In~So Kweon}.} \bibinfo{year}{2018}\natexlab{}.
\newblock \showarticletitle{Bam: Bottleneck attention module}.
\newblock \bibinfo{journal}{\emph{arXiv preprint arXiv:1807.06514}} (\bibinfo{year}{2018}).
\newblock


\bibitem[Ribeiro et~al\mbox{.}(2016)]%
        {ribeiro2016should}
\bibfield{author}{\bibinfo{person}{Marco~Tulio Ribeiro}, \bibinfo{person}{Sameer Singh}, {and} \bibinfo{person}{Carlos Guestrin}.} \bibinfo{year}{2016}\natexlab{}.
\newblock \showarticletitle{" Why should i trust you?" Explaining the predictions of any classifier}. In \bibinfo{booktitle}{\emph{Proceedings of the 22nd ACM SIGKDD international conference on knowledge discovery and data mining}}. \bibinfo{pages}{1135--1144}.
\newblock


\bibitem[Schlegel et~al\mbox{.}(2019)]%
        {schlegel2019towards}
\bibfield{author}{\bibinfo{person}{Udo Schlegel}, \bibinfo{person}{Hiba Arnout}, \bibinfo{person}{Mennatallah El-Assady}, \bibinfo{person}{Daniela Oelke}, {and} \bibinfo{person}{Daniel~A Keim}.} \bibinfo{year}{2019}\natexlab{}.
\newblock \showarticletitle{Towards a rigorous evaluation of XAI methods on time series}. In \bibinfo{booktitle}{\emph{2019 IEEE/CVF International Conference on Computer Vision Workshop (ICCVW)}}. IEEE, \bibinfo{pages}{4197--4201}.
\newblock


\bibitem[Schreiber and Van~Loan(1989)]%
        {schreiber1989storage}
\bibfield{author}{\bibinfo{person}{Robert Schreiber} {and} \bibinfo{person}{Charles Van~Loan}.} \bibinfo{year}{1989}\natexlab{}.
\newblock \showarticletitle{A storage-efficient WY representation for products of Householder transformations}.
\newblock \bibinfo{journal}{\emph{SIAM J. Sci. Statist. Comput.}} \bibinfo{volume}{10}, \bibinfo{number}{1} (\bibinfo{year}{1989}), \bibinfo{pages}{53--57}.
\newblock


\bibitem[Siddiqui et~al\mbox{.}(2019)]%
        {siddiqui2019tsviz}
\bibfield{author}{\bibinfo{person}{Shoaib~Ahmed Siddiqui}, \bibinfo{person}{Dominique Mercier}, \bibinfo{person}{Mohsin Munir}, \bibinfo{person}{Andreas Dengel}, {and} \bibinfo{person}{Sheraz Ahmed}.} \bibinfo{year}{2019}\natexlab{}.
\newblock \showarticletitle{Tsviz: Demystification of deep learning models for time-series analysis}.
\newblock \bibinfo{journal}{\emph{IEEE Access}}  \bibinfo{volume}{7} (\bibinfo{year}{2019}), \bibinfo{pages}{67027--67040}.
\newblock


\bibitem[Simonyan et~al\mbox{.}(2013)]%
        {simonyan2013deep}
\bibfield{author}{\bibinfo{person}{Karen Simonyan}, \bibinfo{person}{Andrea Vedaldi}, {and} \bibinfo{person}{Andrew Zisserman}.} \bibinfo{year}{2013}\natexlab{}.
\newblock \showarticletitle{Deep inside convolutional networks: Visualising image classification models and saliency maps}.
\newblock \bibinfo{journal}{\emph{arXiv preprint arXiv:1312.6034}} (\bibinfo{year}{2013}).
\newblock


\bibitem[Sundararajan et~al\mbox{.}(2017)]%
        {sundararajan2017axiomatic}
\bibfield{author}{\bibinfo{person}{Mukund Sundararajan}, \bibinfo{person}{Ankur Taly}, {and} \bibinfo{person}{Qiqi Yan}.} \bibinfo{year}{2017}\natexlab{}.
\newblock \showarticletitle{Axiomatic attribution for deep networks}. In \bibinfo{booktitle}{\emph{International conference on machine learning}}. PMLR, \bibinfo{pages}{3319--3328}.
\newblock


\bibitem[Suresh et~al\mbox{.}(2017)]%
        {suresh2017clinical}
\bibfield{author}{\bibinfo{person}{Harini Suresh}, \bibinfo{person}{Nathan Hunt}, \bibinfo{person}{Alistair Johnson}, \bibinfo{person}{Leo~Anthony Celi}, \bibinfo{person}{Peter Szolovits}, {and} \bibinfo{person}{Marzyeh Ghassemi}.} \bibinfo{year}{2017}\natexlab{}.
\newblock \showarticletitle{Clinical intervention prediction and understanding using deep networks}.
\newblock \bibinfo{journal}{\emph{arXiv preprint arXiv:1705.08498}} (\bibinfo{year}{2017}).
\newblock


\bibitem[Taheri et~al\mbox{.}(2021)]%
        {taheri2021fault}
\bibfield{author}{\bibinfo{person}{Saman Taheri}, \bibinfo{person}{Amirhossein Ahmadi}, \bibinfo{person}{Behnam Mohammadi-Ivatloo}, {and} \bibinfo{person}{Somayeh Asadi}.} \bibinfo{year}{2021}\natexlab{}.
\newblock \showarticletitle{Fault detection diagnostic for HVAC systems via deep learning algorithms}.
\newblock \bibinfo{journal}{\emph{Energy and Buildings}}  \bibinfo{volume}{250} (\bibinfo{year}{2021}), \bibinfo{pages}{111275}.
\newblock


\bibitem[Tolstikhin et~al\mbox{.}(2021)]%
        {tolstikhin2021mlp}
\bibfield{author}{\bibinfo{person}{Ilya~O Tolstikhin}, \bibinfo{person}{Neil Houlsby}, \bibinfo{person}{Alexander Kolesnikov}, \bibinfo{person}{Lucas Beyer}, \bibinfo{person}{Xiaohua Zhai}, \bibinfo{person}{Thomas Unterthiner}, \bibinfo{person}{Jessica Yung}, \bibinfo{person}{Andreas Steiner}, \bibinfo{person}{Daniel Keysers}, \bibinfo{person}{Jakob Uszkoreit}, {et~al\mbox{.}}} \bibinfo{year}{2021}\natexlab{}.
\newblock \showarticletitle{Mlp-mixer: An all-mlp architecture for vision}.
\newblock \bibinfo{journal}{\emph{Advances in neural information processing systems}}  \bibinfo{volume}{34} (\bibinfo{year}{2021}), \bibinfo{pages}{24261--24272}.
\newblock


\bibitem[Tonekaboni et~al\mbox{.}(2020)]%
        {tonekaboni2020went}
\bibfield{author}{\bibinfo{person}{Sana Tonekaboni}, \bibinfo{person}{Shalmali Joshi}, \bibinfo{person}{Kieran Campbell}, \bibinfo{person}{David~K Duvenaud}, {and} \bibinfo{person}{Anna Goldenberg}.} \bibinfo{year}{2020}\natexlab{}.
\newblock \showarticletitle{What went wrong and when? Instance-wise feature importance for time-series black-box models}.
\newblock \bibinfo{journal}{\emph{Advances in Neural Information Processing Systems}}  \bibinfo{volume}{33} (\bibinfo{year}{2020}), \bibinfo{pages}{799--809}.
\newblock


\bibitem[Turb{\'e} et~al\mbox{.}(2023)]%
        {turbe2023evaluation}
\bibfield{author}{\bibinfo{person}{Hugues Turb{\'e}}, \bibinfo{person}{Mina Bjelogrlic}, \bibinfo{person}{Christian Lovis}, {and} \bibinfo{person}{Gianmarco Mengaldo}.} \bibinfo{year}{2023}\natexlab{}.
\newblock \showarticletitle{Evaluation of post-hoc interpretability methods in time-series classification}.
\newblock \bibinfo{journal}{\emph{Nature Machine Intelligence}} \bibinfo{volume}{5}, \bibinfo{number}{3} (\bibinfo{year}{2023}), \bibinfo{pages}{250--260}.
\newblock


\bibitem[Van~der Maaten and Hinton(2008)]%
        {van2008visualizing}
\bibfield{author}{\bibinfo{person}{Laurens Van~der Maaten} {and} \bibinfo{person}{Geoffrey Hinton}.} \bibinfo{year}{2008}\natexlab{}.
\newblock \showarticletitle{Visualizing data using t-SNE.}
\newblock \bibinfo{journal}{\emph{Journal of machine learning research}} \bibinfo{volume}{9}, \bibinfo{number}{11} (\bibinfo{year}{2008}).
\newblock


\bibitem[Vaswani et~al\mbox{.}(2017)]%
        {vaswani2017attention}
\bibfield{author}{\bibinfo{person}{Ashish Vaswani}, \bibinfo{person}{Noam Shazeer}, \bibinfo{person}{Niki Parmar}, \bibinfo{person}{Jakob Uszkoreit}, \bibinfo{person}{Llion Jones}, \bibinfo{person}{Aidan~N Gomez}, \bibinfo{person}{{\L}ukasz Kaiser}, {and} \bibinfo{person}{Illia Polosukhin}.} \bibinfo{year}{2017}\natexlab{}.
\newblock \showarticletitle{Attention is all you need}.
\newblock \bibinfo{journal}{\emph{Advances in neural information processing systems}}  \bibinfo{volume}{30} (\bibinfo{year}{2017}).
\newblock


\bibitem[Wang et~al\mbox{.}(2015)]%
        {wang2015encoding}
\bibfield{author}{\bibinfo{person}{Zhiguang Wang}, \bibinfo{person}{Tim Oates}, {et~al\mbox{.}}} \bibinfo{year}{2015}\natexlab{}.
\newblock \showarticletitle{Encoding time series as images for visual inspection and classification using tiled convolutional neural networks}. In \bibinfo{booktitle}{\emph{Workshops at the twenty-ninth AAAI conference on artificial intelligence}}, Vol.~\bibinfo{volume}{1}. AAAI Menlo Park, CA, USA.
\newblock


\bibitem[West et~al\mbox{.}(2011)]%
        {west2011automated}
\bibfield{author}{\bibinfo{person}{Samuel~R West}, \bibinfo{person}{Ying Guo}, \bibinfo{person}{X~Rosalind Wang}, {and} \bibinfo{person}{Joshua Wall}.} \bibinfo{year}{2011}\natexlab{}.
\newblock \showarticletitle{Automated fault detection and diagnosis of HVAC subsystems using statistical machine learning}.
\newblock  (\bibinfo{year}{2011}).
\newblock


\bibitem[Woo et~al\mbox{.}(2018)]%
        {woo2018cbam}
\bibfield{author}{\bibinfo{person}{Sanghyun Woo}, \bibinfo{person}{Jongchan Park}, \bibinfo{person}{Joon-Young Lee}, {and} \bibinfo{person}{In~So Kweon}.} \bibinfo{year}{2018}\natexlab{}.
\newblock \showarticletitle{Cbam: Convolutional block attention module}. In \bibinfo{booktitle}{\emph{Proceedings of the European conference on computer vision (ECCV)}}. \bibinfo{pages}{3--19}.
\newblock


\bibitem[Xu et~al\mbox{.}(2019)]%
        {xu2019explainable}
\bibfield{author}{\bibinfo{person}{Feiyu Xu}, \bibinfo{person}{Hans Uszkoreit}, \bibinfo{person}{Yangzhou Du}, \bibinfo{person}{Wei Fan}, \bibinfo{person}{Dongyan Zhao}, {and} \bibinfo{person}{Jun Zhu}.} \bibinfo{year}{2019}\natexlab{}.
\newblock \showarticletitle{Explainable AI: A brief survey on history, research areas, approaches and challenges}. In \bibinfo{booktitle}{\emph{Natural Language Processing and Chinese Computing: 8th CCF International Conference, NLPCC 2019, Dunhuang, China, October 9--14, 2019, Proceedings, Part II 8}}. Springer, \bibinfo{pages}{563--574}.
\newblock


\bibitem[Yang et~al\mbox{.}(2021)]%
        {yang2021simam}
\bibfield{author}{\bibinfo{person}{Lingxiao Yang}, \bibinfo{person}{Ru-Yuan Zhang}, \bibinfo{person}{Lida Li}, {and} \bibinfo{person}{Xiaohua Xie}.} \bibinfo{year}{2021}\natexlab{}.
\newblock \showarticletitle{Simam: A simple, parameter-free attention module for convolutional neural networks}. In \bibinfo{booktitle}{\emph{International conference on machine learning}}. PMLR, \bibinfo{pages}{11863--11874}.
\newblock


\bibitem[Yeh et~al\mbox{.}(2002)]%
        {yeh2002using}
\bibfield{author}{\bibinfo{person}{Shi-Tao Yeh} {et~al\mbox{.}}} \bibinfo{year}{2002}\natexlab{}.
\newblock \showarticletitle{Using trapezoidal rule for the area under a curve calculation}.
\newblock \bibinfo{journal}{\emph{Proceedings of the 27th Annual SAS{\textregistered} User Group International (SUGI’02)}} (\bibinfo{year}{2002}), \bibinfo{pages}{1--5}.
\newblock


\bibitem[Zhang and Wen(2019)]%
        {zhang2019systematic}
\bibfield{author}{\bibinfo{person}{Liang Zhang} {and} \bibinfo{person}{Jin Wen}.} \bibinfo{year}{2019}\natexlab{}.
\newblock \showarticletitle{A systematic feature selection procedure for short-term data-driven building energy forecasting model development}.
\newblock \bibinfo{journal}{\emph{Energy and Buildings}}  \bibinfo{volume}{183} (\bibinfo{year}{2019}), \bibinfo{pages}{428--442}.
\newblock


\bibitem[Zhang et~al\mbox{.}(2018)]%
        {zhang2018shufflenet}
\bibfield{author}{\bibinfo{person}{Xiangyu Zhang}, \bibinfo{person}{Xinyu Zhou}, \bibinfo{person}{Mengxiao Lin}, {and} \bibinfo{person}{Jian Sun}.} \bibinfo{year}{2018}\natexlab{}.
\newblock \showarticletitle{Shufflenet: An extremely efficient convolutional neural network for mobile devices}. In \bibinfo{booktitle}{\emph{Proceedings of the IEEE conference on computer vision and pattern recognition}}. \bibinfo{pages}{6848--6856}.
\newblock


\bibitem[Zhao et~al\mbox{.}(2020)]%
        {zhao2020deep}
\bibfield{author}{\bibinfo{person}{Zhibin Zhao}, \bibinfo{person}{Tianfu Li}, \bibinfo{person}{Jingyao Wu}, \bibinfo{person}{Chuang Sun}, \bibinfo{person}{Shibin Wang}, \bibinfo{person}{Ruqiang Yan}, {and} \bibinfo{person}{Xuefeng Chen}.} \bibinfo{year}{2020}\natexlab{}.
\newblock \showarticletitle{Deep learning algorithms for rotating machinery intelligent diagnosis: An open source benchmark study}.
\newblock \bibinfo{journal}{\emph{ISA transactions}}  \bibinfo{volume}{107} (\bibinfo{year}{2020}), \bibinfo{pages}{224--255}.
\newblock


\end{thebibliography}
% \appendix
% \input{Appendix/camera_ready_appendix}

\newpage
\onecolumn
\appendix
\begin{bibunit}[ACM-Reference-Format]

\section{Motivation Regarding Feature-Based Importance}
\label{appendix:motivation}
This section provides motivating examples to further illustrate the benefits of the proposed GI and CWRI measures in the MTS context and how they can be utilized in the real-world scenarios, especially for both engineers and domain experts. 

\subsection{Case Studies}
\textbf{Case Study 1: Optimization of Fault Detection and Diagnosis (FDD) in Heating, Ventilation, and Air Conditioning (HVAC) Systems} is an important issue in energy preservation as it can prevent excessive energy consumption in buildings \cite{liu2020data,namburu2007data}. Preemptive maintenance in HVAC systems aims to predict potential equipment failure, using data from Internet of Things (IoT) sensors embedded within the system \cite{es2022review}. Despite the integration of numerous sensors and controllers to monitor a variety of system faults, fault detection in HVAC systems remains a complex task due to the non-linearity of fault patterns and strong feature correlations. The situation is further complicated by the hundreds of sensors indicating different types of defects across varying operational modes, such as air conditioning and heating \cite{schein2006hierarchical,tang2010hvac}. Thus, identifying the sensors most contributory to a specific fault can aid maintenance specialists in accurately diagnosing the malfunction. Prioritizing sensors capable of diagnosing a broad spectrum of faults can enhance system efficiency. This comprehension of the global and class-specific importance of features provides invaluable insights into sensor design and placement, beneficial for manufacturers. Armed with these insights, domain experts can elucidate the role of each feature in fault detection, aiding engineers in the development of more effective fault detection models.

\noindent\textbf{Case Study 2. The Smart insole} contains numerous sensors, such as an accelerometer, force-sensitive resistor (FSR), and gyroscope, to capture various gait characteristics. The data acquired from these sensors serve to monitor a range of human activities. Notably, in the process of recognizing these activities, some sensors may prove multi-functional across several classes, while others may be specifically critical for a distinct task. For example, an accelerometer may be instrumental in detecting varying speed ranges of movement, whereas an FSR sensor could effectively discern the user's posture. Yet, how can we ascertain that the black-box model is appropriately employing these features? In this scenario, the Global Importance (GI) and Class-Wise Relevant Importance (CWRI), calculated from CAFO, offer invaluable insights into the model's sensor prioritization. While GI denotes the sensor that contributes most significantly to enhanced classification accuracy, a high GI score for an accelerometer, for instance, does not necessarily confer high importance to each class (e.g., Squat, Lunge). CWRI, on the other hand, provides class-specific insights into feature prioritization.

Understanding the role of each sensor affords substantial benefits for both engineers and practical applications. Engineers, for instance, can harness this information to implement feature selection and extraction strategies for predictive models. Moreover, manufacturers can leverage this knowledge to design more efficient and cost-effective smart insoles, by eliminating redundant sensors in the production phase, thus lowering both production and data transfer costs for cloud processing.

\section{Image Encoding Methods}
\label{appendix:UseOfOtherImageEncodings}
\subsection{Recurrence Plot}
A univariate time series $\mathbf{x}=[x_1,...,x_T]^\top$ is converted to a two-dimensional recurrence plot (RP) image by composing $L\triangleq T-(m-1)\tau$ number of new vectors $\mathbf{v}_1,..., \mathbf{v}_L$, where $\mathbf{v}_k$ is a vector consisting of raw time series as in \cref{rp_equation}. Here $\tau$ is the time delay, and $m$ is the embedding dimension selected as a hyperparameter. 
\begin{equation}
\label{rp_equation}
    \mathbf{v}_k = [x_k,x_{k+\tau},x_{k+2\tau},...,x_{k+(m-1)\tau}]^\top
\end{equation}
These vectors are finally transformed into an RP image by measuring the pairwise distance between all vectors $\mathbf{v}_1,...,\mathbf{v}_L$, leading to an $L\times L$ image. Each element of the RP image is defined in \cref{heaviside}. The threshold distance $\varepsilon$, Heaviside step function $\mathbb{H}$, and a norm function $||\cdot||$ should be properly selected.
\begin{equation}
\label{heaviside}
    \text{RP}_{i,j} = \mathbb{H}(\varepsilon - ||\mathbf{v}_j - \mathbf{v}_l||), \forall i,j \in [L]
\end{equation}
In all experiments, we used $\tau=1, m=1$. The threshold $\varepsilon$ was set to 10\% of the maximum distance. 
\subsection{Gramian Angular Field}
The Gramian Angular Field (GAF) is an image encoding method used for time series data. It has two subtypes: the Gramian Angular Summation Field (GASF) and the Gramian Angular Difference Field (GADF). In our supplementary experiment, we utilized the GASF subtype of GAF and referred to them as GAF unless otherwise indicated.

The data is first scaled between the range $[-1, 1]$ to represent the time series in a polar coordinate system using $\phi_i=\arccos{(x_i)}$ for $1,...,T$. A Gram matrix is calculated between all pairs of $\phi_{i=1,...,T}$ as in \cref{sup:gasf} and is used as the GAF. 

\begin{equation}
\label{sup:gasf}
\mathbf{GAF} = 
\begin{pmatrix}
\cos(\phi_1+\phi_1) & \cos(\phi_1+\phi_2) & \cdots &  \cos(\phi_1+\phi_T)\\
\cos(\phi_2+\phi_1) & \cos(\phi_2+\phi_2) & \cdots & \cos(\phi_2+\phi_T)\\
\vdots  & \vdots  & \ddots & \vdots  \\
\cos(\phi_T+\phi_1)& \cos(\phi_T+\phi_2) & \cdots & \cos(\phi_T+\phi_T)\\
\end{pmatrix}
\end{equation}\\

\begin{figure}[ht]
\centerline{\includegraphics[scale=0.35]{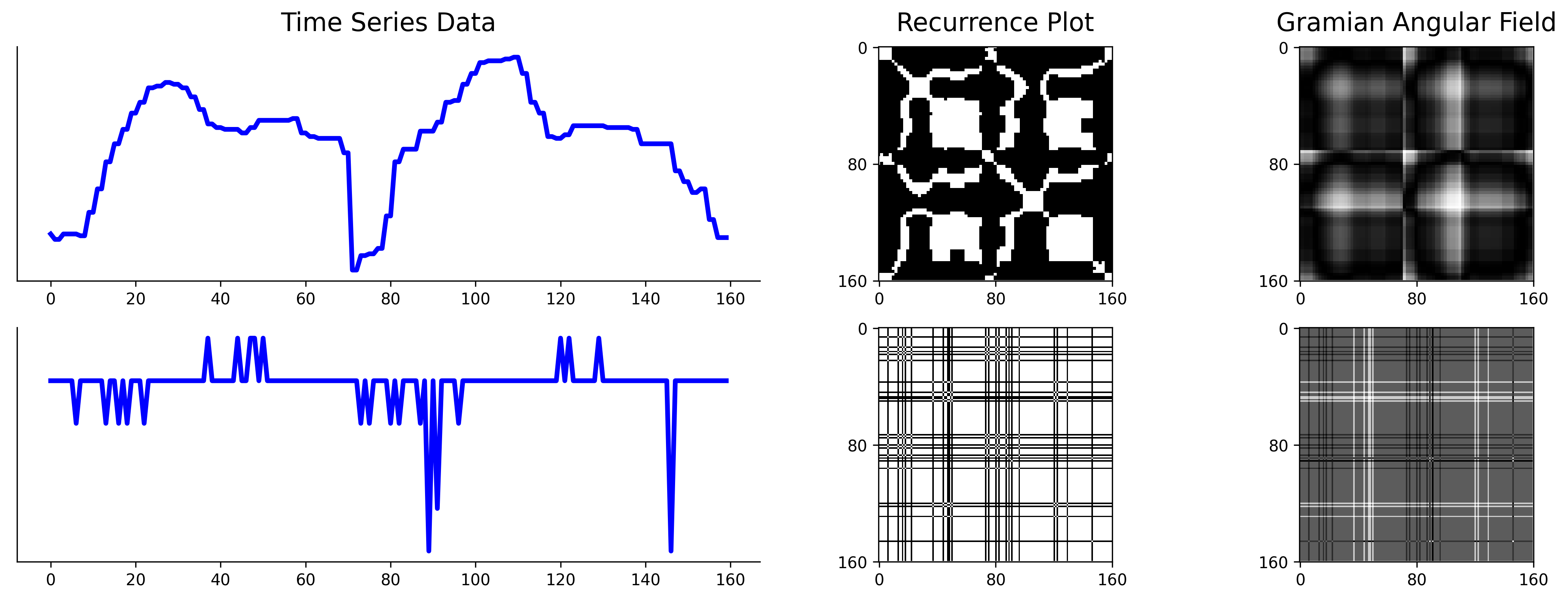}}
\caption{\textbf{Different Image Encoding Methods} The time series presented in the left column is encoded with Recurrence Plot (middle) and Gramian Angular Field (right).}
\label{figsup:imageconvert}
\vspace{-8pt}
\end{figure}

An illustrative figure of how time series is converted to an RP and GAF is shown in \cref{figsup:imageconvert}.

\begin{figure}[ht]
\centerline{\includegraphics[scale=0.3]{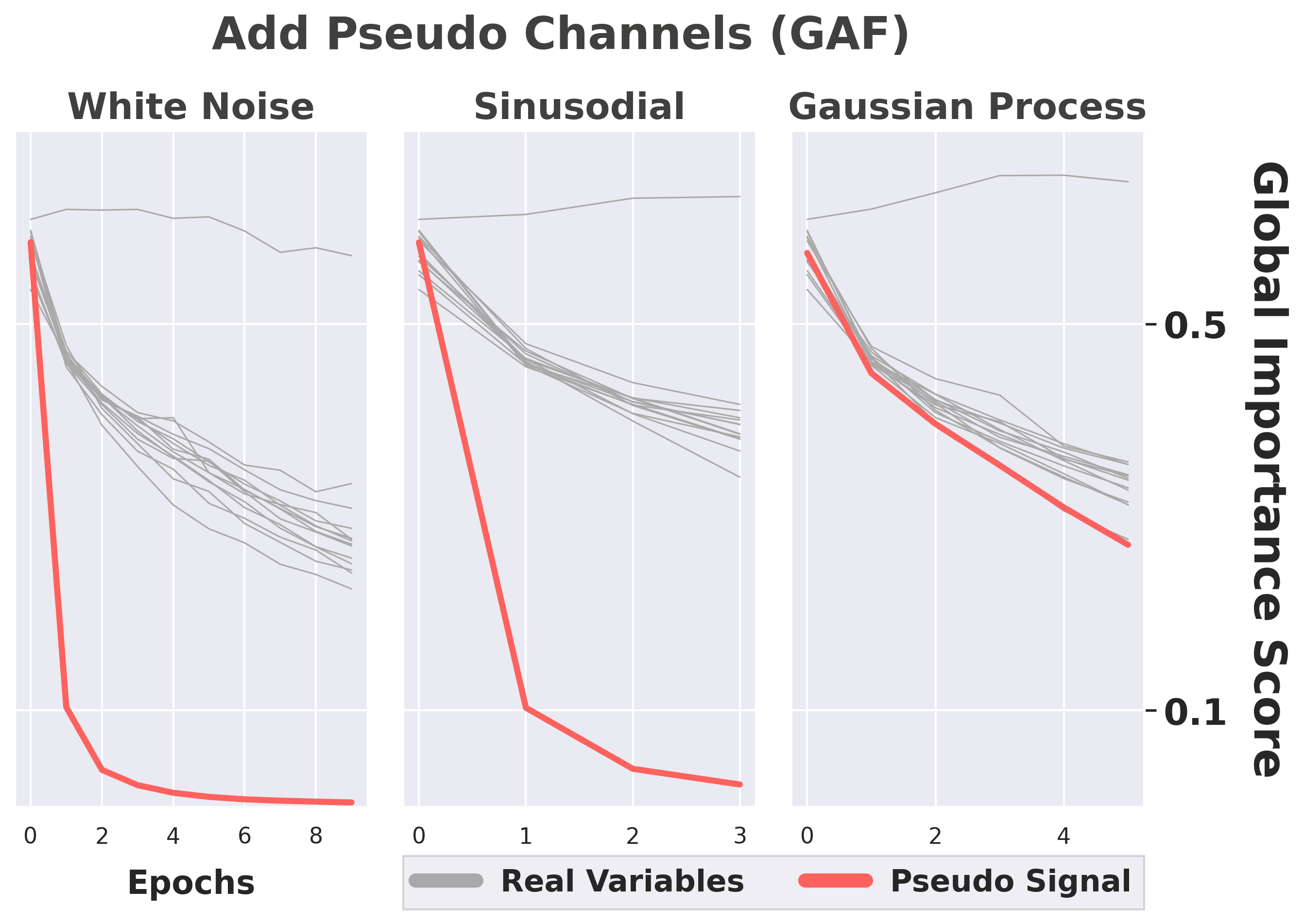}}
\caption{\textbf{Robustness to Pseudo Signals (GAF)} Pseudo signals were incorporated and trained on the Gilon activity task, as illustrated in \cref{fig:spearman}. In this instance, we employed the GAF image encoding method rather than RP. The findings demonstrate that encoding techniques other than RP exhibit robustness against pseudo signals.}
\label{figsup:gaf_pseudo}
\vspace{-8pt}
\end{figure}

% \usepackage{multirow}
\begin{table}[!htb]
\centering
\fontsize{8}{9}\selectfont{
\begin{tabular}{cccccc}
\textbf{Models}      & \textbf{Methods} & \multicolumn{4}{c}{\textbf{SquidGame}}                                                                            \\ \hline
\multicolumn{1}{l}{} &                  & \textbf{F1($\uparrow$)} & \textbf{Jaccard($\uparrow$)} & \textbf{IACC($\uparrow$)} & \textbf{Jaccard($\uparrow$)} \\ \hline
                     & CE               & 0.609\tiny{$\pm$0.07}   & 0.441\tiny{$\pm$0.07}        & 0.527\tiny{$\pm$0.09}     & 0.441\tiny{$\pm$0.07}        \\
                     & CE+QR            & 0.815\tiny{$\pm$0.06}   & 0.692\tiny{$\pm$0.09}        & 0.869\tiny{$\pm$0.05}     & 0.692\tiny{$\pm$0.09}        \\
\multirow{-3}{*}{ShuffleNet} &
  \cellcolor[HTML]{EFEFEF}\textbf{GAINS} &
  \cellcolor[HTML]{EFEFEF}{\color[HTML]{FE0000} \textbf{0.21}} &
  \cellcolor[HTML]{EFEFEF}{\color[HTML]{FE0000} \textbf{0.25}} &
  \cellcolor[HTML]{EFEFEF}{\color[HTML]{FE0000} \textbf{0.34}} &
  \cellcolor[HTML]{EFEFEF}{\color[HTML]{FE0000} \textbf{0.25}} \\ \hline
                     & CE               & 0.673\tiny{$\pm$0.06}   & 0.510\tiny{$\pm$0.06}        & 0.598\tiny{$\pm$0.04}     & 0.510\tiny{$\pm$0.06}        \\
                     & CE+QR            & 0.817\tiny{$\pm$0.05}   & 0.693\tiny{$\pm$0.07}        & 0.871\tiny{$\pm$0.03}     & 0.693\tiny{$\pm$0.07}        \\
\multirow{-3}{*}{ResNet} &
  \cellcolor[HTML]{EFEFEF}\textbf{GAINS} &
  \cellcolor[HTML]{EFEFEF}{\color[HTML]{FE0000} \textbf{0.14}} &
  \cellcolor[HTML]{EFEFEF}{\color[HTML]{FE0000} \textbf{0.18}} &
  \cellcolor[HTML]{EFEFEF}{\color[HTML]{FE0000} \textbf{0.27}} &
  \cellcolor[HTML]{EFEFEF}{\color[HTML]{FE0000} \textbf{0.18}} \\ \hline
                     & CE               & 0.475\tiny{$\pm$0.06}   & 0.314\tiny{$\pm$0.06}        & 0.622\tiny{$\pm$0.10}     & 0.314\tiny{$\pm$0.06}        \\
                     & CE+QR            & 0.703\tiny{$\pm$0.10}   & 0.551\tiny{$\pm$0.12}        & 0.778\tiny{$\pm$0.08}     & 0.551\tiny{$\pm$0.12}        \\
\multirow{-3}{*}{MLP-Mixer} &
  \cellcolor[HTML]{EFEFEF}\textbf{GAINS} &
  \cellcolor[HTML]{EFEFEF}{\color[HTML]{FE0000} \textbf{0.23}} &
  \cellcolor[HTML]{EFEFEF}{\color[HTML]{FE0000} \textbf{0.24}} &
  \cellcolor[HTML]{EFEFEF}{\color[HTML]{FE0000} \textbf{0.16}} &
  \cellcolor[HTML]{EFEFEF}{\color[HTML]{FE0000} \textbf{0.24}} \\ \hline
                     & CE               & 0.751\tiny{$\pm$0.08}   & 0.607\tiny{$\pm$0.10}        & 0.678\tiny{$\pm$0.10}     & 0.607\tiny{$\pm$0.10}        \\
                     & CE+QR            & 0.811\tiny{$\pm$0.04}   & 0.684\tiny{$\pm$0.06}        & 0.860\tiny{$\pm$0.03}     & 0.684\tiny{$\pm$0.06}        \\
\multirow{-3}{*}{ViT} &
  \cellcolor[HTML]{EFEFEF}\textbf{GAINS} &
  \cellcolor[HTML]{EFEFEF}{\color[HTML]{FE0000} \textbf{0.06}} &
  \cellcolor[HTML]{EFEFEF}{\color[HTML]{FE0000} \textbf{0.08}} &
  \cellcolor[HTML]{EFEFEF}{\color[HTML]{FE0000} \textbf{0.18}} &
  \cellcolor[HTML]{EFEFEF}{\color[HTML]{FE0000} \textbf{0.08}} \\ \hline
\end{tabular}}
\caption{\textbf{Performance Evaluation of CWRI Metrics using GAF encoding on the SquidGame task.} We observe that using QR-Ortho Loss improves the identification of important features. Here, we set~$\lambda=1$ for all models. }
\label{table_gaf}
\end{table}

\newpage
\section{GI Full Results}
\label{appendix:GI_FULL}
\begin{table}
% \usepackage{multirow}
\centering
\fontsize{7}{7}\selectfont{
} & DM                                  & \textbf{0.354}                                                & -0.843                                                        & \textbf{0.308}                                                & \textbf{0.495\tiny{$\pm$0.19}}                                                & 0.358\tiny{$\pm$0.16}                                                         & \multicolumn{1}{c|}{\multirow{-7}{*}{0.952}}                                       & \textbf{0.150}                         & -2.127                                                         & \textbf{0.075}                                                & 0.063\tiny{$\pm$0.36}                                                         & 0.040\tiny{$\pm$0.28}                                                         & \multirow{-7}{*}{0.807}                                       \\ \hline
                                                                            & GS                                  & 0.555                                                         & 2.209                                                         & 0.226                                                         & 0.399\tiny{$\pm$0.22}                                                         & 0.275\tiny{$\pm$0.17}                                                         & \multicolumn{1}{c|}{}                                                              & -0.048                                 & 2.513                                                          & 0.058                                                         & 0.189\tiny{$\pm$0.37}                                                         & 0.173\tiny{$\pm$0.31}                                                         &                                                               \\
                                                                            & SVS                                 & 0.113                                                         & -0.858                                                        & 0.078                                                         & 0.060\tiny{$\pm$0.24}                                                         & 0.046\tiny{$\pm$0.19}                                                         & \multicolumn{1}{c|}{}                                                              & -0.335                                 & 1.653                                                          & -0.008                                                        & 0.257\tiny{$\pm$0.37}                                                         & 0.253\tiny{$\pm$0.34}                                                         &                                                               \\
                                                                            & Saliency                            & 0.528                                                         & \textbf{3.093}                                                & 0.248                                                         & \textbf{0.618\tiny{$\pm$0.14}}                                                & \textbf{0.468\tiny{$\pm$0.11}}                                                & \multicolumn{1}{c|}{}                                                              & 0.007                                  & 3.054                                                          & 0.105                                                         & 0.234\tiny{$\pm$0.29}                                                         & 0.200\tiny{$\pm$0.30}                                                         &                                                               \\
                                                                            & FA                                  & 0.230                                                         & -2.185                                                        & 0.027                                                         & 0.165\tiny{$\pm$0.15}                                                         & 0.134\tiny{$\pm$0.11}                                                         & \multicolumn{1}{c|}{}                                                              & -0.065                                 & 1.927                                                          & 0.017                                                         & \textbf{0.474\tiny{$\pm$0.27}}                                                & \textbf{0.360\tiny{$\pm$0.26}}                                                &                                                               \\
                                                                            & IG                                  & \textbf{0.666}                                                & 2.298                                                         & \textbf{0.240}                                                & 0.346\tiny{$\pm$0.18}                                                         & 0.235\tiny{$\pm$0.14}                                                         & \multicolumn{1}{c|}{}                                                              & -0.048                                 & 0.764                                                          & 0.030                                                         & 0.189\tiny{$\pm$0.37}                                                         & 0.173\tiny{$\pm$0.31}                                                         &                                                               \\
\multirow{-6}{*}{\begin{tabular}[c]{@{}c@{}}Baseline\\ (TCN)\end{tabular}}  & DM                                  & -0.280                                                        & -0.270                                                        & -0.024                                                        & 0.077\tiny{$\pm$0.21}                                                         & 0.046\tiny{$\pm$0.14}                                                         & \multicolumn{1}{c|}{\multirow{-6}{*}{0.928}}                                       & \textbf{0.186}                         & \textbf{4.270}                                                 & \textbf{0.140}                                                & 0.000\tiny{$\pm$0.56}                                                         & -0.013\tiny{$\pm$0.49}                                                        & \multirow{-6}{*}{0.815}                                       \\ \hline
\begin{tabular}[c]{@{}c@{}}Baseline\\ (None)\end{tabular}                   & Laxcat                              & -0.458                                                        & -4.217                                                        & -0.230                                                        & -0.016\tiny{$\pm$0.34}                                                        & -0.007\tiny{$\pm$0.25}                                                        & \multicolumn{1}{c|}{0.711}                                                         & -0.225                                 & 1.003                                                          & 0.139                                                         & 0.074\tiny{$\pm$0.46}                                                         & 0.093\tiny{$\pm$0.40}                                                         & 0.759                                                         \\ \hline
\end{tabular}}

    \caption{Performance Evaluation of GI Metrics: This table presents the evaluation of GI metrics such as ABC, DA, WDA, $\rho_{S}$, $\rho_{K}$ from a five-fold cross-validation (CV) process. As ABC, DA, and WDA are metrics that are derived from the averaged outcomes of the five-fold CV, it is not possible to calculate a standard deviation for these metrics. For clarity, the top-performing result for each model is indicated in red bold, while the second-highest performance is denoted in black bold. The evaluation is divided into two sections: Panel A focuses on the outcomes of vision-based deep learning models, whereas Panel B details the performance of LSTM and TCN models, assessed using various explainer methods including Gradient Shap (GS), Shapley Value Sampling (SVS), Saliency, Feature Ablation (FA), Integrated Gradients (IG), FIT, and DynaMask (DM). Here, Laxcat is featured both as an explainer method and as a model in its own.}
    \label{appendix_table_GI}
\end{table}

\textbf{CAFO Models.} ShuffleNet~\cite{zhang2018shufflenet}, ResNet~\cite{he2016deep}, MLP-Mixer~\cite{tolstikhin2021mlp}, and ViT (Vision Transformer)~\cite{dosovitskiy2020image}.\\
\textbf{Baseline Models.} LSTM (long short-term memory)~\cite{hochreiter1997long} and TCN (temporal convolutional network)~\cite{lea2017temporal}.\\
\textbf{Baseline Explainers.} Gradient Shap (GS) \cite{lundberg2017unified}, Shapley Value Sampling (SVS)~\cite{castro2009polynomial}, Saliency~\cite{simonyan2013deep}, Feature Ablation (FA)~\cite{suresh2017clinical}, Integrated Gradients (IG)~\cite{sundararajan2017axiomatic}, DynaMask (DM) \cite{crabbe2021explaining}, FIT~\cite{tonekaboni2020went}, and LAXCAT~\cite{hsieh2021explainable}.

\begin{figure*}[ht]
\centerline{\includegraphics[scale=0.47]{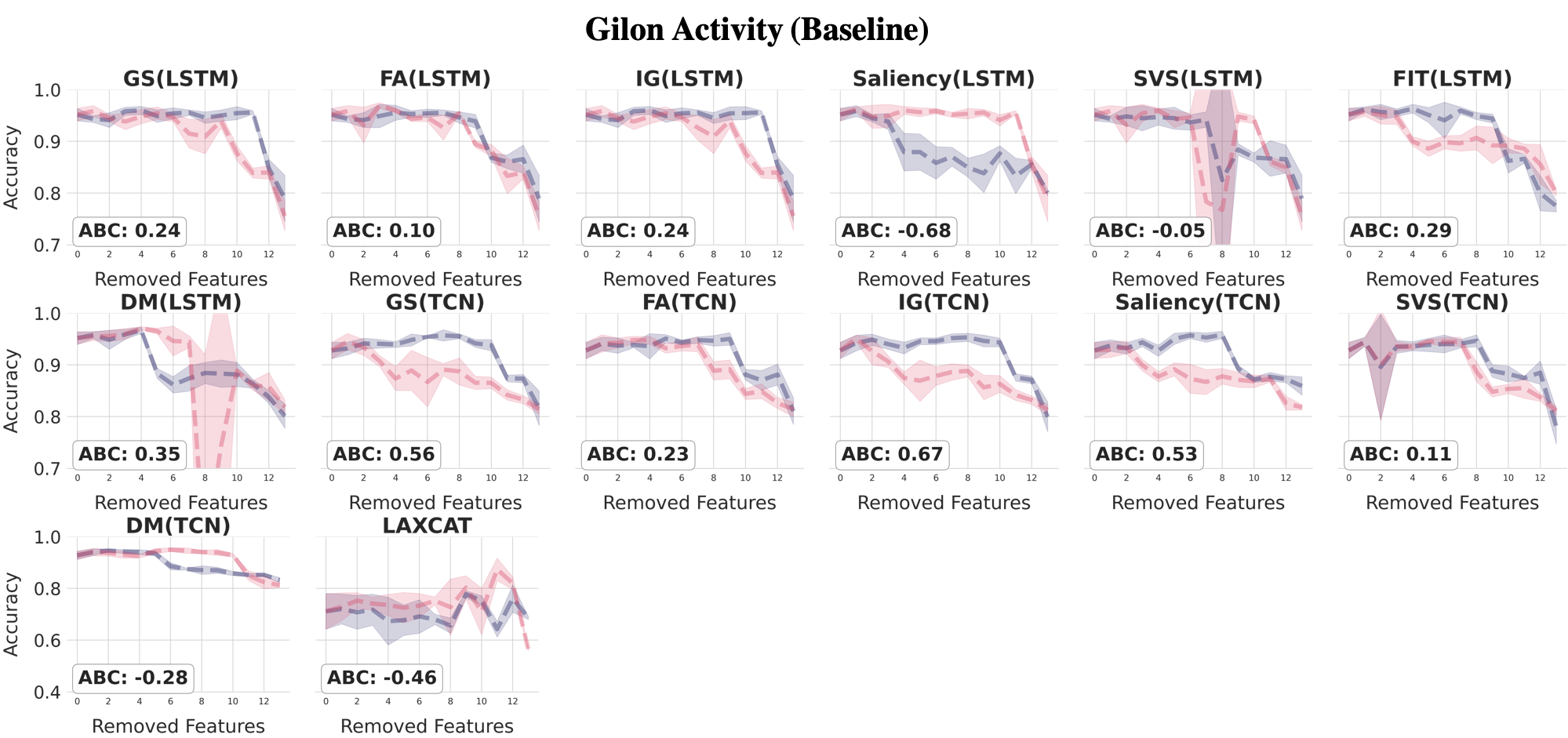}}
\vspace{-4pt}
\caption{The ROAR plot on the Gilon Activity task for all baseline models.}
\vspace{-8pt}
\end{figure*}

\begin{figure*}[ht]
\centerline{\includegraphics[scale=0.47]{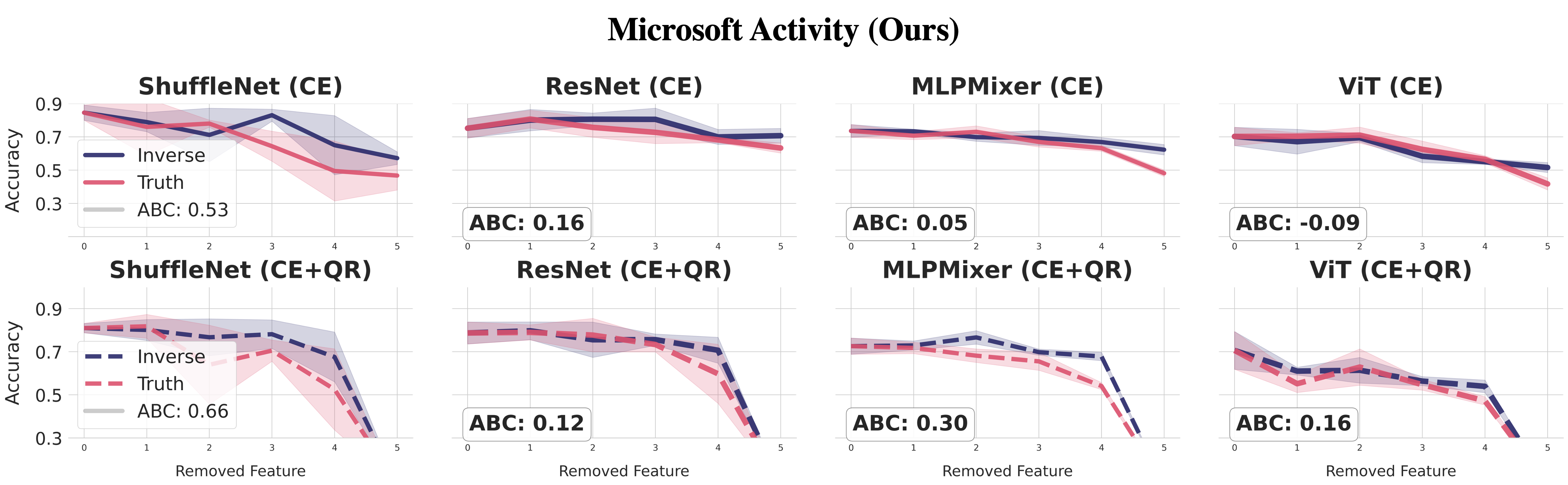}}
\vspace{-4pt}
\caption{The ROAR plot on the Microsoft Activity task using CAFO. (Top row) Only CE(cross entropy) loss is applied. (Bottom row) CE+QR-Ortho Loss is applied. We observe that applying CE+QR-Ortho Loss leads to increased ABC (Area Between Curves) for all models except ResNet.}
\vspace{-8pt}
\end{figure*}

\begin{figure*}[ht]
\centerline{\includegraphics[scale=0.47]{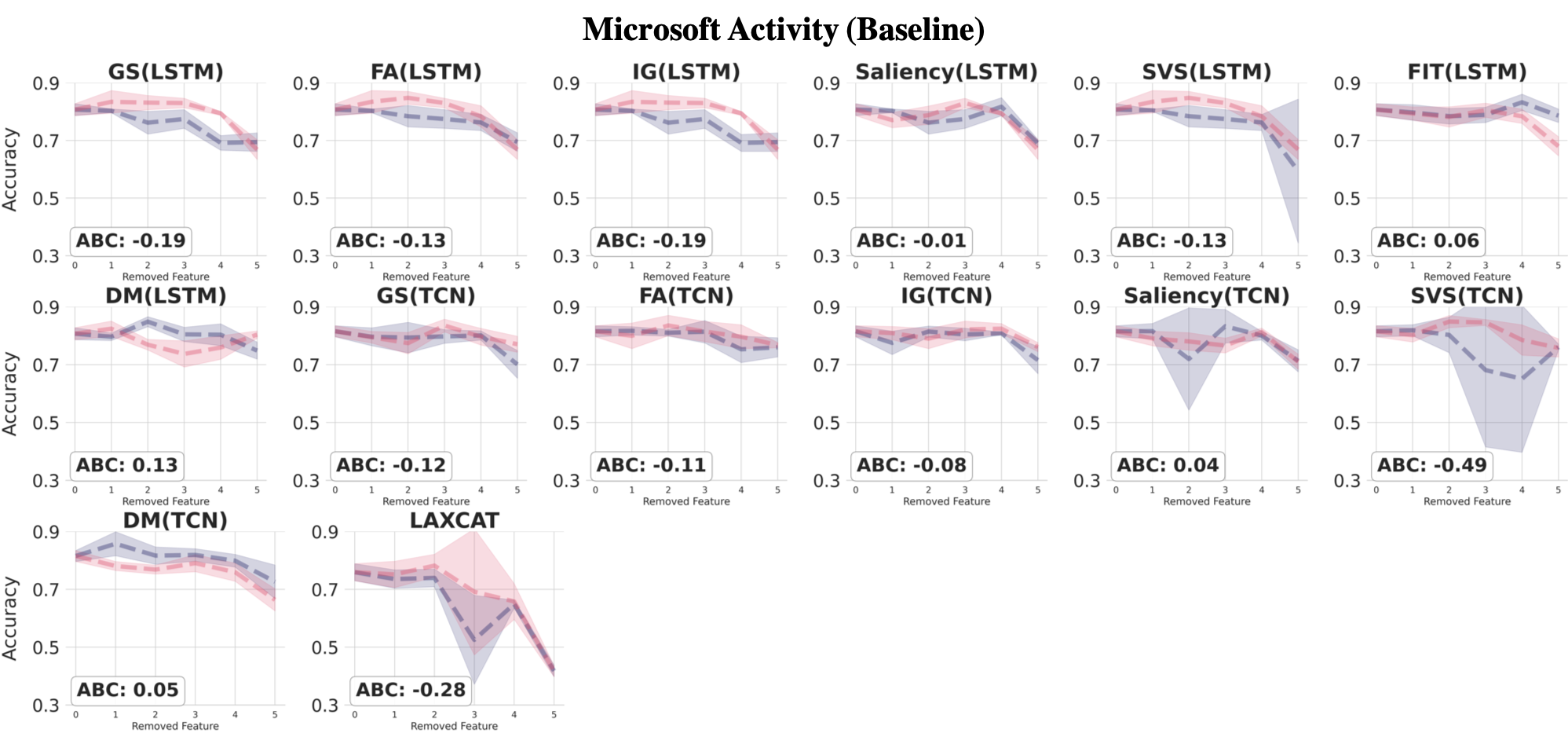}}
\vspace{-4pt}
\caption{The ROAR plot on the Microsoft Activity task for all baseline models.}
\end{figure*}

\newpage
\subsection{GI Metric Calculation}

\label{appendix:gi_metric_calculation}
\textbf{Area between curve (ABC)}. Here, we assume that we have a training set which is split into five-fold cross-validation (CV), and a left-out test set. Each CV fold serves as a validation set against this test set. We start with a complete set of features, performing five-fold CV to derive feature rankings based on Global Importance (GI) scores from each run, which are then averaged to establish the final GI rankings. These features are arranged in descending order for the `truth' rank and ascending order for the `inverse' rank. Sequentially, we remove each feature from both training and test sets to assess model performance, hypothesizing that removing a critical feature decreases accuracy, while removing a non-essential feature doesn't impact performance significantly. After eliminating all features, we calculate the ABC metric by measuring the area between the two resulting performance curves ($f(x), g(x)$),
    \[
    \textbf{ABC}\triangleq\int_a^b (f(x)-g(x))dx \simeq \frac{1}{2}\sum_{i=1}^{n-1}(f(x_i)-g(x_i)+f(x_{i+1}) - g(x_{i+1})) \Delta x_i
    \]

\noindent\textbf{Drop in accuracy (DA)}. The drop in accuracy (DA) calculates the percentage decrease in accuracy of a model when $K$\% percentage of the most important features (denoted as K\_acc) identified by the model are dropped, compared to the base accuracy (when no features are dropped; denoted as base\_acc). This is done by subtracting the accuracy after dropping $K$\% of the total features from the base accuracy and then dividing by the base accuracy. The result is multiplied by 100 to express it as a percentage. Here, we set $K=20\%$. This metric helps in understanding the impact of removing the most important features on the model's performance. Mathematically, DA is given as 
   \[
   \textbf{DA} \triangleq \left( \frac{\text{base\_acc} - \text{K\_acc}}{\text{base\_acc}} \right) \times 100
   \]

\noindent\textbf{Weighted drop in accuracy (WDA)}. While DA provides a snapshot of the degraded model performance at a specific $K$, the weighted drop in accuracy (WDA) complements DA by considering the impact of dropping each feature individually and combines these impacts in a weighted manner. To compute WDA, consider $D$ as the total number of features in the dataset. For each feature indexed by $d$ (where $d$ ranges from $0,...,D-1$), the weight $w_d$ is calculated as follows:
   \[
     w_d = \frac{1}{D-1} \times (D-d-1)
   \]
In this formula, the weight $w_d$ decreases linearly with the index d, reflecting the diminishing marginal impact of removing additional features. For instance, given 14 features as in the Gilon task, the weights are given as $1.0, 0.923, 0.846, 0.769, 0.692, 0.615, 0.538, 0.461, 0.384, 0.307, \\0.230, 0.153, 0.077$. The WDA is calculated as below. Here, base\_acc is the base accuracy, and d\_acc represents the accuracy of model when $d$ features are removed.
   \[
   \textbf{WDA} \triangleq \sum_{d=0}^{D-1}\left( \text{base\_{acc}} - \text{d\_{acc}}\right) w_d
   \]

\noindent\textbf{Spearman $\rho_{s}$ and Kendall $\rho_{k}$}. As in the area between curve (ABC) metric computation, we assume a training-set consisting of five-fold CV, with a left-out test set. Using each fold as a validation set, we can obtain five feature ranks. We calculate the pairwise spearman correlation and Kendall correlation, and state the average value.

\newpage
\section{CWRI Full Results}
\label{appendix:cwri_fullresults}
\begin{table}[ht]
% \usepackage{multirow}
\centering
\fontsize{8}{8}\selectfont{
\begin{tabular}{cccccccc}
\hline
\multicolumn{8}{l}{\textit{\textbf{Panel A. Performance Comparison of Various Models on CE and QR loss}}}                                                                                                                                                                                                                                                                                                                                                                                                                                                                             \\
\textbf{Models}                        & \textbf{Methods}                             & \multicolumn{3}{c}{\textbf{SquidGame}}                                                                                                                                                                                                        & \multicolumn{3}{c}{\textbf{WhichFinger}}                                                                                                                                                                                                      \\ \hline
\multicolumn{1}{l}{}                   &                                              & \textbf{F1($\uparrow$)}                                                       & \textbf{Jaccard($\uparrow$)}                                                  & \textbf{IACC($\uparrow$)}                                                     & \textbf{F1($\uparrow$)}                                                       & \textbf{Jaccard($\uparrow$)}                                                  & \textbf{IACC($\uparrow$)}                                                     \\
\hline
\end{tabular}}

    \caption{Performance Evaluation of CWRI Metrics. This table presents the evaluation of CWRI metrics using F1 Score, Jaccard, and Accuracy (IACC). We use the term IACC to differentiate with model accuracy. The metrics are measured based on the predicted class-wise importance of each explainer methods against the established ground truth class-wise importance. All experiments were conducted with a five-fold cross validation. }
    \label{appendix_table_CWRI}
\end{table}

\newpage
\section{Evaluation of CWRI Metrics}
\label{appendix:cwri_metrics}
During the evaluation of the CWRI measure, we encountered a dilemma regarding the establishment of ground truth for each class. While we can identify the exact features relevant to each class, the model is not obligated to base its predictions on these features exclusively. For instance, flexing the thumb (class 1-Thumb only; \cref{suppfig:fingerclass})) implies that the remaining four fingers remain still, which also serves as a useful identifier for the class. This deviates from spurious correlation, as the association between features and class predictions is not merely coincidental.

\begin{figure*}[ht]
\centerline{\includegraphics[scale=0.2]{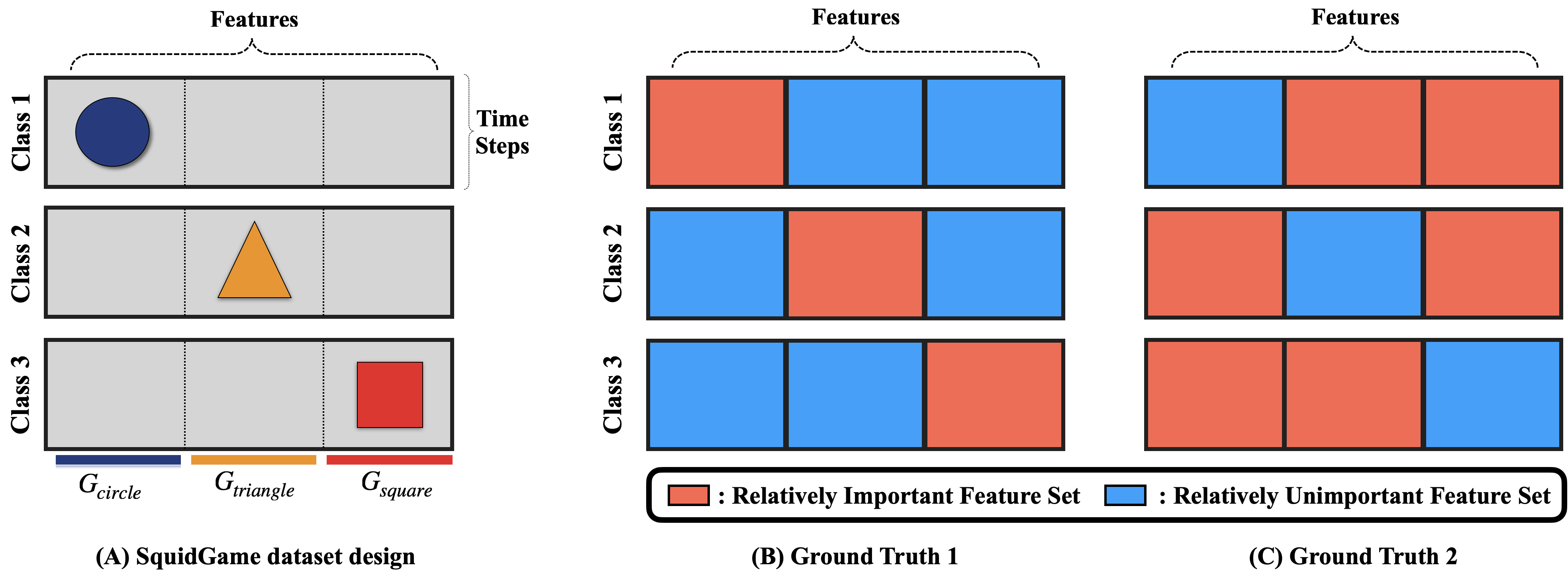}}
\vspace{-4pt}
\caption{\textbf{Ground Truth Design of SquidGame.} \textbf{(A)} The design of SquidGame task. \textbf{(B) \& (C)}. Two distinct ground truth label types are established for the SquidGame task. In this context, the red feature sets signify more critical or influential features, whereas the blue feature sets represent those that are comparatively less important or impactful. }
\label{supple:squidgame_gt}
\end{figure*}

\begin{figure*}[ht]
\centerline{\includegraphics[scale=0.2]{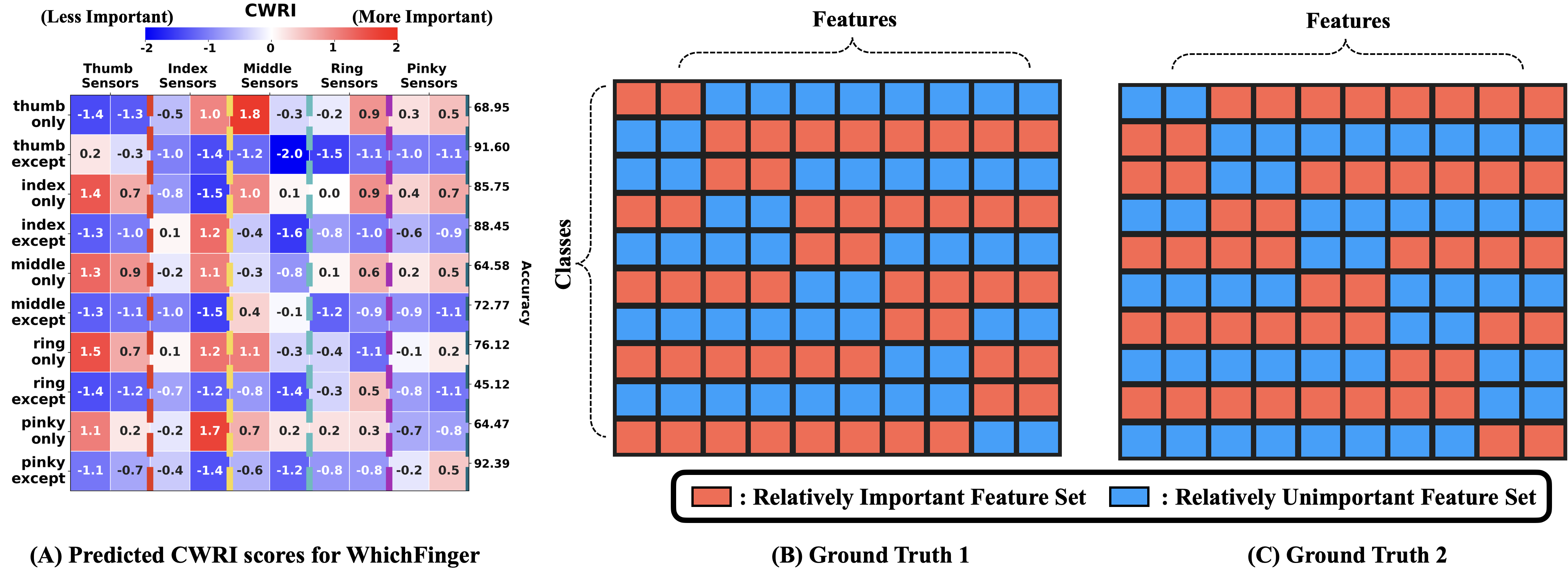}}
\caption{\textbf{An example of a predicted CWRI score for the WhichFinger task.} \textbf{(A)} The CWRI score is predicted using MLP-Mixer. \textbf{(B) \& (C)}. Two distinct ground truth label types are established for the WhichFinger task. In this context, (A) follows the ground truth label of (C) rather than (B). }
\label{supple:whichfinger_gt}
\end{figure*}

In order to quantitatively assess the CWRI score, we generated two distinct sets of ground truth labels, as illustrated in \cref{supple:squidgame_gt} and \cref{supple:whichfinger_gt}. Each of these ground truth label sets was utilized to evaluate the predicted CWRI score. We selected the ground truth label that produced a higher F1 score, as it represented the only feasible approach for evaluating the CWRI score. This decision was grounded in the understanding that the model may have considered these feature sets crucial during that particular run.

Throughout our experimentation, we observed that a model adheres to a consistent ground truth label, notwithstanding variations in seed and cross-validation folds. Nevertheless, the choice of ground truth may shift between different architectures and the application of QR-Ortho Loss. Determining which ground truth label the model will adopt remains an open question and constitutes a challenge that we intend to investigate in future work.

\newpage
\section{Datasets}
\label{appendix:gi_data}

\begin{table}[htbp]
\centering
\begin{tabular}{c c c c c c}
\toprule
    Dataset &  \#Samples &  Length & \#Users &  \# Features &  \# Class \\
\midrule
Gilon & 47,647 & 160 & 72 & 14 & 7 \\ 
MS & 14,201 & 200 & 93 & 6 & 10 \\ 
SquidGame & 54,000 & 32 & - & 30 & 3 \\
WhichFinger & 18,010 & 120 & 19 & 10 & 10 \\ \hline
\end{tabular}
\caption{Statistics of the datasets used in our experiment}
\end{table}
\vspace{-8pt}

\begin{figure}[ht]
\centerline{\includegraphics[scale=0.15]{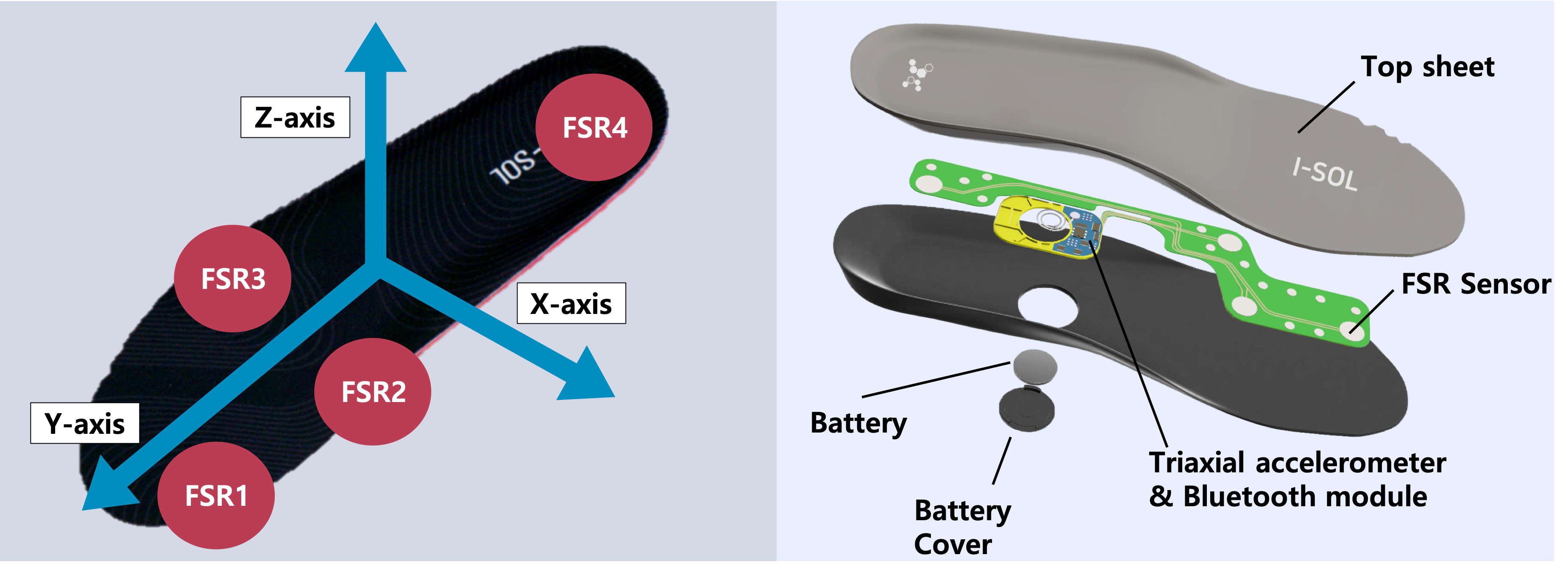}}
\caption{Smart insole used to collect the Gilon Activity task. Used with permission \cite{kim2023multi}.}
\label{figsup:gilon}
\end{figure}

\noindent\textbf{Gilon Activity}~\cite{kim2023multi}. This dataset comprises smart insole measurements gathered from 72 distinct users in South Korea. The smart insole records 14 different sensor measures, split evenly between the left and right feet. These metrics include three-dimensional accelerometer readings (X, Y, Z) and data from four force-sensitive resistors (FSRs). These participants engaged in seven specific activities while equipped with the smart insole. The activities include standing still (0), walking on the ground (1), walking on a treadmill at a constant speed (2), running on a treadmill at a constant speed (3),  performing lunges (4), squats (5), and jumping jacks (6). The original study established predefined train/test splits based on the users. As a result, the training set is composed of data from 50 users, amounting to 33,368 samples, while the test set consists of data from 22 users, totaling 14,279 samples. Further, within the training set, the data is divided into five-folds. Each fold uses a different group of users as a validation set, ensuring a balanced evaluation across the dataset. The dataset was segmented into windows size of 160 with an overlapping length of 120. The five-fold cross validation in our research was conducted by using each validation split with the fixed test set. The dataset (https://github.com/Gilon-Inc/GILON-MULTITASK-DATASET) can be accessed by request. The dataset is protected under CC-BY-NC-ND-4.0 license.

\noindent\textbf{MS Activity}~\cite{morris2014recofit}. The original dataset (exercise\_data.50.0000\_singleonly.mat) contains 114 users performing 72 distinct gym exercises, recorded in 4,686 sessions with wearable arm sensors. The armband includes a 3-axis accelerometer (X, Y, Z) and gyroscope (X, Y, Z). To address the imbalance in the number of sessions across activities, we refined the dataset by selecting exercises with session counts ranging between 25 and 70, based on an empirical analysis of the data distribution. We then focused on activities featuring similar physical motions to effectively showcase our class-wise importance metric. This resulted in a curated set of activities, including Bicep Curl (0), Biceps Curl with Band (1), Jump Rope (2), Plank (3), Pushups (4), Squat (5), Squat with Hands Behind Head (6), Squat Jump (7), Walk (8), and Walking Lunge (9). The data was segmented into non-overlapping windows of 200 samples each. The final processed dataset comprises training (65 users; 10,892 samples) and testing subsets (28 users; 3,309 samples), with the training data further divided into five-fold cross validation, following the approach used in the Gilon dataset. The raw dataset can be accessed in (https://github.com/microsoft/Exercise-Recognition-from-Wearable-Sensors).

\begin{figure}[!htb]
\centerline{\includegraphics[scale=0.13]{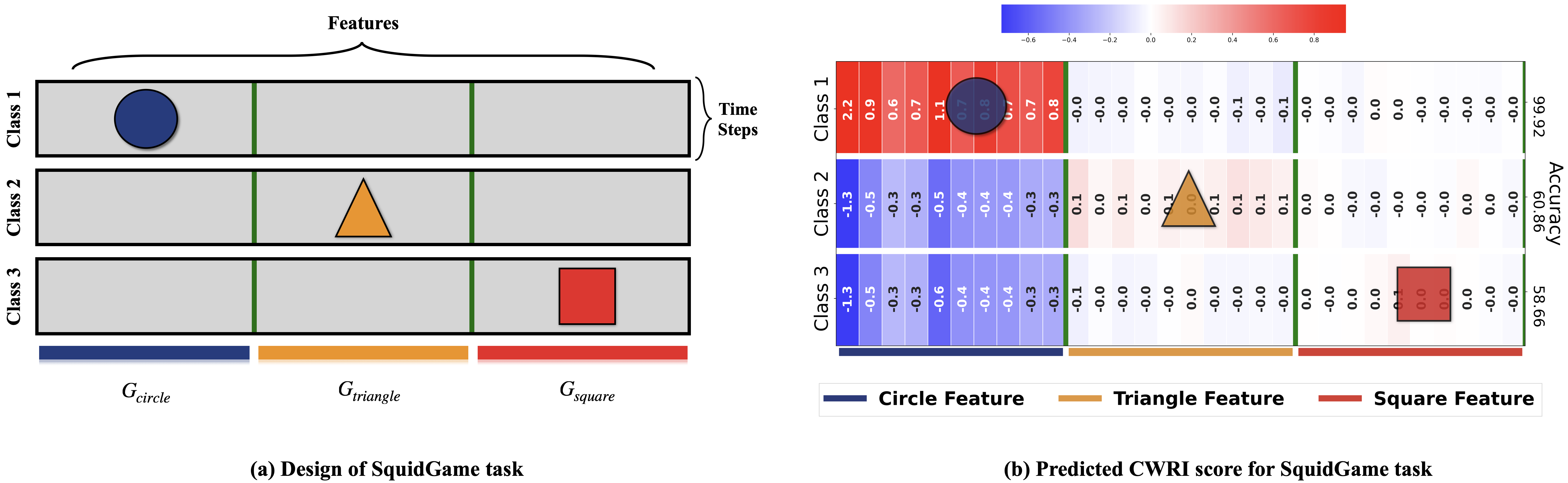}}
\caption{(a) The design of SquidGame task. (b) The predicted CWRI score for the SquidGame. }
\label{figsup:gilon}
\vspace{-8pt}
\end{figure}

\noindent\textbf{SquidGame.} The SquidGame task is a synthetic 3-class classification dataset designed to validate the efficacy of CWRI scores. This task features distinct, non-overlapping feature sets, each representing a specific class. Initially, all features are populated with random Gaussian noise. These time series instances are then treated as $T\times D$ images, and an empty mask is created with circle, triangle, and square shapes. Each mask is filled with characteristic time signals from the sinusoidal series, with the length, size, and center coordinates generated randomly for each instance. Here, $T=32$ and $D=30$. Feature indices are divided into three groups: $\textbf{\textit{G}}_{circle}=\{1,...,10\}$, $\textbf{\textit{G}}_{triangle}=\{11,...,20\}$, and $\textbf{\textit{G}}_{square}=\{21,...,30\}$. $\textbf{\textit{G}}_{circle}$ indicates that feature indices 1 to 10 contain a circular mask with characteristic time signals representing the first class. The complement of the circle masks in indices 1 to 10 and the remaining indices (11 to 30) are filled with Gaussian noise for the first class. This scheme is also applied to the second and third classes. As a result, only the first class contains characteristic time signals in feature indices 1 to 10, while the other classes have random Gaussian noises. The dataset's name was inspired by the popular Netflix series "Squid Game."

\noindent\textbf{WhichFinger.} The WhichFinger task utilizes a real-world smart glove dataset that we have gathered in order to verify the effectiveness of the CWRI scores. In this task, we set $T=120$ and $D=10$. A comprehensive discussion of the dataset's details can be found in \cref{appendix:whichfinger}, which is dedicated to this particular topic.
\section{WhichFinger Dataset}
\label{appendix:whichfinger}
\begin{figure*}[h]
\centerline{\includegraphics[scale=0.25]{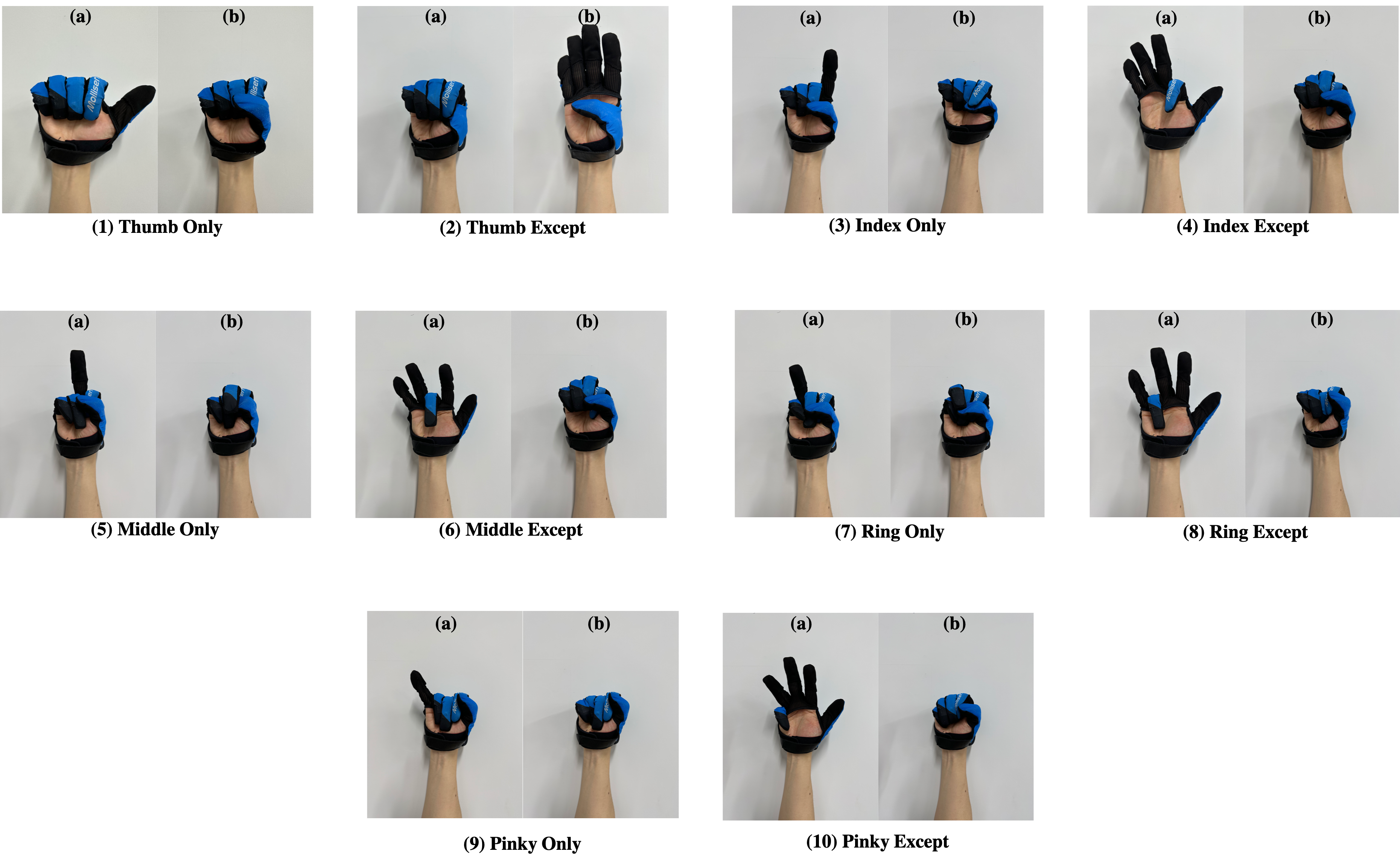}}
\vspace{-4pt}
\caption{The figure illustrates the detailed finger movement for each class. (a) and (b) for each class are repeated for more than one minute.}
\label{suppfig:fingerclass}
\vspace{-5pt}
\end{figure*}

The WhichFinger Dataset is a multivariate time series (MTS) dataset, designed for eXplainable Artificial Intelligence (XAI) applications. This dataset offers comprehensive information on the data collection process for each class, as well as the features relevant to specific classes, which facilitates the validation of the CWRI measure. We created this dataset because, to the best of our knowledge, no public MTS datasets met the following three criteria: (1) strong prior knowledge or information regarding each feature's contribution to specific classes, (2) a sufficient number of classes~$(C\geq2)$ and features~$(D\geq2)$, and (3) an adequate number of samples $(N\geq1,000)$. In this section, we describe the detailed data collection process and the preprocessing steps involved in the creation of the dataset.

\subsection{Recruitment of Subjects}
We recruited 20 volunteer participants through flyer advertisements and provided each participant with a fee of approximately \$7 (USD) for their participation in the experiment. The experiment was approved by the UNIST Institutional Review Board (IRB) (UNISTIRB-23-008-A).

\subsection{Smart Glove}
\textbf{Sensors}.
The smart glove is equipped with an integrated soft sensor that captures precise finger movements. Each finger contains two sensors, resulting in a total of 10 sensor recordings. These sensors detect changes in resistance as the fingers flex and relax. The data is recorded at approximately~66.7 Hz.

\subsection{Data Collection Process}
\textbf{Experiment Condition}. Participants were instructed to wear the smart glove on their right hand. The supervisor sat adjacent to each participant, closely monitoring the data collection process during the entire experiment. The supervisor also provided a thorough explanation of the experimental procedures and specific actions required from the participants.

\noindent\textbf{Experiment Procedure}. 
A total of 10 unique finger movements, corresponding to 10 different categories, were recorded. The finger movements captured included: (1) Thumb only, (2) Thumb except, (3) Index only, (4) Index except, (5) Middle only, (6) Middle except, (7) Ring only, (8) Ring except, (9) Pinky only, and (10) Pinky except. These specific finger movements are depicted in detail in \cref{suppfig:fingerclass}. For each category, participants carried out the actions for one minute. Upon completing every two categories, a one-minute break was provided, or extended if requested by the participant. The entire data recording process took approximately 20 minutes for each participant to complete.

As the experiment aimed to identify specific features contributing to each class, it was necessary to minimize undesired finger movements. However, due to the interconnected nature of finger muscles, it is impossible to keep the other fingers completely still while moving one finger. Therefore, participants were encouraged to use their left hand to support and constrain the fingers that needed to remain stationary. When required, duct tape was also employed to reinforce these constraints. Despite these measures to reduce extraneous movements, they could not be entirely eliminated, leading to sensor measurements that still captured some unintended finger motions. 

\noindent\textbf{Data Exclusion}. One participant had to be excluded from the study as a result of data records being lost during the data collection process, resulting in 19 user information in the final dataset.

\subsection{Data Preparation for Model Training}
\textbf{Data Shape}. We dropped the first and last 150 recordings (approximately 2 seconds) from each class measurement. The resulting dataset for each class contains on average $4190 \pm 268$ recordings (rows) for each finger activity.  

\noindent\textbf{Data Preprocessing}. We adopted a similar preprocessing approach for the recorded data as described in the work of ~\cite{kim2023multi}. The multivariate time series (MTS) data was segmented into consecutive 2-second time windows (120 rows) with a 75\% overlap (80 rows) between windows. This segmentation was performed within each class and participant. Consequently, for each class, we obtained an average of 102 MTS instances, calculated as $(\frac{(4190 - 120)}{120 - 80} + 1)=102$ for each class recording.

\noindent\textbf{Data Split}. Out of the 19 participants, four were randomly chosen for the test set. The remaining 15 participants were divided into five groups, with each group serving as a validation set. This resulted in a 5-fold cross-validation (CV) setup.

\subsection{License for Data Use}
We apply the Creative Commons Attribution-NonCommercial 4.0 International License.

\newpage
\section{Implementation Details}
\label{appendix:ModelHyperParameters}
\subsection{Model Training}
In our experiments, we fixed all random seeds to $42$ and used the AdamW \cite{loshchilov2017decoupled} optimizer with learning rates of $0.002$ for Gilon and SquidGame, $0.005$ for WhichFinger, and $0.01$ for Microsoft Activity. Learning rate scheduling was not used.
\subsection{CAFO}
Throughout our experiments, we set the expansion filter number $\gamma=3$ in the DepCA module. The task-specific hyperparameter $\lambda$ in QR-Ortho loss requires a grid search, which we perform for $\lambda~\in~\{0.1, 0.2, 0.5, 1.0\}$ in each task, selecting based on the best validation accuracy. Consequently, we used $\lambda=0.5$ for Gilon, $\lambda=0.1$ for Microsoft, and $\lambda=1.0$ for both SquidGame and WhichFinger. 
\subsection{Deep Architectures}
\textbf{ShuffleNet}  We utilized grouped convolution, setting the group number to 3, and configuring the output channels as follows: $[D, 24, 120, 240, 480]$. Here, $D$ is the number of input feature channels.\\
\textbf{ResNet}. We used the original implementation of the ResNet but with 9 layers.\\
\textbf{MLP-Mixer}. For MLP-Mixer, we used the implementation from https://github.com/lucidrains/mlp-mixer-pytorch. We set the number of Mixer layers to 3 and the feed-forward layer dimension to 256.\\
\textbf{Vision Transformer}. We used the implementation from https://github.com/lucidrains/vit-pytorch. We configured the Transformer architecture with three blocks, each containing a multi-head attention mechanism with three heads. Additionally, the dimensions of all feed-forward layers were set to 256. For both MLP-Mixer and Vision Transformer, the patch size was set to 1/10th of the original input image for Gilon, Microsoft, and WhichFinger, while 1/8th was used for SquidGame due to its smaller image size.

\subsection{Baselines}
All explainers produce attributions for each time step, yielding an attribution size of $\mathbb{R}^{T\times D}$. We average the attributions feature-wise and utilize these values to compute metrics in our study.\\
\textbf{FIT}. We trained a feature generator using a gated recurrent unit (GRU) as in the original implementation of \cite{tonekaboni2020went}. The hidden dimension was set to 256, and 1 layer was used for GRU. We generated ten monte carlo samples for each time stamp.\\
\textbf{DynaMask}. We used the original implementation of \cite{crabbe2021explaining}. However, due to the immense computational complexity in optimizing the perturbation mask, we reduced the optimization step to 50 for each instance.\\
\textbf{LAXCAT}. We set a task specific kernel and stride size for 1-D convolution. As the original paper did not release the code, we used a third party implementation from https://github.com/Shuheng-Li/UniTS-Sensory-Time-Series-Classification. \\
\textbf{Others}. We used the implementations from the CAPTUM library \cite{kokhlikyan2020captum}. \\

\subsection{Hardware and Software}
\label{supp:hardware}
We conducted experiments on three GPU types: Nvidia TITAN RTX-24GB, Tesla V100-PCIE-32GB, and RTX A6000-48GB, utilizing PyTorch 1.8.1 and PyTorch-Lightning 1.6.5.

\newpage
\section{QR Feature Orthogonality Regularization Algorithm}
\label{appendix:qr_algo}

\begin{algorithm*}
\setlength{\baselineskip}{17pt}
\caption{QR-Ortho Loss Algorithm}
\label{algorithm_qr_ortho_loss}
\begin{algorithmic}[1]

\State \textbf{Input:} Training data $\textbf{\textit{K}} = \{(\mathbf{X}^{(i)}, y^{(i)})\}_{i=1}^N $, Instances in class $c$ denotes $\textit{\textbf{K}}_c$, Batch size $B$, Class set $C$, The matrix of class prototypes $\mathbf{\bar{A}} \in \mathbb{R}^{C \times D}$, Sum of the number of off-diagonal elements in $\mathbf{R}$ denoted as $r$, Strength of orthogonality $\lambda$, Depthwise Channel Attention denoted as $\textbf{\text{DepCA}}$, Image encoder denoted as $\textbf{RP}$
% \State $\mathbf{Q} \mathbf{R}$ = []
\State \textbf{For} mini-batch $\{(\mathbf{X}^{(i)}, y^{(i)})\}_{i=1}^{B} \subseteq \textbf{\textit{K}}$ \textbf{do} \textcolor{gray}{\Comment{Orthogonality phase for training data}}
\State $\quad \quad$ $\mathcal{X}^{(i)} = \textbf{\text{RP}}(\mathbf{X}^{(i)})$ \textcolor{gray}{\Comment{Encode raw time series using RP}}
\State $\quad \quad$ $\mathbf{a}^{(i)} = \textbf{\text{DepCA}}(\mathcal{X}^{(i)})$ \textcolor{gray}{\Comment{Calculate channel attention score}}
\State $\quad \quad$ $\mathbf{\bar{a}}_{c} = \frac{1}{|\textit{\textbf{K}}_c|} \sum_{i \in \textit{\textbf{K}}_c}\mathbf{a}^{(i)}$ \textcolor{gray}{\Comment{Calculate class prototypes for channel attention score}}
\State $\quad \quad$ $\mathbf{\bar{A}} = [\mathbf{\bar{a}}_{i}]_{i=1}^{C}$ \textcolor{gray}{\Comment{Stack $\mathbf{\bar{a}}_{c}$ in a row-wise manner}}
\State $\quad \quad$ $\text{Decomp}(\mathbf{\bar{A}})=\mathbf{Q} \mathbf{R}$ \textcolor{gray}{\Comment{QR decomposition}}
%\State $\quad \quad$ $\mathbf{\bar{A}} = \mathbf{Q} \mathbf{R} $ \textcolor{gray}{\Comment{QR decomposition}}
\State $\quad \quad$ $\mathcal{L}_\texttt{QR} = \frac{1}{r} \sum_{i<j} |\mathbf{R}_{ij}|$ \textcolor{gray}{\Comment{Compute QR-Ortho Loss}}
\State $\quad \quad$ $\mathcal{L}_\texttt{Total}=\mathcal{L}_\texttt{CE}+\lambda\mathcal{L}_\texttt{QR}$ \textcolor{gray}{\Comment{Total loss is the sum of cross-entropy (CE) Loss with QR-Ortho Loss}}
\State \textbf{end for}
\end{algorithmic}
\end{algorithm*}

\section{Pseudo Variables}
\label{appendix:pseudo_vars}
We generate random pseudo signals from three different time series process using the TimeSynth~\cite{maat2017timesynth} library. The pseudo signal was generated for each MTS instance. We used the default setting from the library.

\begin{enumerate}
\item \textbf{WhiteNoise.} Gaussian noise with a zero mean and a 0.3 standard deviation
\item \textbf{Sinusoidal.} Sine waves with an amplitude of one and a frequency of 0.25
\item \textbf{Gaussian Process.} Matern kernel was used with nu=1.5
\end{enumerate}

\section{Effect of $\lambda$}

\begin{figure}[ht]
\centerline{\includegraphics[scale=0.4]{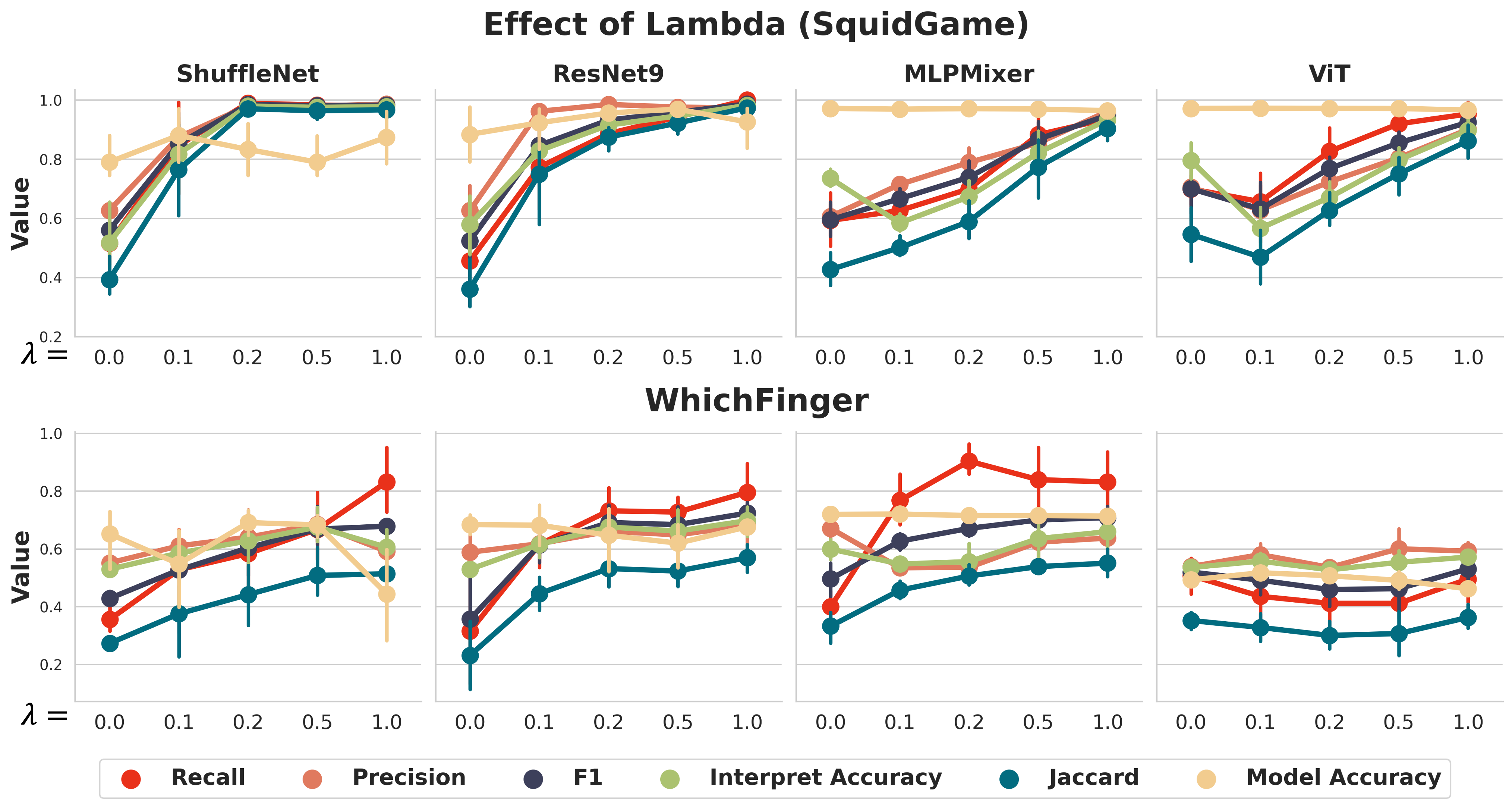}}
\caption{\textbf{(Top row)} CWRI related metrics are visualized for the SquidGame task. Here,~$\lambda=0$ is equivalent to using the CE loss only. We observe a general upward trend as~$\lambda$ is increased. \textbf{(Bottom row)} The general upward trend observed in the SquidGame is weaker in the WhichFinger.
}
\label{fig:lambda_grid}
\end{figure}

\label{appendix:lambdaeffect}
We conduct a grid search of $\lambda\in\{0, 0.1, 0.2, 0.5, 1.0\}$, a task-specific hyperparameter that controls the strength of QR-Ortho Loss(~\cref{eq:qr_ortho}), for SquidGame and WhichFinger tasks.  As shown in \cref{fig:lambda_grid}, metrics related to the evaluation of CWRI generally improve with larger~$\lambda$. However, excessive orthogonality can decrease model accuracy, as exemplified by the ShuffleNet on the WhichFinger task, which shows that identifying an optimal $\lambda$ is a critical consideration.

\begin{figure}[ht]
\centerline{\includegraphics[scale=0.35]{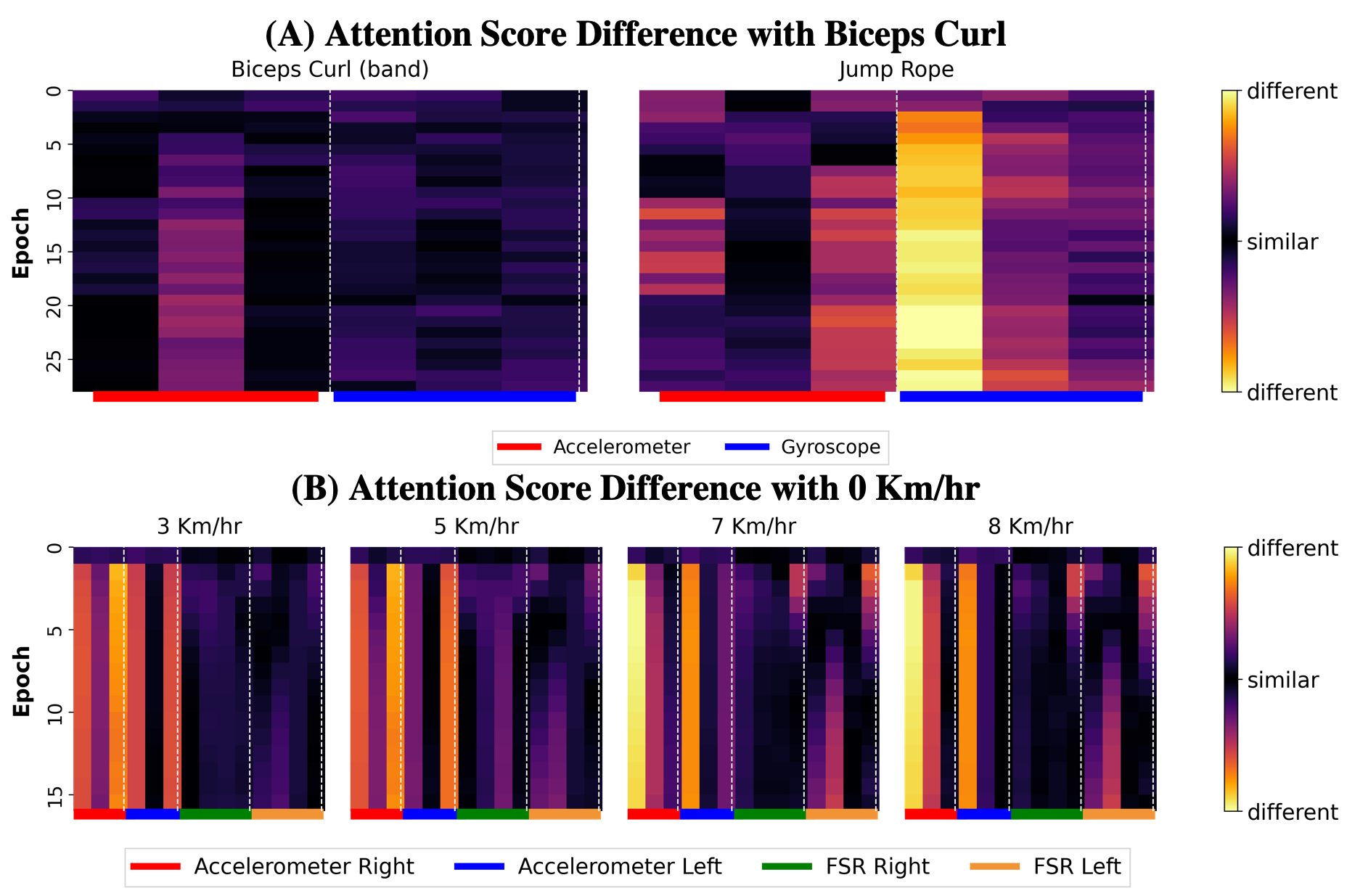}}
\caption{Visualization of Attention Score Difference. The Y-axis is the epoch and the X-axis corresponds to the feature. A greater disparity in attention scores is indicated by a more intense color brightness.
}
\label{fig:att_diff}
\end{figure}

\section{Alignment with Domain Knowledge}
\label{appendix:alignmentwithdomain}
In the domains of vision and natural language processing, XAI methodologies have aimed to demonstrate the alignment of their explanations with human intuition. This is typically achieved by superimposing attention maps onto images highlighting class-related features~\cite{selvaraju2017grad} or visualizing the intensity of attention signals connecting pairs of related words~\cite{vaswani2017attention}. Our methodology adopts a parallel strategy but by visualizing the differences between attention vectors derived from closely related classes. By doing so, we show that the explanations made with CAFO align with domain knowledge but also provide a deeper insight into the subtleties differentiating one class from another. 

We employ real-world datasets, namely Gilon and MS, to illustrate this methodology. For visual analysis, we gather class attention prototypes $\mathbf{\bar{a}}_{c}\in\mathbb{R}^D$(\cref{eq:class_prototype}) for each epoch $E$, forming a class attention matrix $\mathbf{\tilde{A}}_{c}\in\mathbb{R}^{E\times D}$. Here, we visualize $|\mathbf{\tilde{A}}_{c} - \mathbf{\tilde{A}}_{c'}| (c\neq c')$. Our first validation, showcased in \cref{fig:att_diff}-A, tests the hypothesis that \textit{similar activities should yield similar attention scores}. The comparison between (a) Bicep Curl (w/o band) and Bicep Curl with Band, and (b) Bicep Curl (w/o band) and Jump Rope from the MS data, supports this. The attention scores of similar activities (Bicep Curl (w/o band) and Bicep Curl with Band) show minor divergence, whereas there is a large difference in the attention scores when comparing Bicep Curl to Jump Rope.

Subsequently, we verify that \textit{accelerometer features should be the main feature in differentiating varying speed ranges} in \cref{fig:att_diff}-B. For this experiment, we utilized a downstream regression task from the Gilon, where the objective is to regress the speed at which the users are moving on a treadmill. Here, we train with mean square error loss and without QR-Ortho regularizer on the Gilon dataset. We visualize the difference between the 0km/hr (stationary position) and the remaining speed labels (3, 5, 7, and 8 km/hr). We first observe that the attention divergence is large for accelerometer features across all speed ranges. Moreover, this divergence escalates with larger speed difference, denoted by the brightness in the heatmap. This change in color difference highlights the role of accelerometer features in differentiating speed ranges, a finding that is consistent with established knowledge.

\putbib[appendix]
\end{bibunit}

\end{document}